\pdfoutput=1
\documentclass{article}

\PassOptionsToPackage{numbers, compress}{natbib}

\usepackage[final,main]{neurips_2026}

\usepackage[utf8]{inputenc} %
\usepackage[T1]{fontenc}    %
\usepackage{hyperref}       %
\usepackage{url}            %
\usepackage{xspace}
\usepackage{booktabs}       %
\usepackage{amsfonts}       %
\usepackage{nicefrac}       %
\usepackage{microtype}      %
\usepackage{xcolor}         %
\usepackage{graphicx}       %
\usepackage{amsmath}        %
\usepackage{amsthm}         %
\usepackage{subcaption}     %
\usepackage{float}          %
\usepackage[ruled,vlined]{algorithm2e}
\usepackage{comment}
\usepackage{import}
\usepackage{soul}
\usepackage{enumitem}

\usepackage{marvosym}
\usepackage[T1]{fontenc}

\newtheorem{thm}{Theorem}[section]

\newtheorem{remark}[thm]{Remark}
 
\newtheorem{theorem}[thm]{Theorem}
\newtheorem{corollary}[thm]{Corollary}
\newtheorem{lemma}[thm]{Lemma}
\newtheorem{proposition}[thm]{Proposition}

\newtheorem{definition}[thm]{Definition}

\newtheorem{assumption}[thm]{Assumption}

\newcommand{\x}{\times}

\newcommand{\dee}[1]{\mbox{$\mathrm{d}#1$}}

\makeatletter

\newdimen\w@dth

\def\setw@dth#1#2{\setbox\z@\hbox{\scriptsize $#1$}\w@dth=\wd\z@
\setbox\@ne\hbox{\scriptsize $#2$}\ifnum\w@dth<\wd\@ne \w@dth=\wd\@ne \fi
\advance\w@dth by 1.2em}

\def\t@^#1_#2{\allowbreak\def\n@one{#1}\def\n@two{#2}\mathrel
{\setw@dth{#1}{#2}
\mathop{\hbox to \w@dth{\rightarrowfill}}\limits
\ifx\n@one\empty\else ^{\box\z@}\fi
\ifx\n@two\empty\else _{\box\@ne}\fi}}
\def\t@@^#1{\@ifnextchar_ {\t@^{#1}}{\t@^{#1}_{}}}

\def\t@left^#1_#2{\def\n@one{#1}\def\n@two{#2}\mathrel{\setw@dth{#1}{#2}
\mathop{\hbox to \w@dth{\leftarrowfill}}\limits
\ifx\n@one\empty\else ^{\box\z@}\fi
\ifx\n@two\empty\else _{\box\@ne}\fi}}
\def\t@@left^#1{\@ifnextchar_ {\t@left^{#1}}{\t@left^{#1}_{}}}

\def\two@^#1_#2{\def\n@one{#1}\def\n@two{#2}\mathrel{\setw@dth{#1}{#2}
\mathop{\vcenter{\hbox to \w@dth{\rightarrowfill}\kern-1.7ex
                 \hbox to \w@dth{\rightarrowfill}}%
       }\limits
\ifx\n@one\empty\else ^{\box\z@}\fi
\ifx\n@two\empty\else _{\box\@ne}\fi}}
\def\tw@@^#1{\@ifnextchar_ {\two@^{#1}}{\two@^{#1}_{}}}

\def\tofr@^#1_#2{\def\n@one{#1}\def\n@two{#2}\mathrel{\setw@dth{#1}{#2}
\mathop{\vcenter{\hbox to \w@dth{\rightarrowfill}\kern-1.7ex
                 \hbox to \w@dth{\leftarrowfill}}%
       }\limits
\ifx\n@one\empty\else ^{\box\z@}\fi
\ifx\n@two\empty\else _{\box\@ne}\fi}}
\def\t@fr@^#1{\@ifnextchar_ {\tofr@^{#1}}{\tofr@^{#1}_{}}}

\newdimen\W@dth
\def\setW@dth#1#2{\setbox\z@\hbox{$#1$}\W@dth=\wd\z@
\setbox\@ne\hbox{$#2$}\ifnum\W@dth<\wd\@ne \W@dth=\wd\@ne \fi
\advance\W@dth by 1.2em}

\def\T@^#1_#2{\allowbreak\def\N@one{#1}\def\N@two{#2}\mathrel
{\setW@dth{#1}{#2}
\mathop{\hbox to \W@dth{\rightarrowfill}}\limits
\ifx\N@one\empty\else ^{\box\z@}\fi
\ifx\N@two\empty\else _{\box\@ne}\fi}}
\def\T@@^#1{\@ifnextchar_ {\T@^{#1}}{\T@^{#1}_{}}}

\def\T@left^#1_#2{\def\N@one{#1}\def\N@two{#2}\mathrel{\setW@dth{#1}{#2}
\mathop{\hbox to \W@dth{\leftarrowfill}}\limits
\ifx\N@one\empty\else ^{\box\z@}\fi
\ifx\N@two\empty\else _{\box\@ne}\fi}}
\def\T@@left^#1{\@ifnextchar_ {\T@left^{#1}}{\T@left^{#1}_{}}}

\def\Tofr@^#1_#2{\def\N@one{#1}\def\N@two{#2}\mathrel{\setW@dth{#1}{#2}
\mathop{\vcenter{\hbox to \W@dth{\rightarrowfill}\kern-1.7ex
                 \hbox to \W@dth{\leftarrowfill}}%
       }\limits
\ifx\N@one\empty\else ^{\box\z@}\fi
\ifx\N@two\empty\else _{\box\@ne}\fi}}
\def\T@fr@^#1{\@ifnextchar_ {\Tofr@^{#1}}{\Tofr@^{#1}_{}}}

\def\Two@^#1_#2{\def\N@one{#1}\def\N@two{#2}\mathrel{\setW@dth{#1}{#2}
\mathop{\vcenter{\hbox to \W@dth{\rightarrowfill}\kern-1.7ex
                 \hbox to \W@dth{\rightarrowfill}}%
       }\limits
\ifx\N@one\empty\else ^{\box\z@}\fi
\ifx\N@two\empty\else _{\box\@ne}\fi}}
\def\Tw@@^#1{\@ifnextchar_ {\Two@^{#1}}{\Two@^{#1}_{}}}

\def\to{\@ifnextchar^ {\t@@}{\t@@^{}}}
\def\from{\@ifnextchar^ {\t@@left}{\t@@left^{}}}
\def\two{\@ifnextchar^ {\tw@@}{\tw@@^{}}}
\def\tofro{\@ifnextchar^ {\t@fr@}{\t@fr@^{}}}
\def\To{\@ifnextchar^ {\T@@}{\T@@^{}}}
\def\From{\@ifnextchar^ {\T@@left}{\T@@left^{}}}
\def\Two{\@ifnextchar^ {\Tw@@}{\Tw@@^{}}}
\def\Tofro{\@ifnextchar^ {\T@fr@}{\T@fr@^{}}}

\makeatother

\newcommand{\cD}{\mathcal{D}}

\newcommand{\cL}{\mathcal{L}}

\newcommand{\cP}{\mathcal{P}}

\newcommand{\cX}{\mathcal{X}}
\newcommand{\cY}{\mathcal{Y}}

\newcommand{\bbE}{\mathbb{E}}

\newcommand{\bbP}{\mathbb{P}}

\newcommand{\bbR}{\mathbb{R}}

\DeclareMathOperator*{\argmax}{arg\,max}

\newcommand{\squarebrackets}[1]{\left[#1\right]}

\usepackage{bbm}
\newcommand{\indic}{\mathbbm{1}}

\newcommand{\Xsup}{X_{\mathrm{sup}}}
\newcommand{\ysup}{y_{\mathrm{sup}}}

\newcommand{\Xanc}{X_{\mathrm{anc}}}
\newcommand{\yanc}{y_{\mathrm{anc}}}

\newcommand{\Xeval}{X_{\mathrm{eval}}}

\usepackage[most]{tcolorbox}
\tcbuselibrary{skins,breakable}
\usetikzlibrary{arrows.meta, positioning, decorations.pathreplacing, calligraphy, shapes, calc, patterns, backgrounds}

\tcbset{
  leftbar/.style={
    enhanced jigsaw, breakable,
    frame hidden, sharp corners,
    boxrule=0pt,
    left=12pt, right=8pt, top=6pt, bottom=6pt,
    before skip=10pt, after skip=10pt,
    borderline west={2.5pt}{0pt}{#1},
  },
}

\tcolorboxenvironment{thm}{leftbar=blue!55!black, colback=blue!4}
\tcolorboxenvironment{theorem}{leftbar=blue!55!black, colback=blue!4}
\tcolorboxenvironment{proposition}{leftbar=blue!55!black, colback=blue!4}
\tcolorboxenvironment{prop}{leftbar=blue!55!black, colback=blue!4}
\tcolorboxenvironment{lemma}{leftbar=orange!70!black, colback=orange!5}
\tcolorboxenvironment{lem}{leftbar=orange!70!black, colback=orange!5}
\tcolorboxenvironment{corollary}{leftbar=green!45!black, colback=green!4}
\tcolorboxenvironment{cor}{leftbar=green!45!black, colback=green!4}
\tcolorboxenvironment{assumption}{leftbar=gray!60!black, colback=gray!6}
\tcolorboxenvironment{definition}{leftbar=purple!55!black, colback=purple!4}
\tcolorboxenvironment{defin}{leftbar=purple!55!black, colback=purple!4}
\tcolorboxenvironment{remark}{leftbar=gray!50!black, colback=gray!4}
\tcolorboxenvironment{rem}{leftbar=gray!50!black, colback=gray!4}
\tcolorboxenvironment{proof}{leftbar=teal!60!black, colback=teal!3, before skip=-5pt}

\newcommand{\cald}{\textbf{CALD}\xspace}

\title{When to Trust a PFN: Detecting Harmful Shift in Tabular Foundation Models}

\author{%
  Viet Nguyen\\
  University of Toronto\\
  \texttt{viet@cs.toronto.edu}\\
  \And
  Herman Bergström\\
  Chalmers University of Technology\\
  \texttt{hermanb@chalmers.se}\\
  \And
  Stephan Rabanser\\
  Princeton University\\
  \texttt{rabanser@princeton.edu}\\
  \And
  Rahul G. Krishnan\\
  University of Toronto\\
  \texttt{rahulgk@cs.toronto.edu}\\
}

\begin{document}

\maketitle

\begin{abstract}
  Tabular prior-fitted networks (PFNs) work well across tabular classification problems but can fail silently at deployment when distributions shift. We propose \textbf{CALD} (Calibrated Anchor-Loss Disagreement), a lightweight
  unsupervised monitor that triggers an alarm to warn when a PFN might fail. By finetuning the decoder to maximize disagreement with the test-set prediction while maintaining accuracy on an ID anchor set, \cald identifies harmful OOD batches through the resulting anchor loss, requiring no labels on test batches. We prove that the post-finetune anchor loss has a strictly lower expectation under sufficiently severe shift than under ID eval, yielding a one-sided test. Across 14 datasets covering tabular and structured distribution shift, on both TabPFN and TabICL, \cald matches or outperforms baselines on harmful shifts and is markedly more resilient to benign-shift false alarms.
\end{abstract}

\section{Introduction}

Several machine learning (ML) areas underwent a paradigm shift in the last decade, transitioning from training custom models on task-specific data toward leveraging in-context learning (ICL) to obtain predictions without any parameter updates~\citep{brown2020language,hoffmann2022training,wei2022chain,touvron2023llama,achiam2023gpt4}. The tabular modality enjoyed its own such paradigm shift after the development of amortized pre-trained models known as prior-fitted networks (PFNs)~\citep{muller2021transformers,hollmann2022tabpfn,hollmann2025accurate,qu2025tabicl,qu2026tabiclv2}, which provide classification and regression predictions on novel datasets in a single forward pass, outperforming the status quo of gradient boosted trees~\citep{chen2016xgboost,ke2017lightgbm,prokhorenkova2018catboost,grinsztajn2022why}. PFNs for tabular data, such as TabPFN~\citep{hollmann2022tabpfn,hollmann2025accurate,grinsztajn2025tabpfn} and TabICL~\citep{qu2025tabicl,qu2026tabiclv2}, are trained on millions of synthetic datasets to ingest a training set (the ``support set'') and test features (the ``query set'') as in-context prompts and return accurate predictions, capturing rich and complex inductive biases which make up the tabular modality.

Still, PFNs largely inherit the same fundamental limitations as classical ML, remaining brittle under distribution shift~\citep{quinonero2009dataset,recht2019imagenet,hendrycks2019benchmarking,koh2021wilds,taori2020measuring,miller2021accuracy,geirhos2020shortcut,yao2022wildtime}. Because their synthetic priors construct the support and query sets as independent draws from the same generative process, PFNs expect by design that the support and the query follow the same distribution. Consequently, the model may experience silent performance deterioration when deployment data drifts substantially from the support set%
, while it is unclear whether existing out-of-distribution (OOD) detectors designed for the classical paradigm~\citep{hendrycks2017baseline,liang2018enhancing,lee2018simple,liu2020energy,ren2021simple,sun2022knn,wang2022vim,huang2021importance,yang2024generalized} are suitable for the in-context regime. Reliable deterioration detection is crucial if we are to leverage PFNs' strong performance in high-stakes applications, such as healthcare~\citep{amodei2016concrete,obermeyer2019dissecting,finlayson2021clinician}.

A partial solution is to compute certain distributional statistics between training and deployment data to flag deviations in the covariate distribution%
~\citep{gretton2012kernel,lopezpaz2017revisiting,lipton2018detecting,rabanser2019failing,liu2020learning,zhao2022comparing}. However, these approaches are inherently label-agnostic and therefore sensitive to \emph{benign} shifts, where changes in the input space do not necessarily degrade model performance. %
More recently, methods that employ model disagreement have emerged as a more robust alternative~\citep{nakkiran2021deep,jiang2022assessing,ginsberg2023learning,tifrea2022semi,rosenfeld2023almost,nguyen2025disagreement}, leveraging the differences in a trained model's ability to disagree with its own predictions in- and out-of-distribution as an indicator of deterioration. %
However, these methods rely on perturbing the weights of a model that has been fine-tuned for a specific task, whereas an amortized PFN's weights remain fixed across tasks, and retraining is prohibitive and can degrade performance. Exploring how to apply disagreement-based detection in this regime is therefore a natural extension.

We introduce \textbf{CALD} (Calibrated Anchor-Loss Disagreement), a deployment deterioration monitoring mechanism for PFNs split into two phases (Figure~\ref{fig:overview}). In the \textbf{Calibration} phase (2), \cald samples a small labeled \emph{anchor} set from an in-distribution (ID) validation batch and treats the remainder as a held-out ID \emph{eval} set. The decoder is then briefly fine-tuned under a dual objective consisting of an agreement term on the anchor and a disagreement term on the eval set with respect to the PFN's own pseudolabels. Repeating this procedure across many validation draws yields a distribution of anchor losses in the ID setting. %
During the \textbf{Deployment} phase (3), we run the same fine-tuning procedure, substituting a (possibly OOD) unlabeled deployment batch for the eval set. We  compare the resulting anchor loss against the losses in the ID distribution, and flag it if it falls in the lower tail. This is motivated by the idea that a lower loss is achievable when the evaluated samples occupy a different region of the PFN's internal class-conditioned representations compared to the anchor samples.

Our contributions are as follows:
\begin{enumerate}
    \item \textbf{A detector for tabular ICL.} We propose \textbf{CALD}, to our knowledge the first deployment-time deterioration detector for PFNs that exploits their specific structure at inference.%
    \item \textbf{Supporting theory.} We derive conditions under which the post-finetune anchor loss is provably lower in expectation under sufficiently severe shift than under ID eval, with the gap controlled by the alignment of the anchor-agreement and eval-disagreement gradient fields.%
    \item \textbf{Benchmark evaluation.} We evaluate \cald on 14 datasets spanning covariate, domain, and structured covariate shift, across four shift severities, for both TabPFN and TabICL. \cald matches or outperforms 15 post-hoc baselines on harmful shifts, and on benign shifts achieves the most favorable trade-off among all methods evaluated (Table~\ref{tab:baselines}).%
    \item \textbf{Ablations and mechanism.} We show that the detection signal scales with the harm a shift inflicts rather than with its magnitude, and we verify the gradient-interference mechanism behind the theory directly on the deployed models, by measuring the proof's premise along the finetuning trajectories. We further provide a large suite of ablations, among which are the finetuning budget and its deployment cost, the disagreement weight, a frozen deployment configuration covered by the false-alarm guarantee, robustness to anchor quality, and detectability across shift types.
\end{enumerate}

\section{Related Work}
\label{sec:relatedwork}

\paragraph{Post-hoc OOD detection.}
A long line of work scores each input under a frozen classifier by inspecting the softmax shape~\citep{hendrycks2017baseline,liang2018enhancing,ovadia2019can,liu2020energy}, the distance from the training data in the model's feature space~\citep{lee2018simple,ren2021simple,sun2022knn,wang2022vim}, and auxiliary-loss gradients~\citep{huang2021importance}. \citet{yang2024generalized} survey the space. Label-agnostic two-sample tests~\citep{lopezpaz2017revisiting,liu2020learning,zhao2022comparing,rabanser2019failing} target distributional change directly and can therefore lead to false positives on benign covariate shifts. Every detector in this family relies on a single forward pass through a frozen model and we adopt the strongest of these as baselines.

\paragraph{Disagreement-based monitoring.}
The observation that independently trained models often agree on ID samples but disagree on OOD samples~\citep{nakkiran2021deep,jiang2022assessing} has been leveraged in several lines of work. Detectron~\citep{ginsberg2023learning} trains a small ensemble under a constrained-disagreement objective and derives its verdict from the deployment-batch disagreement rate. \citet{rosenfeld2023almost} add to this by introducing \emph{disagreement discrepancy} and provable error bounds. Finally, \textsc{D3M}~\citep{nguyen2025disagreement} constructs a harm-calibrated Bayesian three-stage test with explicit FPR/TPR guarantees. All three assume a fixed converged classifier whose weights are perturbed externally, which is prohibitive in the PFN regime, where training a second copy is not an option. \cald instead perturbs the decoder of a model %
and reads its signal from the \emph{anchor loss} rather than from the disputed prediction itself.

\paragraph{Adaptation under shift.}
A separate body of work tries to account for, rather than detect, distributional shifts. Methods include test-time adaptation~\citep{wang2021tent,niu2022eata,iwasawa2021t3a,sun2020ttt,wang2022cotta}, source-free DA~\citep{liang2020shot}, and feature alignment~\citep{saito2018maximum,vu2019advent}. This line of work is distinct from detection, answering the question of \emph{how should we change the model?} as opposed to \emph{should this prediction be trusted?}, and treating one as a substitute for the other is risky when the shift falls outside the family the adapter was designed for. The in-context translation of this idea is Drift-Resilient TabPFN~\citep{helli2024driftresilient}, which retrains the PFN prior on synthetic data sampled from \emph{evolving} structural causal models. On our 14-dataset benchmark it does \emph{not} close the deterioration gap: mean $\delta\mathrm{OOD} = -1.4$\,pp across all datasets and $-2.2$\,pp on the harmful subset, with PACS, CIFAR-10, and Waterbirds made measurably \emph{worse} by the adapted prior, and every harmful dataset remains harmful by the same Hoeffding criterion. Full numbers are in Appendix~\ref{sec:drift-resilient}. If anything, the result argues for placing a detector ahead of the adapter, including to flag when the adapter is itself the source of the problem.

\section{Background and Method}
\label{sec:background}

\subsection{Prior-Fitted Networks}

Prior-fitted networks (PFNs)~\citep{muller2021transformers} are tabular foundation models pretrained to perform classification in-context, without any task-specific gradient updates at deployment.
During pretraining, a PFN is exposed to millions of synthetic supervised tasks drawn from a designer-specified prior $p(\theta)$ over data-generating processes.
Each task contributes a labeled \textbf{support set} $\cD^{\mathrm{sup}} = (\Xsup, \ysup)$ of $n$ examples that serve as the in-context prompt, together with a query batch with known labels.
The model is optimized by minimizing the in-context cross-entropy between its predictions on the query and the ground-truth labels from the same generative process.
After pretraining, a PFN approximates the posterior predictive distribution (PPD):
\begin{equation}
    p(y \mid x,\, \cD^{\mathrm{sup}}) = \int p(y \mid x, \theta)\, p(\theta \mid \cD^{\mathrm{sup}})\, \dee\theta,
    \label{eq:ppd}
\end{equation}
amortizing Bayesian inference into a single forward pass over the sequence $(\Xsup, \ysup, x)$.

We view TabPFN and TabICL as PFN instantiations specialized to tabular classification and regression, with priors built from structural causal models (SCMs), Bayesian neural networks, and (for TabICL) tree-like inductive biases via XGBoost-fitted SCM edges. Both represent the current frontier in tabular supervised learning and are the empirical subjects of this work, though we note that the following exposition applies to the broader PFN family~\citep{ma2024tabdpt,balazadeh2025causalpfn,robertson2025dopfn}. 

\subsection{Failure Monitoring of PFNs for Classification}

PFN priors are largely ID: for each pretraining task, support and query sets are sampled from the same generative model. Thus, they do not account for queries whose covariates are OOD with respect to the support, which can produce post-hoc performance gaps beyond acceptable ranges. Detecting this requires a computable signal that distinguishes ID from OOD batches, from which we may build a hypothesis test. Per the model disagreement literature~\citep{ginsberg2023learning,nguyen2025disagreement,tifrea2022semi,rosenfeld2023almost}, it is easier to force a model to disagree with its own predictions on OOD than on ID data; classical-paradigm methods such as \textsc{Detectron}~\citep{ginsberg2023learning} and \textsc{D3M}~\citep{nguyen2025disagreement} exploit this by perturbing the weights of a fixed converged classifier. In the following, we show how forcing model disagreement can be done
for PFNs and extract a relevant signal that helps distinguish ID and OOD, building the foundations of \textbf{CALD}.

\subsection{In-Context Model Disagreement Optimization}

Let us consider how disagreement could be maximized in the ICL paradigm with PFNs. Denote the ID distribution by $p_0$ and let $\cD^{\mathrm{sup}} = (\Xsup, \ysup) \sim p_0^n$ be a size-$n$ ID support. We partition the queries into two disjoint roles: 

\textbf{Anchor} $(\Xanc, \yanc)\sim p_0$ with $\Xanc \cap \Xsup = \emptyset$. A held-out set of \emph{labeled} ID samples. Their true labels $\yanc$ are known and serve as a reference ID signal.

\textbf{Eval} $(\Xeval) \sim p_1 \overset{?}{=} p_0$. A batch of \emph{unlabeled} query points whose distributional origin is unknown at test time, the target of our disagreement forcing. 

Let $q_\theta$ denote a PFN. Conditional on $\cD^{\mathrm{sup}}$, its forward pass returns logits and predictions:
\begin{equation}
    z \in \bbR^{C},\quad q_\theta(\cdot \mid x, \cD^{\mathrm{sup}}) := \mathrm{softmax}(z) \in \Delta^{C-1},\quad \hat{y} = \mathrm{argmax}_y\, q_\theta(y \mid x, \cD^{\mathrm{sup}})
\end{equation}
for each $x \in \Xanc \cup \Xeval$, where $C$ is the number of classes. Write $\hat{y}_\mathrm{eval}$ for predictions (pseudolabels) the model assigns to $X_\mathrm{eval}$.

To force $q_\theta$ to disagree with itself on $X_{\mathrm{eval}}$ while remaining in agreement with the true labels from the anchor, we compute the cross-entropy (CE) loss rewarding correct predictions for the anchor:  
\begin{equation}
    \cL_{\mathrm{anc}}(\theta) = \frac{1}{|X_\mathrm{anc}|} \sum_{(x,y) \in (X_\mathrm{anc}, y_\mathrm{anc})} -\log q_\theta(y | x, \cD^\mathrm{sup}) 
\end{equation}
and the Disagreement Cross-Entropy (DCE) loss~\citep{ginsberg2023learning} for the eval set, rewarding the model for placing low probability on its own initial pseudolabels:
\begin{equation}
    \cL_\mathrm{eval}(\theta) = \frac{1}{|X_\mathrm{eval}|} \sum_{(x,y) \in (X_\mathrm{eval}, \hat{y}_\mathrm{eval})} \squarebrackets{\frac{1}{C-1} \sum_{c \not= y} -\log q_\theta(c \mid x, \cD^\mathrm{sup})}
\end{equation}
If minimizing the CE loss bumps logits for the correct class up, minimizing the DCE loss pushes logits for the pseudolabel class down. 
Performing gradient descent on the joint objective
\begin{equation}
    \cL(\theta) = \cL_\mathrm{anc}(\theta) + \beta\,\cL_\mathrm{eval}(\theta)
    \label{eq:objective}
\end{equation}
with disagreement weight $\beta = 1$ throughout unless otherwise stated (Appendix~\ref{sec:abl-beta} ablates it), thus has the effect of forcing $q_\theta$ to disagree on the eval set while remaining performant on ID samples.

\subsection{CALD: Calibrated Anchor-Loss Disagreement}
\label{sec:cald}

\input{figure1.tex}

Finetuning $q_\theta$ under~\eqref{eq:objective} exerts two competing pressures on $\theta$: $\cL_\mathrm{anc}$ pulls the decoder toward the true labels on the anchor, while $\cL_\mathrm{eval}$ pushes it away from its own pretrained pseudolabels on the eval batch. The driving observation behind \cald is the way these pressures resolve.

The \textit{post-finetune anchor loss} $\lambda := \cL_\mathrm{anc}(\theta_T)$ is systematically \emph{lower} when $\Xeval$ is OOD than when $\Xeval$ is ID (Figure~\ref{fig:overview}, Panel~2). \cald turns this one-dimensional signal into a calibrated one-sided test by collecting a null record of $\lambda$ values under known-ID conditions and rejecting at deployment when the observed $\lambda^*$ falls into the lower tail of that record. We defer justification to Section~\ref{sec:method}.

Given hyperparameters $T$ (finetune steps), $K$ (calibration rounds), $\eta$ (learning rate), and $\alpha$ (test level), \cald proceeds in two phases (full pseudocode in Appendix~\ref{sec:algorithm}):

\begin{enumerate}
    \item \textbf{Calibration phase.} For each round $k = 1, \ldots, K$, draw an ID eval batch $\Xeval^{(k)}\sim p_0$, freeze its pseudolabels $\hat{y}_\mathrm{eval}^{(k)}$ assigned by $q_{\theta_0}$ to $\Xeval^{(k)}$, and reset $\theta \leftarrow \theta_0$. At each of $T$ gradient steps, the PFN forward-passes the joint sequence $(\Xsup, \ysup,\, \Xanc,\, \Xeval^{(k)})$: the anchor logits supply $\cL_\mathrm{anc}(\theta)$ against the true labels $\yanc$, the eval logits supply $\cL_\mathrm{eval}(\theta)$ against the frozen pseudolabels $\hat{y}_\mathrm{eval}^{(k)}$, and the joint loss is descended via $\theta \leftarrow \theta - \eta\,\nabla_\theta\,\cL(\theta)$ with only the decoder trainable. Record the post-finetune anchor loss $\lambda^{(k)} := \cL_\mathrm{anc}(\theta_T)$. After $K$ rounds, we have an empirical null record $\{\lambda^{(k)}\}_{k=1}^{K}$ under known-ID conditions.
    \item \textbf{Deployment phase.} On a deployment batch $\Xeval^*$ of unknown origin, freeze its pseudolabels $\hat{y}_\mathrm{eval}^*$, reset $\theta \leftarrow \theta_0$, and run the same $T$-step finetune, forward-passing $(\Xsup, \ysup,\, \Xanc,\, \Xeval^*)$ at each step and recording the deployment statistic $\lambda^* := \cL_\mathrm{anc}(\theta_T)$. The verdict is a one-sided test against the calibration record:
    \begin{equation}
        \lambda^* < \mathrm{Quantile}_\alpha\!\left(\lambda^{(1)}, \ldots, \lambda^{(K)}\right) \;\Longrightarrow\; \text{declare OOD}.
        \label{eq:verdict}
    \end{equation}
\end{enumerate}
The support set $\cD^\mathrm{sup}$ and anchor $(\Xanc, \yanc)$ are held fixed across both phases.

\section{Theory}
\label{sec:method}

This section justifies the headline of \S\ref{sec:cald}: that $\lambda_\mathrm{OOD} < \lambda_\mathrm{ID}$ in expectation, and that the resulting one-sided test has controlled Type-I error. \S\ref{sec:method-intuition} gives an informal mechanistic argument, \S\ref{sec:method-theory} states the formal results, and \S\ref{sec:method-empirical-calibration} flags one related claim (monotone scaling of the signal with harm severity) that we support empirically but do not prove.

\subsection{Intuitive argument}
\label{sec:method-intuition}

The two terms of $\cL(\theta)$ impose complementary forces on the shared decoder, and the way they resolve depends on whether the pretrained PFN is confident on its eval pseudolabels. When $\Xeval \sim p_0$, the PFN places high probability on its own (correct) pseudolabels, so $\nabla\cL_\mathrm{eval}$ is large in magnitude and points in a direction that conflicts with $\nabla\cL_\mathrm{anc}$; forcing disagreement on in-distribution points actively interferes with maintaining anchor agreement, and the anchor loss stays elevated after $T$ steps. When $\Xeval$ is shifted, however, the PFN is uncertain on the same pseudolabels: $\nabla\cL_\mathrm{eval}$ is small in magnitude and approximately orthogonal to $\nabla\cL_\mathrm{anc}$, so the disagreement pressure does not impede anchor descent and the anchor loss drops below its ID counterpart. The post-finetune $\lambda := \cL_\mathrm{anc}(\theta_T)$ therefore summarizes in a single scalar whether the eval-side optimization pressure can co-exist with anchor agreement, with $\bbE[\lambda_\mathrm{OOD}] < \bbE[\lambda_\mathrm{ID}]$.

\subsection{Theoretical guarantees}
\label{sec:method-theory}

We summarize the two main results that justify \cald; full statements, assumptions, and proofs are deferred to Appendix~\ref{sec:theory}.

The first result establishes that the post-finetune anchor loss is, in expectation, smaller under harmful shift than under in-distribution eval. It rests on a gradient-geometry analysis of the joint objective~\eqref{eq:objective}: the disagreement gradient on ID eval interferes with anchor descent (driving the anchor loss up by a structural amount $\gamma > 0$), while on far-OOD eval it becomes approximately orthogonal to the anchor gradient (controlled by a parameter $\epsilon$ that vanishes as $\|p_1 - p_0\|_{\mathrm{TV}} \to 1$).

\begin{theorem}[Main separation, simplified]
\label{thm:main-separation-main}
Let $\bbE[\lambda_{\mathrm{ID}}]$ and $\bbE[\lambda_{\mathrm{OOD}}]$ denote the expected post-finetune anchor losses when the eval batch is drawn from $p_0$ and $p_1$ respectively. Under the assumptions stated in Appendix~\ref{sec:theory}, there exist constants $\gamma > 0$ and $\epsilon \ge 0$ (the ID interference strength and the OOD feature-orthogonality parameters, respectively) such that
\begin{equation}
    \bbE[\lambda_{\mathrm{ID}}] - \bbE[\lambda_{\mathrm{OOD}}]
    \;\ge\; T\eta\,(\gamma - \epsilon\, B)
    \;-\; O(T^2 \eta^2),
    \label{eq:main-separation-main}
\end{equation}
where $B$ collects bounded architecture- and data-dependent constants. Whenever $\epsilon < \gamma / B$ and $T\eta$ is in the small-step regime, the right-hand side is strictly positive and $\bbE[\lambda_{\mathrm{OOD}}] < \bbE[\lambda_{\mathrm{ID}}]$.
\end{theorem}

The separation gap grows linearly in the finetune budget $T\eta$ (within the small-step regime), is driven by a structural ID quantity $\gamma$ that depends only on the pretrained classifier and $p_0$, and admits its severity dependence through $\epsilon$. The full version (Theorem~\ref{thm:main-separation}) makes the constants explicit.

It is important to acknowledge that Theorem~\ref{thm:main-separation-main} certifies detection in the \emph{far-OOD regime only}. At finite shift severity its condition $\epsilon < \gamma/B$ is not satisfied on any of the ten benchmarks on which we estimated it. Outside the far-OOD regime we therefore make no formal claim and instead measure the proof's operative premise directly: the time-averaged gradient-interference gap $\Delta_{\mathrm{traj}} := \bbE_{\mathrm{OOD}}[\langle \bar g_a, \bar g_e\rangle] - \bbE_{\mathrm{ID}}[\langle \bar g_a, \bar g_e\rangle]$ is positive on nine of the ten datasets, and the two cells where it is weakest or negative are exactly the two with near-zero $\lambda$-separation (\S\ref{sec:abl-cross-term}, Figure~\ref{fig:mechanism}(b); Table~\ref{tab:abl-traj}). These measurements support the theorem's mechanism at moderate shift severity and help predict where the detector will struggle. A useful extension for future work would be to derive a finite-$\epsilon$ bound using constants estimated from data instead of worst-case values.

The second result certifies the practical decision rule. The detector observes only $K$ calibration anchor losses and a single deployment statistic, and rejects ID when the deployment value falls below the empirical $\alpha$-quantile of the calibration set. A standard exchangeability argument controls the false-alarm rate.

\begin{proposition}[Finite-sample Type-I error control, simplified]
\label{prop:conformal-main}
Suppose the anchor set, support set, and finetune hyperparameters are fixed across rounds, the eval batches are i.i.d.\ draws, and the inner finetune loop is deterministic given its inputs. Under the null $H_0\!:\!\Xeval^* \sim p_0$, the one-sided test in~\eqref{eq:verdict} satisfies
\begin{equation}
    \bbP_{H_0}(\text{reject}\ H_0) \;\le\; \alpha + \frac{1}{K+1}.
    \label{eq:conformal-main}
\end{equation}
\end{proposition}

The slack $1/(K+1)$ is the cost of using a finite calibration set; it vanishes as $K \to \infty$. Combining Theorem~\ref{thm:main-separation-main} with Proposition~\ref{prop:conformal-main} gives a detector with controlled Type-I error and a non-vanishing power gap whenever the OOD shift is severe enough to satisfy the feature-orthogonality regime; we expand on the corresponding power analysis in Appendix~\ref{sec:theory-conformal}.

Proposition~\ref{prop:conformal-main} requires the calibration and deployment batches to be exchangeable under the null hypothesis. Here, ``in-distribution'' means that the batches come from the same covariate distribution; no geometric boundary between ID and OOD data is needed (Definition~\ref{def:id-batch}, Appendix~\ref{sec:theory-conformal}). The hyperparameters must also be fixed in advance so that calibration and deployment use the same scoring procedure. The per-cell learning-rate selection in Table~\ref{tab:baselines} (Appendix~\ref{sec:main-results-full}) uses both ID and known OOD trajectories, so the guarantee does not cover that procedure. A fixed learning rate avoids this dependence on known OOD data and appears practical: it gives similar average detection performance, while observed no-shift false-alarm rates remain close to the theoretical bound across the tested settings (see Appendix~\ref{sec:abl-frozen}).

\subsection{Calibration to harm severity is empirical}
\label{sec:method-empirical-calibration}

Theorem~\ref{thm:main-separation-main} delivers a separation in the far-OOD regime, but it does not predict that the detection signal scales \emph{monotonically} with the magnitude of the shift or with the model's accuracy drop. We observe this calibration-to-harm property empirically (Figure~\ref{fig:calibration_to_harm}): on a controlled 2D Gaussian sweep and on a Waterbirds severity sweep, AUROC rises smoothly from $0.5$ at zero shift to $1.0$ at catastrophic shift, while small benign translations remain near chance; large benign translations do fire, as Lemma~\ref{lem:density-ratio} predicts for any feature-space monitor. We treat this as a desirable structural property of the gradient cross-term $\langle g_a, g_e\rangle$ tracking shift severity; we explore it in \S\ref{sec:abl-calibration}.

\section{Experiments}
\label{sec:experiments}

We evaluate \cald against post-hoc OOD-detection baselines on 14 datasets and the two PFN architectures (TabPFN and TabICL). Our experiments are designed to support three claims: (i) \cald reliably flags harmful distribution shift, (ii) it is markedly more resilient to benign-shift false alarms than label-blind baselines, though not immune to them, and (iii) the gradient mechanism of \S\ref{sec:method} is the one operating in the deployed models. Extended experimental details, dataset descriptions, and baselines are deferred to Appendix~\ref{sec:experimental-details}; hyperparameter sweeps, deployment cost, design choices, the frozen deployment configuration, anchor robustness, feature collapse, a two-signal harm detector, and detectability across shift types are ablated in Appendix~\ref{sec:appendix-ablations}. The code repository for \cald{} is available at \url{https://github.com/teivng/cald}.

\subsection{Setup}
\label{sec:experimental-setup}

\textbf{Datasets:} we evaluate on 14 datasets covering two regimes. (i) \emph{Raw tabular shift} (7 datasets): UCI Heart Disease, Diabetes Readmission, Adult Income, Bank Marketing, Credit Default, Covertype, and Folktables ACS Income; and (ii) \emph{structured shift via foundation-model representations} (7 datasets): Camelyon17, CIFAR-10$\to$10.1, PACS, Waterbirds, CelebA, CivilComments, and Amazon Reviews, where image and text inputs are encoded by pretrained foundation models (DINOv2, ResNet-18, Jina-v3) and the resulting representations are passed to TabPFN/TabICL as ordinary tabular features. Per-dataset descriptions, ID/OOD splits, and a survey of the resulting accuracy deteriorations across the severity spectrum (catastrophic, moderate, negligible, benign) are given in Appendices~\ref{sec:dataset-description} and~\ref{sec:deterioration-survey}. \textbf{Baselines:} we compare against 15 post-hoc OOD detection signals computed without finetuning: MSP, Entropy, Energy, Mahalanobis distance, Relative Mahalanobis (RMD), KNN distance, ViM, BBSD-softmax, BBSD-embedding, MMD, H-divergence, GradNorm, GradOrth, GAIA-Z, and WeiPer. Detailed descriptions are in Appendix~\ref{sec:baselines-detail}.

\textbf{PFN models.} We apply \cald to two SoTA PFN models: \textbf{TabPFN v2.5}~\citep{grinsztajn2025tabpfn} and \textbf{TabICL v2}~\citep{qu2026tabiclv2}. 

\textbf{Evaluation metrics.} We report \textbf{TPR@5\%\,FPR} (primary), the true-positive rate at a fixed 5\% false-positive operating point set by the empirical $\alpha = 0.05$ quantile of the calibration anchor-loss distribution, and \textbf{AUROC} (secondary).

\subsection{Main Results: Deterioration Detection}
\label{sec:main-results}

Table~\ref{tab:baselines} reports TPR@5\%FPR for \cald and the 15 post-hoc baselines across 14 datasets and both PFN architectures. All results use 10 seeds, 50 calibration rounds, $T = 50$ finetune steps, and eval batches of 128 points; the full table with standard deviations and the learning-rate selection protocol are deferred to Appendix~\ref{sec:main-results-full}.

\textbf{\cald{} reliably flags harmful shift.} Across the seven harmful columns, the better \cald variant scores TPR $\ge 0.99$ on CivilComments, Camelyon, Waterbirds, CIFAR, PACS, and Covertype, whereas for UCI Heart, both variants reach only $0.71$ TPR at the chosen calibration size.

\textbf{\cald{} is more resilient to benign shifts than baselines, but not immune.} On four of seven benign datasets the better \cald variant achieves correct-silence rate $\ge 0.92$ (Adult, Amazon, Folktables-TabPFN, CelebA-TabICL); on Credit Default, TabICL recovers $0.76$ silence where TabPFN false-alarms at $0.00$. The honest weaknesses are Bank ($\le 0.02$), Diabetes Readmission ($\le 0.08$), and Folktables under TabICL ($0.42$). All three have small accuracy gaps ($|\Delta| \le 4.4\%$) close to the Hoeffding floor, and Appendix~\ref{sec:scope} dissects the mechanism: the post-finetune anchor loss is a feature-space quantity, sensitive to any reweighting of the input domain, whereas the harm label is determined entirely by reweighting on the model's mistake region (Lemma~\ref{lem:density-ratio}). When a benign shift moves substantial mass on the complement of the mistake region, the gradient geometry registers a real distributional difference and \cald fires.

\textbf{TabICL outperforms TabPFN, especially on subpopulation shift.} TabICL strictly dominates TabPFN on four datasets whose shift is defined at the subgroup level: Camelyon ($0.19 \!\to\! 1.00$), Waterbirds ($0.46 \!\to\! 0.99$), CelebA ($0.27 \!\to\! 0.97$), and Credit Default ($0.00 \!\to\! 0.76$). On eight of the other ten datasets the two architectures track within $0.05$; Folktables ($0.98 \!\to\! 0.42$) is the one cell where TabPFN dominates. The likely culprit is decoder capacity: TabICL's trainable decoder (Linear $512 \!\to\! 1024 \!\to\! C$) is roughly an order of magnitude larger than TabPFN's ($192 \!\to\! 384 \!\to\! C$), and on subtle group-level shifts the disagree-finetune dynamics need enough decoder expressivity to separate ID from OOD anchor descent. TabPFN's tighter decoder appears to bottleneck the signal whenever the shift does not register as a global feature-space displacement.

\textbf{No baseline matches \cald's harm/benign trade-off.} D-MMD, H-divergence, and KNN each score $\ge 0.95$ on a majority of harmful columns but $\le 0.10$ on a majority of benign columns: their headline harmful-side TPR comes from \emph{always alarming}, not from harm-calibration. RMD shows the opposite trade-off: benign-side silence comparable to \cald, but harmful-side TPR collapsing to $0.40$--$0.67$ on the moderate shifts (Camelyon, Waterbirds, CIFAR). \cald is the only detector in the table that simultaneously achieves high harmful-side TPR on a majority of datasets and benign-side silence on a majority of datasets. We emphasize that this is a relative claim, not an absolute one: \cald does false-alarm on a minority of benign cells (Bank and Diabetes for both PFNs; Credit, Folktables, and CelebA for one of the two), and we view this remaining benign sensitivity as the realistic ceiling of a feature-space monitor that is blind to test labels.

\begin{table}[t]
\centering
\setlength{\tabcolsep}{2.5pt}
\renewcommand{\arraystretch}{0.95}
\scriptsize
\caption[Detection performance (TPR@5\% FPR) across 14 shifts]{\textbf{Detection performance (TPR@5\%\,FPR) across 14 shifts ordered by harm.} Columns are sorted by ID$\rightarrow$OOD accuracy drop $\Delta$ (second header row, \%). Harmful/benign split via Hoeffding bound at $\delta\!=\!0.05$, $N\!=\!\min(\text{rounds}\!\times\!\text{eval},\,|\mathcal{D}_{\text{OOD}}|)\!=\!1{,}280$ from the deterioration survey of Appendix~\ref{sec:deterioration-survey}: $\Delta^{*}\!\approx\!4.8\%$. For \textit{harmful} shifts ($\Delta\!>\!\Delta^{*}$), TPR = rate of correct alarms. For \textit{benign} shifts ($\Delta\!\le\!\Delta^{*}$), TPR = rate of correct silence (1\,$-$\,false-alarm rate). A good detector scores high in \emph{both} regions.}
\label{tab:baselines}
\begin{tabular*}{\textwidth}{@{\extracolsep{\fill}}lccccccc|ccccccc@{}}
\toprule
 & \multicolumn{7}{c}{\textit{Benign} ($\Delta\!\le\!4.8\%$)} & \multicolumn{7}{c}{\textit{Harmful} ($\Delta\!>\!4.8\%$)} \\
\cmidrule(lr){2-8} \cmidrule(lr){9-15}
Method & Adult & Diab. & Amazon & Credit & Folktbl. & Bank & CelebA & CivilC. & Camel. & Waterb. & CIFAR & UCI & PACS & Covert. \\
$\Delta$ & $-$9.4 & $-$2.3 & $+$1.5 & $+$2.8 & $+$3.1 & $+$4.4 & $+$4.5 & $+$7.4 & $+$8.9 & $+$11.6 & $+$12.4 & $+$15.6 & $+$27.9 & $+$83.5 \\
\midrule
MSP & .00 & .64 & .78 & .55 & .53 & .00 & .63 & \textbf{1.00} & .97 & \textbf{1.00} & .96 & .74 & \textbf{1.00} & \textbf{1.00} \\
Entropy & .00 & .61 & .80 & .61 & .50 & .00 & .39 & \textbf{1.00} & .98 & \textbf{1.00} & .97 & .77 & \textbf{1.00} & \textbf{1.00} \\
Energy & .00 & .00 & .77 & .00 & .86 & \textbf{.42} & .04 & \textbf{1.00} & \textbf{1.00} & \textbf{1.00} & .97 & .66 & \textbf{1.00} & .50 \\
Mahal. & .00 & .00 & .71 & .00 & .12 & .05 & .00 & .32 & .43 & .18 & .55 & \textbf{1.00} & \textbf{1.00} & \textbf{1.00} \\
RMD & .69 & .59 & .87 & \textbf{.92} & .72 & .05 & .42 & .51 & .40 & .52 & .67 & .47 & \textbf{1.00} & \textbf{1.00} \\
KNN & .00 & .00 & .78 & .00 & .23 & .00 & .00 & \textbf{1.00} & .95 & .97 & .10 & .99 & \textbf{1.00} & \textbf{1.00} \\
ViM & .00 & .10 & .87 & .00 & .63 & .13 & .81 & .23 & .54 & .18 & .15 & .93 & .70 & \textbf{1.00} \\
BBSD-S & .00 & .40 & .75 & .76 & .71 & .02 & .64 & .61 & .94 & .88 & .89 & .81 & \textbf{1.00} & \textbf{1.00} \\
BBSD-E & .00 & .00 & .76 & .67 & .66 & .00 & .49 & .58 & \textbf{1.00} & .99 & .72 & \textbf{1.00} & \textbf{1.00} & \textbf{1.00} \\
D-MMD & .00 & .00 & .66 & .00 & .00 & .00 & .54 & .91 & \textbf{1.00} & \textbf{1.00} & .68 & \textbf{1.00} & \textbf{1.00} & \textbf{1.00} \\
H-Div & .00 & .00 & .19 & .00 & .10 & .32 & .00 & .70 & \textbf{1.00} & \textbf{1.00} & .50 & \textbf{1.00} & \textbf{1.00} & \textbf{1.00} \\
GradNorm & .00 & .44 & .85 & .01 & .78 & .27 & .60 & \textbf{1.00} & .92 & .98 & .91 & .54 & .98 & .70 \\
GradOrth & .83 & \textbf{.77} & .83 & .69 & .86 & .02 & .46 & .66 & .54 & \textbf{1.00} & .95 & .23 & .97 & \textbf{1.00} \\
GAIA-z & .00 & .07 & --- & .01 & .80 & .29 & --- & --- & --- & --- & --- & .41 & --- & .59 \\
WeiPer & .18 & .17 & .89 & .00 & .67 & .17 & .81 & .34 & .45 & .26 & .21 & .89 & .62 & .73 \\
\midrule
\textsc{Cald} (TabPFN) & \textbf{1.00} & .08 & \textbf{.92} & .00 & \textbf{.98} & .02 & .27 & 1.00 & .19 & .46 & \textbf{1.00} & .71 & \textbf{1.00} & \textbf{1.00} \\
\textsc{Cald} (TabICL) & \textbf{1.00} & .00 & .87 & .76 & .42 & .00 & \textbf{.97} & .99 & \textbf{1.00} & .99 & .99 & .71 & \textbf{1.00} & \textbf{1.00} \\
\bottomrule
\end{tabular*}
\end{table}

\subsection{Calibration to harm}
\label{sec:abl-calibration}

A deterioration monitor is only useful if its alarm scales with the \emph{harm} a shift inflicts rather than with how novel the shifted inputs look. While Theorem~\ref{thm:main-separation-main} certifies a separation in the far-OOD limit, it does not determine how the statistic $\lambda$ behaves as severity grows (\S\ref{sec:method-empirical-calibration}). We therefore ask directly whether $\lambda$ tracks severity, and whether it stays silent on shifts of comparable feature-space magnitude that do not hurt the model.

We sweep shift severity in two settings. The first is a controlled 2D Gaussian in which harm and novelty can be dialed independently: rotating the class boundary is harmful, while translating the inputs along the boundary is benign, and both are matched in feature-space magnitude. The second is real subpopulation shift on Waterbirds, where the fraction $\alpha$ of spuriously-correlated minority samples in the eval batch sets the accuracy drop.

\begin{figure}[t]
    \centering
    \begin{subfigure}[b]{0.48\linewidth}
        \centering
        \includegraphics[width=\linewidth]{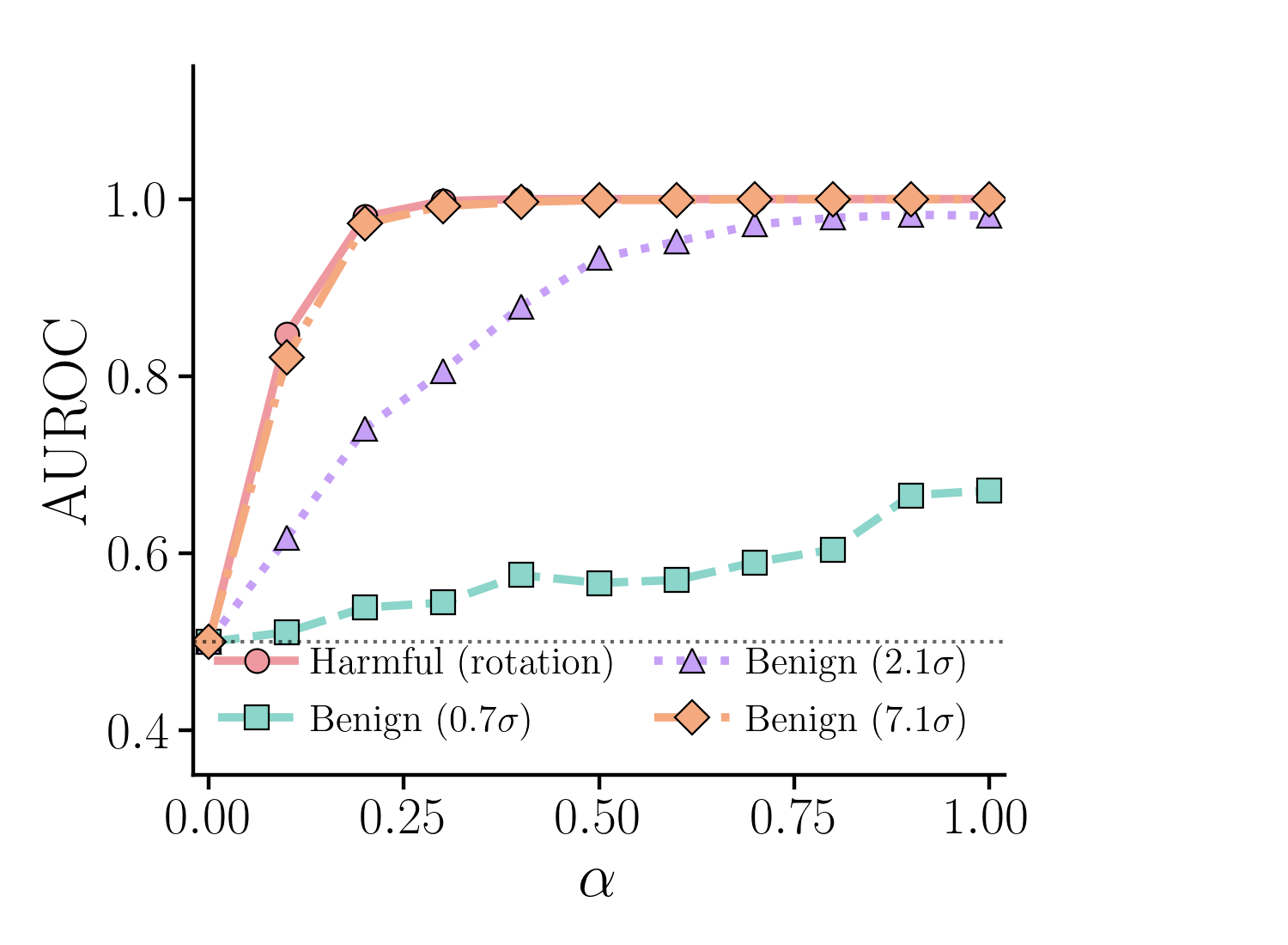}
        \caption{Synthetic 2D Gaussian.}
        \label{fig:calibration_synthetic}
    \end{subfigure}
    \hfill
    \begin{subfigure}[b]{0.48\linewidth}
        \centering
        \includegraphics[width=\linewidth]{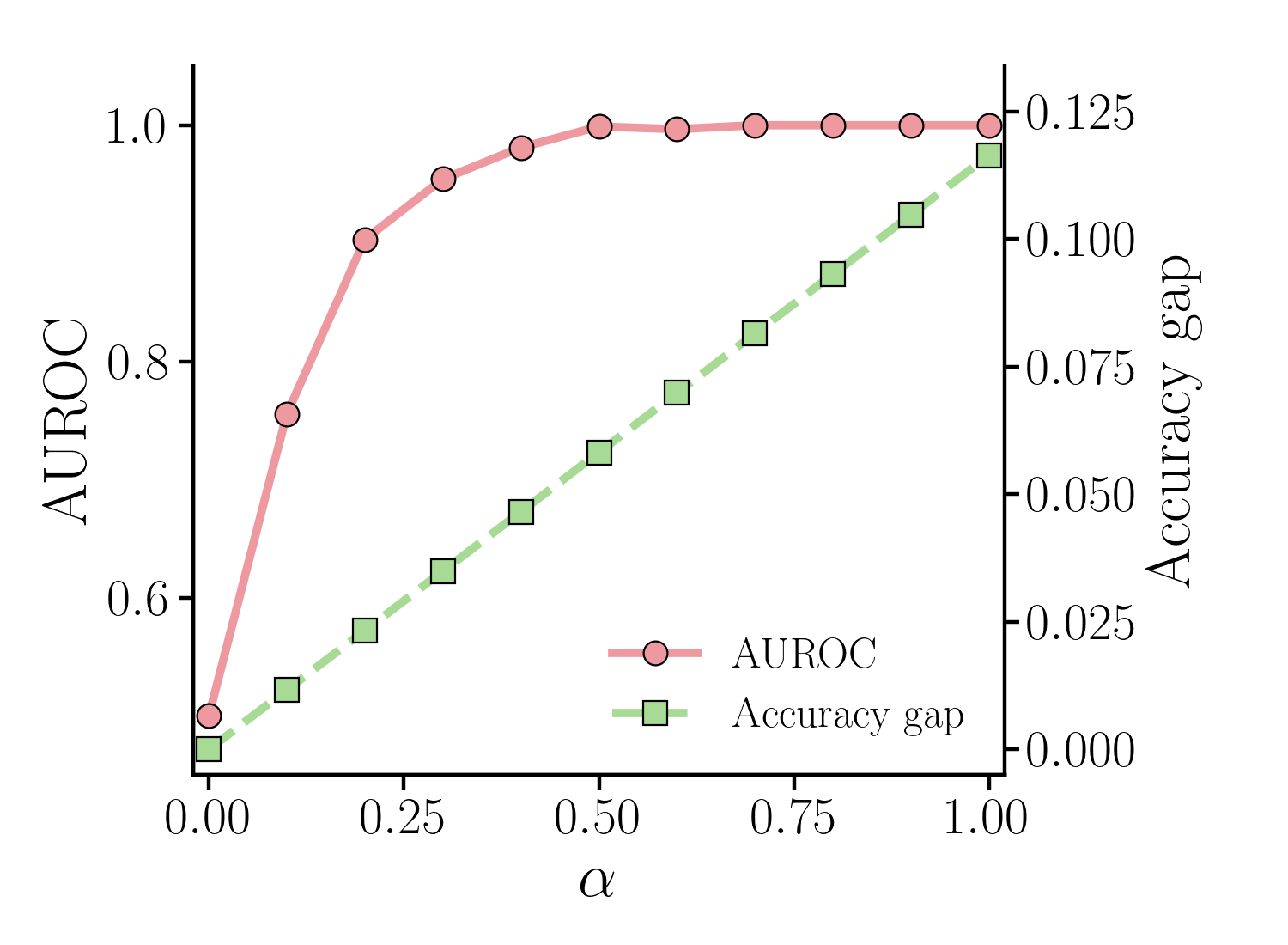}
        \caption{Real (Waterbirds, $\alpha$ sweep).}
        \label{fig:calibration_real}
    \end{subfigure}
    \caption{\textbf{Detection AUROC scales with shift severity.}
    \cald's AUROC rises monotonically with the induced accuracy gap on both controlled synthetic shift (left) and real subpopulation shift (right). At zero shift severity the AUROC is at chance; at maximum severity it saturates at $1.0$. On the synthetic sweep, a small benign translation ($0.7\sigma$) stays near chance, while larger benign translations ($2.1\sigma$, $7.1\sigma$) fire.}
    \label{fig:calibration_to_harm}
\end{figure}

In Figure~\ref{fig:calibration_to_harm}, on both the controlled and the real sweep, AUROC sits at chance when there is no shift, rises monotonically with the accuracy gap, and saturates at the most severe setting. On the benign axis, a small translation stays near chance throughout, while larger translations fire. This suggests that within these controlled regimes, $\lambda$ tracks severity, and distinguishes harm from novelty only when the benign shift is small, which is the property that separates \cald{} from the always-alarming baselines in Table~\ref{tab:baselines}. The property is empirical and should only be treated as such, as on some real-world benign datasets the post-finetune anchor loss separates ID from OOD despite a small accuracy gap, and Lemma~\ref{lem:density-ratio} shows why no feature-space monitor can be perfectly harm-calibrated; we refer the reader to Appendix~\ref{sec:fp-implications} for a closer look.

\subsection{The gradient cross-term}
\label{sec:abl-cross-term}

The proof of Theorem~\ref{thm:main-separation-main} tracks the anchor loss along the $T$ finetuning steps and asks how much of its descent the eval-disagreement term interferes with, under ID versus OOD eval. That interference is carried by the inner product $\langle g_a, g_e\rangle$ between the anchor-agreement gradient $g_a = \nabla\cL_\mathrm{anc}$ and the eval-disagreement gradient $g_e = \nabla\cL_\mathrm{eval}$. Expanding one step of the joint update to first order,
\begin{equation}
    \Delta\cL_\mathrm{anc}^{(t)} \;\approx\; -\eta\,\|g_a^{(t)}\|^2 \;-\; \eta\,\langle g_a^{(t)}, g_e^{(t)}\rangle,
    \label{eq:first-order-anchor}
\end{equation}
so the anchor loss descends at the vanilla rate $-\eta\|g_a\|^2$ corrected by the cross-term. Theorem~\ref{thm:main-separation-main} makes two claims about it. First, that the cross-term is what \emph{governs} anchor descent step by step, so that the first-order analysis describes the regime the detector actually runs in. Second, that its typical \emph{value} differs by regime: strongly negative on ID eval, where the disagreement gradient fights anchor descent and holds $\lambda$ up, and near zero on OOD eval, where the two gradients decouple and $\lambda$ falls. We examine each in turn.

\textbf{The cross-term governs anchor descent.}
We record $\cos(g_a^{(t)}, g_e^{(t)})$ and $\Delta\cL_\mathrm{anc}^{(t)}$ at every finetune step and compute the Spearman correlation between them per dataset, separately for ID and OOD eval. Equation~\eqref{eq:first-order-anchor} predicts a \emph{negative} correlation: the more aligned the two gradients, the faster the anchor loss falls.

\begin{figure}[t]
    \centering
    \begin{subfigure}[t]{0.49\linewidth}
        \centering
        \includegraphics[width=\linewidth]{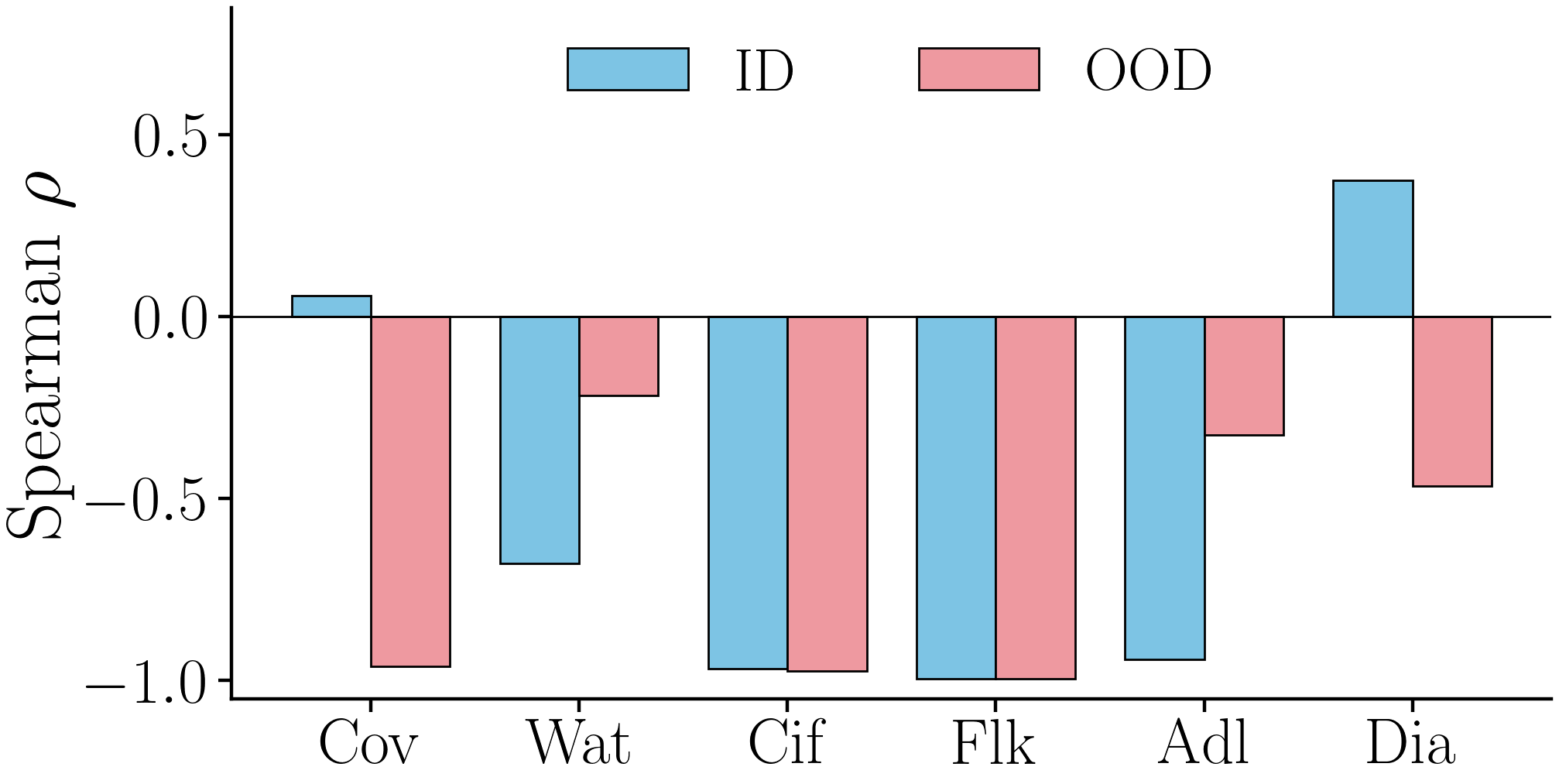}
        \caption{Per-step Spearman $\rho$ between $\cos(g_a, g_e)$ and $\Delta\cL_\mathrm{anc}$, ID vs.\ OOD; negative $\rho$ is the prediction of Equation~\eqref{eq:first-order-anchor}.}
        \label{fig:pred2_grad_cosine}
    \end{subfigure}\hfill
    \begin{subfigure}[t]{0.49\linewidth}
        \centering
        \includegraphics[width=\linewidth]{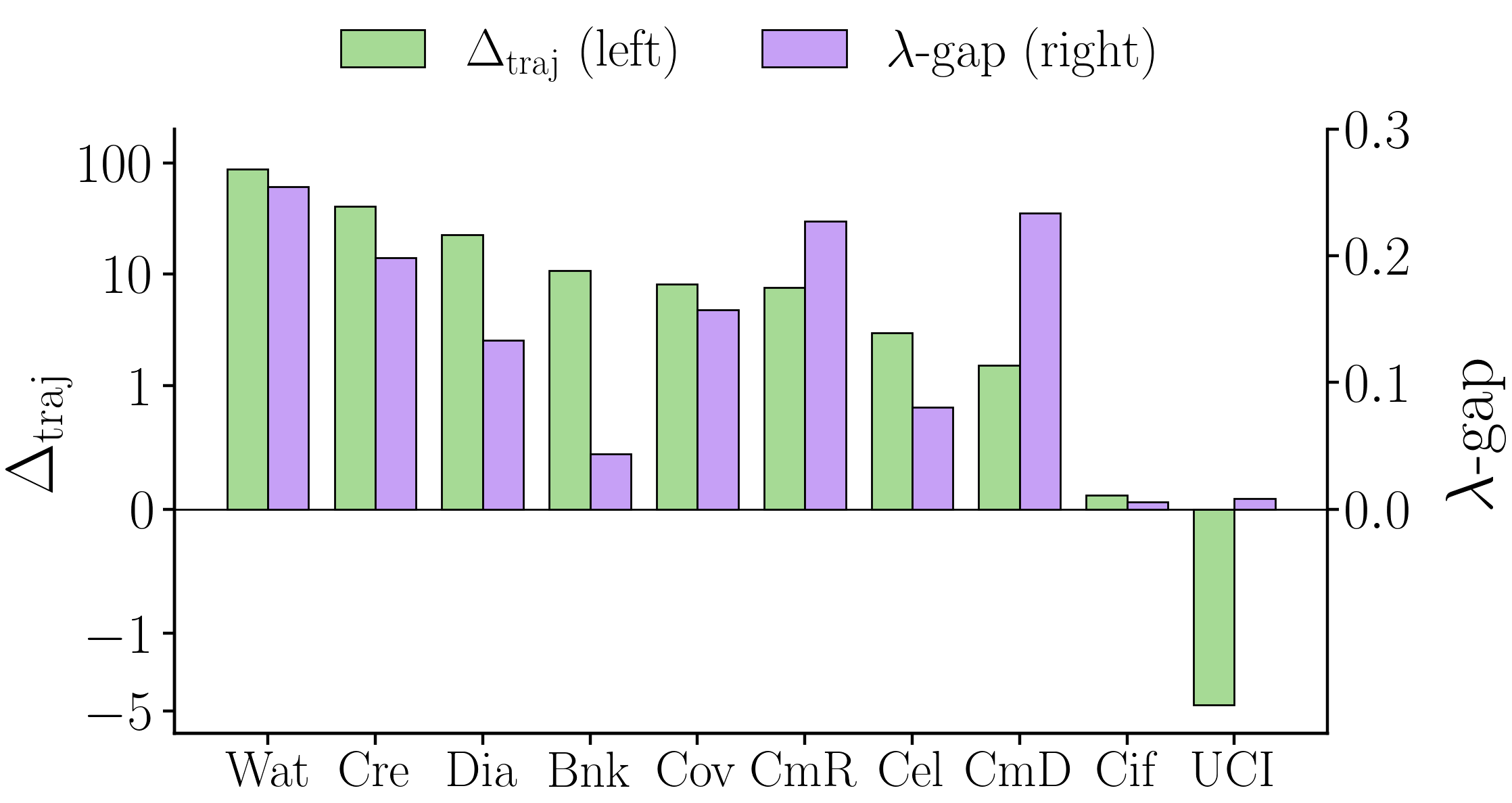}
        \caption{The proof's premise $\Delta_{\mathrm{traj}}$ of Equation~\eqref{eq:delta-traj} (left axis, symlog) and the $\lambda$-gap $\bbE[\lambda_{\mathrm{ID}}] - \bbE[\lambda_{\mathrm{OOD}}]$ it produces (right axis), per dataset; TabPFN, fixed $\eta = 10^{-3}$.}
        \label{fig:premise_delta_traj}
    \end{subfigure}
    \caption{\textbf{The gradient cross-term governs anchor descent, and its ID/OOD gap is what the detector reads.} (a) The first-order law holds in both regimes; the two near-zero or positive ID bars (Covertype, Diabetes Readmission) are cells where the cosine is pinned at its fighting extreme with no spread, see text. (b) The premise $\Delta_{\mathrm{traj}} > 0$ holds on nine of ten datasets, and the two cells where it is weakest or negative (CIFAR-10, UCI Heart) are exactly the two with near-zero $\lambda$-gap. \emph{Datasets:} Cov=Covertype, Wat=Waterbirds, Cif=CIFAR-10, Flk=Folktables, Adl=Adult Income, Dia=Diabetes Readmission, Cre=Credit Default, Bnk=Bank Marketing, CmR/CmD=Camelyon17 (ResNet18/DINOv2), Cel=CelebA, UCI=UCI Heart.}
    \label{fig:mechanism}
\end{figure}

In Figure~\ref{fig:mechanism}(a), the correlation is negative for every dataset under OOD eval, and for every dataset but two under ID eval. The two exceptions, Covertype and Diabetes Readmission under ID eval, are cells where the cosine is pinned near its fighting extreme across every step with the anchor loss rising throughout, which is the ID behavior the theorem describes; with no spread in the cosine there is nothing for a rank correlation to pick up. This suggests that the first-order law holds on all six datasets measured, and that it holds equally under ID and OOD eval. The law itself therefore does not separate the two regimes; what separates them must be the \emph{value} of the cross-term, which we measure next.

\textbf{The proof's premise.}
Theorem~\ref{thm:main-separation-main} lower-bounds the separation $\bbE[\lambda_{\mathrm{ID}}] - \bbE[\lambda_{\mathrm{OOD}}]$ by $T\eta(\gamma - \epsilon B)$, and the proof obtains this bound from the premise that by averaging the gradients over the $T$ steps (Proposition~\ref{prop:T-step}), the cross-term is more negative under ID eval than under OOD eval. Writing $\bar g_a$, $\bar g_e$ for the time-averaged gradients, the premise is
\begin{equation}
    \Delta_{\mathrm{traj}} \;:=\; \bbE_{\mathrm{OOD}}\big[\langle \bar g_a, \bar g_e\rangle\big] \;-\; \bbE_{\mathrm{ID}}\big[\langle \bar g_a, \bar g_e\rangle\big] \;>\; 0,
    \label{eq:delta-traj}
\end{equation}
and the constants $\gamma$ and $\epsilon$ are the theorem's worst-case way of guaranteeing it. The certifying condition $\epsilon < \gamma/B$ turns out not to hold on any of the ten benchmarks on which we estimate it (Appendix~\ref{sec:abl-certificate}, Table~\ref{tab:abl-certificate}): at finite severity, none of them is far-OOD in the theorem's sense. We therefore measure the premise itself, computing $\Delta_{\mathrm{traj}}$ by exact full-decoder autograd along the finetuning trajectories, alongside the $\lambda$-gap $\bbE[\lambda_{\mathrm{ID}}] - \bbE[\lambda_{\mathrm{OOD}}]$ it is supposed to produce.

In Figure~\ref{fig:mechanism}(b), $\Delta_{\mathrm{traj}}$ is positive on nine of the ten datasets, including the moderate-overlap benchmarks on which the formal condition is vacuous. The two datasets on which it is weakest or negative, CIFAR-10 and UCI Heart, are also the two with a near-zero $\lambda$-gap under this configuration, so the measurement anticipates the detector's weakest cells rather than rationalizing its successes after the fact. This suggests that while Theorem~\ref{thm:main-separation-main} certifies detection in the far-OOD regime only, the gradient mechanism it is built on is present across the near-OOD benchmarks and is what drives detection there. We refer the reader to Appendix~\ref{sec:abl-certificate} for the full analysis, including the estimated constants of the condition and the per-dataset $\lambda$-gap (Table~\ref{tab:abl-traj}).

\section{Conclusion and Limitations}
\label{sec:conclusion}

We introduced \cald, a deployment-time deterioration monitor for tabular PFNs that turns a brief decoder finetune of the frozen model into a detection signal. %
We ground our method in theory for harmful shift, and observe that across 14 datasets, \cald matches or outperforms 15 post-hoc baselines on harmful shifts, giving the most favorable harm/benign trade-off among the methods compared (Table~\ref{tab:baselines}).

The theory, however, is lacking on the benign side. Lemma~\ref{lem:density-ratio} essentially states that the accuracy gap depends only on mass moved on the model's mistake region $M$, while the post-finetune $\lambda$ integrates the gradient cross-term over the whole input domain, so a feature-space monitor \emph{cannot} achieve perfect harm-calibration. Empirically this surfaces on Bank, Diabetes, Credit, Folktables, and CelebA (each under at least one of the two PFNs in Table~\ref{tab:baselines}), benign shifts that nonetheless move substantial mass off $M$; Appendix~\ref{sec:scope} dissects the tabular cases. The conformal test also relies on ID exchangeability, so pure label shift, concept shift, open-set or adversarial inputs, and engineering errors are out of scope. Closing the harm gap likely requires a complementary signal tied to the density-ratio integral on $M$, sketched in Proposition~\ref{prop:harm-theorem} and left to future work.

Concretely, a tabular PFN's impending failure leaves a mechanistic trace in the gradient geometry of a brief decoder fine-tune, reachable from a frozen pretrained model with no extra training data. The limitations above bound this particular feature-space readout. \cald thus probes this trace through the decoder, while PFNs expose gradients with respect to the support set itself. Probes that exploit this, such as measuring how disagreement-labeled deployment points placed in the support affect predictions on the anchor, are a natural direction for richer, genuinely in-context deterioration sensors.

\newpage
\section*{Acknowledgements and Disclosure of Funding}

RGK gratefully acknowledges support from the Canada Research Chairs Program (CRC-2022-00049) and the Canada CIFAR AI Chairs Program. This research was funded in part by an NFRF Special Call Award (NFRFR-2022-00526), an NSERC Discovery Grant (RGPIN-2022-04546) and a CIHR Team Grant (RN549848--534930). HB is supported by Swedish Research Council Grant 2022-04748, as well as the Sweden-America Foundation. Resources used in preparing this research were provided, in part, by the Province of Ontario, the Government of Canada through CIFAR, and companies sponsoring the Vector Institute. VN humbly acknowledges the help and support provided by Eugenia Distefano and all members of the technical staff of the University of Toronto's Department of Computer Science in maintaining the Cluster. 

\bibliographystyle{plainnat}
\bibliography{main}

\newpage
\newpage
\appendix

\newpage

\section{Theory}
\label{sec:theory}

This appendix develops the theoretical foundations behind \cald. We first state the algorithm in pseudocode, then build up the formal argument that the post-finetuning anchor loss $\lambda$ is, in expectation, lower on OOD eval batches than on ID eval batches.

\subsection{Algorithm}
\label{sec:algorithm}

We restate \cald in pseudocode for completeness; the verbal description in Section~\ref{sec:cald} suffices for the reader who does not need step-by-step detail.

\begin{algorithm}[H]
  \caption{Calibrated Anchor-Loss Disagreement (\cald)}
  \label{alg:method}
  \SetAlgoLined
  \DontPrintSemicolon
  \KwIn{Support set $\cD^{\mathrm{sup}}$, anchor set $(\Xanc, \yanc)$,
        ID distribution $p_0$,
        steps $T$, rounds $K$, learning rate $\eta$, level $\alpha$}
  \KwOut{OOD verdict for test batch $\Xeval^*$}

  \BlankLine
  \tcp{--- Phase 1: Calibration on known-ID data ---}
  \For{$k = 1, \ldots, K$}{
      Sample $\Xeval^{(k)} \sim p_0$\;
      $\hat{y}^{(k)} \leftarrow \arg\max\; q_{\theta_0}\!\left(y \mid \Xeval^{(k)},\, \cD^{\mathrm{sup}}\right)$
      \tcp*{pseudolabels (no grad)}
      $\theta \leftarrow \theta_0$\;
      \For{$t = 1, \ldots, T$}{
          $\ell_{\mathrm{anc}} \leftarrow \mathcal{L}_{\mathrm{CE}}\!\left(q_\theta(y \mid \Xanc,\,
  \cD^{\mathrm{sup}}),\; \yanc\right)$\;
          $\ell_{\mathrm{dis}} \leftarrow \mathcal{L}_{\mathrm{DCE}}\!\left(q_\theta(y \mid \Xeval^{(k)},\,
  \cD^{\mathrm{sup}}),\; \hat{y}^{(k)}\right)$\;
          $\theta \leftarrow \theta - \eta\,\nabla_\theta\!\left(\ell_{\mathrm{anc}} +
  \ell_{\mathrm{dis}}\right)$\;
      }
      $\lambda^{(k)} \leftarrow \mathcal{L}_{\mathrm{CE}}\!\left(q_\theta(y \mid \Xanc,\, \cD^{\mathrm{sup}}),\;
  \yanc\right)$
      \tcp*{post-finetune anchor loss}
  }

  \BlankLine
  \tcp{--- Phase 2: Detection on test batch ---}
  $\hat{y}^* \leftarrow \arg\max\; q_{\theta_0}\!\left(y \mid \Xeval^*,\, \cD^{\mathrm{sup}}\right)$\;
  $\theta \leftarrow \theta_0$\;
  \For{$t = 1, \ldots, T$}{
      $\ell_{\mathrm{anc}} \leftarrow \mathcal{L}_{\mathrm{CE}}\!\left(q_\theta(y \mid \Xanc,\,
  \cD^{\mathrm{sup}}),\; \yanc\right)$\;
      $\ell_{\mathrm{dis}} \leftarrow \mathcal{L}_{\mathrm{DCE}}\!\left(q_\theta(y \mid \Xeval^*,\,
  \cD^{\mathrm{sup}}),\; \hat{y}^*\right)$\;
      $\theta \leftarrow \theta - \eta\,\nabla_\theta\!\left(\ell_{\mathrm{anc}} + \ell_{\mathrm{dis}}\right)$\;
  }
  $\lambda^* \leftarrow \mathcal{L}_{\mathrm{CE}}\!\left(q_\theta(y \mid \Xanc,\, \cD^{\mathrm{sup}}),\;
  \yanc\right)$\;

  \BlankLine
  \tcp{--- Verdict ---}
  \eIf{$\lambda^* < \mathrm{Quantile}_\alpha\!\left(\lambda^{(1)}, \ldots, \lambda^{(K)}\right)$}{
      \Return \textbf{OOD}\;
  }{
      \Return \textbf{ID}\;
  }
\end{algorithm}

\subsection{Justification of the Detection Signal}
\label{sec:theory-justification}

We develop a formal argument that the post-finetuning anchor loss $\lambda$ is, in expectation, \emph{lower} on OOD eval batches than on ID eval batches. The argument is a first-order gradient-geometry analysis of the joint objective in~\eqref{eq:objective}, combined with the tilted-attention interpretation of transformer PFNs due to \citet{nagler2023statistical}. We state the main proposition as a one-step result, then extend to the full $T$-step trajectory used in Algorithm~\ref{alg:method} using a standard small-step regime. The separation $\bbE[\lambda_\mathrm{OOD}] < \bbE[\lambda_\mathrm{ID}]$ is proven in the far-OOD regime (where the feature-orthogonality assumption holds with small parameter $\epsilon$; see Theorem~\ref{thm:main-separation}). Conversely, in the near-OOD regime the signed direction is a heuristic prediction that is empirically supported by a severity sweep rather than a theorem.

The argument formalizes the informal intuition in Section~\ref{sec:cald} that \emph{forcing disagreement on ID points tugs against anchor agreement, while forcing disagreement on OOD points does not}. We isolate the Frobenius inner product $\langle g_a, g_e\rangle_F$ of the anchor and disagreement gradients at initialization as the key quantity of interest: it controls whether anchor descent is impeded or free. We then show that $\langle g_a, g_e\rangle_F$ is controlled by a feature-space kernel $\langle \phi(x_a), \phi(x_e)\rangle$, and that this kernel collapses in the OOD limit under Nagler's exponential-tilting interpretation of attention.

The appendix builds up in six stages:
\begin{enumerate}[leftmargin=1.5em]
  \item \emph{Algebra} (Sections~\ref{sec:theory-setup}--\ref{sec:theory-gradients}): we write out the last-layer gradients of CE and DCE and observe that they factor as outer products of a label-space residual and a backbone-feature vector. This reduces the gradient cross-term to a sum of products of two scalar quantities (Lemma~\ref{lem:factorization}).
  \item \emph{First-step geometry} (Sections~\ref{sec:theory-prop-first-step}--\ref{sec:theory-id-interference}): a short Taylor expansion yields the anchor-loss decrement after one step as $-\eta\|g_a\|_F^2 - \eta\langle g_a, g_e\rangle_F$ (Proposition~\ref{prop:first-step}). The sign and magnitude of the cross-term is then analyzed in two regimes: upper-bounded under a feature-orthogonality assumption (Corollary~\ref{cor:feature-orth}), and strictly negative under an ID-confidence assumption (Proposition~\ref{prop:id-interference}).
  \item \emph{PFN theory} (Section~\ref{sec:theory-nagler}): we use \citet[Theorems 5.4 \& 6.3]{nagler2023statistical} to argue that the feature-orthogonality assumption is a structural consequence of transformer attention in the \emph{far-OOD} limit. The key object is an exponentially tilted measure over the support set that collapses to the untilted marginal as $\|p_1 - p_0\|_{\mathrm{TV}} \to 1$.
  \item \emph{Main separation results} (Sections~\ref{sec:theory-tstep}--\ref{sec:theory-multiclass}): multi-step lift (Proposition~\ref{prop:T-step}), the Main Separation Theorem that composes all preceding pieces into $\bbE[\lambda_{\mathrm{OOD}}] < \bbE[\lambda_{\mathrm{ID}}]$ (Theorem~\ref{thm:main-separation}), a conformal-prediction argument establishing Type-I error control (Proposition~\ref{prop:conformal}), and a multiclass extension.
  \item \emph{Harm characterization} (Section~\ref{sec:theory-harm}): we extend Nagler's tilting from features to predictions, derive a density-ratio decomposition of the accuracy gap $\Delta$, and introduce a second signal $\hat\mu$ that estimates $\Delta$ directly (Proposition~\ref{prop:harm-theorem}). This bridges the gap between distributional-difference detection and predictive-harm detection.
  \item \emph{Summary of assumptions} (Section~\ref{sec:theory-assumptions}): We summarize our main assumptions for both the main separation theorem as well as the harm detection extension.
\end{enumerate}

\paragraph{Gradient interpretation.} Two gradients live in a large parameter space $\bbR^{C \times d}$. The anchor gradient $g_a$ points in the direction that, if walked along, most rapidly increases the probability assigned to the true labels on the anchor set. The disagreement gradient $g_e$ points in the direction that most rapidly \emph{reduces} the probability assigned to the (frozen) pseudo-labels on the eval set. Gradient descent moves in the direction $-(g_a + g_e)$. The question `does this step also improve the anchor loss?' therefore reduces to: does $-(g_a + g_e)$ have positive inner product with the direction $-g_a$ that purely improves anchor loss? This inner product is $\|g_a\|_F^2 + \langle g_a, g_e\rangle_F$, and its sign determines whether the disagreement gradient aligns with anchor descent or interferes with it.

\subsection{Setup and Notation}
\label{sec:theory-setup}

We document the formal setup in two pieces. We start with an important design choice that the subsequent analysis depends on. Then, we introduce the notation used throughout this appendix.

\paragraph{Why the analysis trains only the last layer.} Our analysis models the finetune as acting on the last linear (or small MLP) layer with every other parameter frozen; the deployed detector updates the whole decoder, and Appendix~\ref{sec:appendix-design} shows the two are empirically indistinguishable. This restriction is what makes our analysis tractable. A frozen backbone provides a fixed feature map $\phi$, and all gradient computations reduce to linear algebra in the last-layer parameter space. The per-sample gradient becomes a rank-one outer product of a label-space residual and a feature vector, which factorizes the gradient cross-term into a label $\times$ feature product.

\paragraph{Notation.} Let $\phi: \cX \times (\cX \times \cY)^n \to \bbR^d$ denote the PFN's backbone map, which takes a query $x$ and a support set $\cD^{\mathrm{sup}} = \{(X_i, Y_i)\}_{i=1}^n$ and returns a $d$-dimensional representation. We write $\phi(x) := \phi(x;\,\cD^{\mathrm{sup}})$ when the support is held fixed; all arguments in this appendix treat $\cD^{\mathrm{sup}}$ as a fixed conditioning set. Concretely, $\phi(x)$ is the transformer's penultimate activation at query $x$, with the support-set context incorporated through attention. The trainable parameter is the last-layer weight matrix $W \in \bbR^{C \times d}$, which absorbs the corresponding bias by the usual one-augmentation of $\phi$ (i.e., appending a constant $1$-coordinate). Predictions take the form
\begin{equation}
  q_W(y \mid x) = \mathrm{softmax}(W \phi(x))_y.
  \label{eq:decoder-softmax}
\end{equation}
All other PFN parameters are frozen at their pretrained values $\theta_0$, so $\phi$ does not depend on $W$. Write $W_0$ for the initial decoder (the decoder weights of the pretrained PFN) and $p_x := q_{W_0}(\cdot \mid x) \in \Delta^{C-1}$ for the initial predicted distribution at query $x$. The initial prediction $p_x$ is the PFN's usual in-context prediction before any finetuning.

Let $(\Xanc, \yanc) = \{(x_a^{(i)}, y_a^{(i)})\}_{i=1}^{n_a}$ be the labeled ID anchor set and $\Xeval = \{x_e^{(j)}\}_{j=1}^{n_e}$ the unlabeled eval batch. Pseudo-labels $\hat{y}^{(j)} := \argmax_c p_{x_e^{(j)}}(c)$ are frozen at initialization and do not change during finetuning (\emph{cf.} the pseudolabel step of Algorithm~\ref{alg:method}). Define the per-sample losses
\begin{align}
  \ell_{\mathrm{CE}}(W; x, y)     &= -\log q_W(y \mid x), \\
  \ell_{\mathrm{DCE}}(W; x, \hat{y}) &= -\frac{1}{C-1} \sum_{c \neq \hat{y}} \log q_W(c \mid x),
\end{align}
and the batch losses
\begin{equation}
  \cL_{\mathrm{anc}}(W) = \frac{1}{n_a} \sum_{i=1}^{n_a} \ell_{\mathrm{CE}}(W; x_a^{(i)}, y_a^{(i)}), \qquad \cL_{\mathrm{dis}}(W) = \frac{1}{n_e} \sum_{j=1}^{n_e} \ell_{\mathrm{DCE}}(W; x_e^{(j)}, \hat{y}^{(j)}).
\end{equation}
The finetuning objective is $\cL(W) = \cL_{\mathrm{anc}}(W) + \cL_{\mathrm{dis}}(W)$. The post-finetuning anchor loss, which is the detection statistic of interest, is
\begin{equation}
  \lambda := \cL_{\mathrm{anc}}(W_T), \qquad W_{t+1} = W_t - \eta\, \nabla_W \cL(W_t).
  \label{eq:detection-stat}
\end{equation}

\subsection{Last-Layer Gradient Geometry}
\label{sec:theory-gradients}

\paragraph{Gradient of CE.} We derive~\eqref{eq:grad-ce} explicitly so that the reader can see why the outer-product structure emerges. Write the pre-softmax logits as $z := W\phi(x) \in \bbR^C$, so $q_W(c \mid x) = e^{z_c} / \sum_{c'} e^{z_{c'}}$. The per-sample CE loss is $\ell_{\mathrm{CE}} = -\log q_W(y \mid x) = -z_y + \log \sum_{c'} e^{z_{c'}}$. The gradient with respect to $z$ is the standard softmax residual
\begin{equation}
  \frac{\partial \ell_{\mathrm{CE}}}{\partial z_c} = q_W(c \mid x) - \indic\{c = y\} = \bigl(q_W(\cdot \mid x) - e_y\bigr)_c.
  \label{eq:grad-z-ce}
\end{equation}
To push this back to $W$, use the chain rule: $\partial z_c / \partial W_{c', d'} = \phi_{d'}(x) \indic\{c = c'\}$, so
\begin{equation*}
  \frac{\partial \ell_{\mathrm{CE}}}{\partial W_{c, d}} = \bigl(q_W(c \mid x) - \indic\{c = y\}\bigr)\, \phi_d(x).
\end{equation*}
Reading the right-hand side as the $(c, d)$ entry of an outer product between the vector $q_W(\cdot \mid x) - e_y \in \bbR^C$ and the vector $\phi(x) \in \bbR^d$ yields
\begin{equation}
  \nabla_W \ell_{\mathrm{CE}}(W; x, y) = \bigl(q_W(\cdot \mid x) - e_y\bigr)\, \phi(x)^\top.
  \label{eq:grad-ce}
\end{equation}
This is a rank-one $C \times d$ matrix: the first factor lives in label space and records the per-class prediction error; the second factor lives in feature space and records `which direction in $W$ to push'.

\paragraph{Gradient of DCE.} The disagreement cross-entropy $\ell_{\mathrm{DCE}}(W; x, \hat{y}) = -\tfrac{1}{C-1}\sum_{c \neq \hat{y}} \log q_W(c \mid x)$ is itself a cross-entropy, against the uniform-on-non-pseudolabel target $u_{\hat{y}} \in \Delta^{C-1}$ defined by $(u_{\hat{y}})_c = \indic\{c \neq \hat{y}\} / (C-1)$. Indeed,
\begin{equation*}
  \mathrm{CE}\bigl(q_W(\cdot \mid x),\, u_{\hat{y}}\bigr)
  = -\sum_c (u_{\hat{y}})_c \log q_W(c \mid x)
  = -\tfrac{1}{C-1} \sum_{c \neq \hat{y}} \log q_W(c \mid x)
  = \ell_{\mathrm{DCE}}.
\end{equation*}
Applying the same chain-rule calculation as for CE, with target $u_{\hat{y}}$ replacing $e_y$,
\begin{equation}
  \nabla_W \ell_{\mathrm{DCE}}(W; x, \hat{y}) = \bigl(q_W(\cdot \mid x) - u_{\hat{y}}\bigr)\, \phi(x)^\top.
  \label{eq:grad-dce}
\end{equation}
This gradient is again rank-one and shares the feature factor $\phi(x)$ with the CE gradient in~\eqref{eq:grad-ce}.

\paragraph{Batch gradients.} Both gradients are rank-one matrices in $\bbR^{C \times d}$. Evaluating at initialization, write
\begin{equation}
  r_a^{(i)} := p_{x_a^{(i)}} - e_{y_a^{(i)}}, \qquad r_e^{(j)} := p_{x_e^{(j)}} - u_{\hat{y}^{(j)}}
  \label{eq:residuals}
\end{equation}
for the anchor- and eval-residual vectors in $\bbR^C$. $r_a^{(i)}$ measures the prediction error on anchor point $i$ in label space, while $r_e^{(j)}$ is the corresponding direction in which the disagreement loss pushes the prediction away from the pseudolabel on eval point $j$. The initial batch gradients are the averaged rank-one matrices
\begin{align}
  g_a &:= \nabla_W \cL_{\mathrm{anc}}(W_0) = \frac{1}{n_a} \sum_{i=1}^{n_a} r_a^{(i)}\, \phi(x_a^{(i)})^\top, \label{eq:ga} \\
  g_e &:= \nabla_W \cL_{\mathrm{dis}}(W_0) = \frac{1}{n_e} \sum_{j=1}^{n_e} r_e^{(j)}\, \phi(x_e^{(j)})^\top. \label{eq:ge}
\end{align}

\paragraph{Why we care about $\langle g_a, g_e\rangle_F$.} The joint update moves $W$ in the direction $-(g_a + g_e)$; whether this also descends the anchor loss depends on the alignment between $-(g_a + g_e)$ and the pure anchor-descent direction $-g_a$. Specifically, the first-order change in anchor loss is $\eta \langle g_a, g_a + g_e\rangle_F = \eta(\|g_a\|_F^2 + \langle g_a, g_e\rangle_F)$, a sum of a non-negative self-term and the gradient cross-term $\langle g_a, g_e\rangle_F$ whose sign determines whether the disagreement gradient aligns with (\emph{sign} $\ge 0$) or interferes with (\emph{sign} $<0$) anchor descent. Proposition~\ref{prop:first-step} makes this Taylor-level statement precise. Before that, it is useful to see that this inner product has a simple structure: it decomposes into a weighted sum over anchor-eval \emph{pairs}, where each pair contributes a product of a label-space scalar and a feature-space scalar. This factorization is what makes the rest of the analysis tractable.

\begin{lemma}[Gradient cross-term factorization]
\label{lem:factorization}
The Frobenius inner product of $g_a$ and $g_e$ factorizes as
\begin{equation}
  \langle g_a,\, g_e\rangle_F = \frac{1}{n_a n_e} \sum_{i=1}^{n_a} \sum_{j=1}^{n_e} C_{ij}\, K_{ij},
  \label{eq:factorization}
\end{equation}
where $C_{ij} := \langle r_a^{(i)}, r_e^{(j)}\rangle \in [-2, 2]$ is a label-space residual inner product and $K_{ij} := \langle \phi(x_a^{(i)}), \phi(x_e^{(j)})\rangle$ is a feature-space kernel.
\end{lemma}

\begin{proof}
We first establish the general outer-product identity. Let $A = u\, v^\top \in \bbR^{C \times d}$ and $B = u'\, v'^\top \in \bbR^{C \times d}$ be rank-one matrices with $u, u' \in \bbR^C$ and $v, v' \in \bbR^d$. The Frobenius inner product is
\begin{align*}
  \langle A, B\rangle_F
  &= \sum_{c=1}^C \sum_{d'=1}^d A_{c, d'}\, B_{c, d'}
  = \sum_{c, d'} (u_c v_{d'})(u'_c v'_{d'}) \\
  &= \Bigl(\sum_c u_c\, u'_c\Bigr)\, \Bigl(\sum_{d'} v_{d'}\, v'_{d'}\Bigr)
  = \langle u, u'\rangle\, \langle v, v'\rangle.
\end{align*}
In words: the Frobenius inner product of two rank-one matrices splits into the product of the two `left' inner products and the two `right' inner products. This is why decoder-only updates make the analysis tractable.

Now expand the cross-term using bilinearity:
\begin{align*}
  \langle g_a, g_e\rangle_F
  &= \Bigl\langle \tfrac{1}{n_a}\sum_i r_a^{(i)} \phi(x_a^{(i)})^\top, \tfrac{1}{n_e}\sum_j r_e^{(j)} \phi(x_e^{(j)})^\top\Bigr\rangle_F \\
  &= \tfrac{1}{n_a n_e} \sum_{i, j} \bigl\langle r_a^{(i)} \phi(x_a^{(i)})^\top, r_e^{(j)} \phi(x_e^{(j)})^\top\bigr\rangle_F \\
  &= \tfrac{1}{n_a n_e} \sum_{i, j} \langle r_a^{(i)}, r_e^{(j)}\rangle\, \langle \phi(x_a^{(i)}), \phi(x_e^{(j)})\rangle
  = \tfrac{1}{n_a n_e} \sum_{i, j} C_{ij}\, K_{ij},
\end{align*}
where the third equality applies the rank-one identity to each pair $(i, j)$.

For the bound $|C_{ij}| \le 2$: by Cauchy--Schwarz, $|C_{ij}| \le \|r_a^{(i)}\|_2\, \|r_e^{(j)}\|_2$. Each residual is a difference of two distributions in $\Delta^{C-1}$, so $\|r_a^{(i)}\|_2 = \|p_{x_a} - e_{y_a}\|_2 \le \sqrt{2}$ (the maximum squared $\ell_2$ distance between two probability vectors on $C$ classes is $2$, attained when both are point masses on different classes). Hence $|C_{ij}| \le \sqrt{2} \cdot \sqrt{2} = 2$.
\end{proof}

Equation~\eqref{eq:factorization} is the basis of the analysis. The entire gradient interaction between the anchor and eval batches collapses onto the elementwise product of a label-space matrix $C \in \bbR^{n_a \times n_e}$ and a feature-space Gram matrix $K \in \bbR^{n_a \times n_e}$. The label part $C_{ij}$ is bounded, easy to interpret, and typically negative under ID (we will see this in Proposition~\ref{prop:id-interference}). The feature part $K_{ij}$ is the part that `knows' whether the eval batch is OOD: as we will show in Section~\ref{sec:theory-nagler}, OOD queries produce feature vectors that collapse to a single input-independent direction, making $K_{ij}$ small under OOD.

\subsection{First-Step Anchor Descent}
\label{sec:theory-prop-first-step}

\paragraph{Why a first-step analysis is meaningful.} Algorithm~\ref{alg:method} takes $T=50$ gradient steps per round at a learning rate selected from $\{10^{-2}, 10^{-3}, 10^{-4}\}$ (default $10^{-3}$). One might worry that the first step tells us nothing about the cumulative effect of fifty. However, there are two insights that make the first-step analysis informative. First, in the small-step regime (Assumption~\ref{assn:small-step}) the cumulative anchor-loss decrement after $T$ steps is a linear combination of first-step quantities up to $O(T^2\eta^2)$ corrections (Proposition~\ref{prop:T-step}). Second, the sign of the first-step cross-term is a discrete feature that typically persists along the trajectory when the backbone is frozen, because the feature-space kernel $K_{ij}$ is the same at every step. This means that the sign analysis at step one carries to step $T$ under mild regularity conditions.

\begin{proposition}[First-step anchor descent]
\label{prop:first-step}
Let $W_1 = W_0 - \eta (g_a + g_e)$ be the update after one gradient step on the joint loss $\cL$. Then
\begin{equation}
  \cL_{\mathrm{anc}}(W_1) - \cL_{\mathrm{anc}}(W_0) = -\eta\, \|g_a\|_F^2 - \eta\, \langle g_a,\, g_e\rangle_F + R(\eta),
  \label{eq:first-step}
\end{equation}
with remainder $|R(\eta)| \le \tfrac{1}{2} L_a \eta^2 \|g_a + g_e\|_F^2$, where $L_a$ is the Lipschitz constant of $\nabla \cL_{\mathrm{anc}}$ along the segment $[W_0, W_1]$.
\end{proposition}

\begin{proof}
The loss $\cL_{\mathrm{anc}}: \bbR^{C\times d} \to \bbR$ is twice continuously differentiable, being a composition of the affine map $W \mapsto W\phi(x)$ with the softmax-cross-entropy function, both of which are smooth. On the bounded segment $[W_0, W_1] = \{W_0 + t(W_1 - W_0) : t \in [0, 1]\}$, $\nabla \cL_{\mathrm{anc}}$ is $L_a$-Lipschitz for some finite $L_a$; we bound $L_a$ explicitly in the proof of Proposition~\ref{prop:T-step}.

\emph{Step 1 (Taylor expansion with integral remainder).} For any twice continuously differentiable $f: \bbR^{C\times d} \to \bbR$ and any $W_0, W_1$,
\begin{align*}
  f(W_1) =& f(W_0) \,+\, \langle \nabla f(W_0),\, W_1 - W_0\rangle_F \\
  &+ \int_0^1 (1-t)\, \bigl\langle W_1 - W_0,\, \nabla^2 f(W_0 + t(W_1 - W_0))[W_1 - W_0]\bigr\rangle_F \, dt.
\end{align*}
Applying this to $f = \cL_{\mathrm{anc}}$ and denoting the integral as $R(\eta)$ gives
\begin{equation*}
  \cL_{\mathrm{anc}}(W_1) - \cL_{\mathrm{anc}}(W_0) = \langle \nabla \cL_{\mathrm{anc}}(W_0),\, W_1 - W_0\rangle_F + R(\eta).
\end{equation*}

\emph{Step 2 (compute the linear term).} By definition $\nabla \cL_{\mathrm{anc}}(W_0) = g_a$, and the update direction is $W_1 - W_0 = -\eta(g_a + g_e)$. Hence
\begin{align*}
  \langle \nabla \cL_{\mathrm{anc}}(W_0),\, W_1 - W_0\rangle_F
  &= \langle g_a,\, -\eta(g_a + g_e)\rangle_F \\
  &= -\eta \langle g_a, g_a\rangle_F - \eta \langle g_a, g_e\rangle_F \\
  &= -\eta \|g_a\|_F^2 - \eta \langle g_a, g_e\rangle_F,
\end{align*}
using bilinearity of the Frobenius inner product.

\emph{Step 3 (bound the remainder).} By $L_a$-Lipschitzness of $\nabla \cL_{\mathrm{anc}}$ on $[W_0, W_1]$, the Hessian has operator norm $\le L_a$ on the segment. Plugging this into the integral remainder,
\begin{align*}
  |R(\eta)|
  &\le \int_0^1 (1-t)\, L_a\, \|W_1 - W_0\|_F^2\, dt \\
  &= L_a\, \eta^2 \|g_a + g_e\|_F^2 \int_0^1 (1-t)\, dt
  = \tfrac{1}{2} L_a\, \eta^2 \|g_a + g_e\|_F^2.
\end{align*}
Combining steps 1--3 gives~\eqref{eq:first-step}.
\end{proof}

\begin{figure}[t]
\centering
\begin{subfigure}[b]{0.29\textwidth}
\centering
\begin{tikzpicture}[scale=0.85, >={Stealth[scale=1.1]}]
  \draw[thin, gray!25] (-2.3,-1.1) rectangle (2.3,1.8);
  \begin{scope}[shift={(-1.55,0)}]
    \draw[->, thick, blue!60!black, line width=1.2pt] (0,0) -- (1.6,0);
    \node[above, blue!60!black, font=\small] at (1.7, 0.02) {$g_a$};
    \draw[->, thick, orange!80!black, line width=1.2pt] (0,0) -- (1.0,1.3);
    \node[left, orange!80!black, font=\small] at (0.8,1.3) {$g_e$};
    \draw[->, gray!60!black, dashed, line width=0.9pt] (0,0) -- (2.6,1.3);
    \node[above, gray!70!black, font=\scriptsize] at (2.6,1.3) {$g_a+g_e$};
    \draw[decorate, decoration={brace, amplitude=5pt}, black, line width=0.8pt] (2.6,-0.25) -- (0,-0.25);
    \node[below, black, font=\scriptsize] at (1.3,-0.55) {descent rate};
  \end{scope}
\end{tikzpicture}
\caption{\footnotesize \textbf{Cooperation.} $\langle g_a, g_e\rangle_F \ge 0$. Descent rate $\ge \eta\|g_a\|_F^2$ (OOD regime).}
\end{subfigure}
\hfill
\begin{subfigure}[b]{0.29\textwidth}
\centering
\begin{tikzpicture}[scale=0.85, >={Stealth[scale=1.1]}]
  \draw[thin, gray!25] (-2.3,-1.1) rectangle (2.3,1.8);
  \begin{scope}[shift={(-0.35,0)}]
    \draw[->, thick, blue!60!black, line width=1.2pt] (0,0) -- (1.6,0);
    \node[above, blue!60!black, font=\small] at (0.8,0.02) {$g_a$};
    \draw[->, thick, orange!80!black, line width=1.2pt] (0,0) -- (-0.7,1.3);
    \node[above, orange!80!black, font=\small] at (-0.7,1.3) {$g_e$};
    \draw[->, gray!60!black, dashed, line width=0.9pt] (0,0) -- (0.9,1.3);
    \node[above, gray!70!black, font=\scriptsize] at (0.9,1.3) {$g_a+g_e$};
    \draw[decorate, decoration={brace, amplitude=5pt}, black, line width=0.8pt] (0.9,-0.25) -- (0,-0.25);
    \node[below, black, font=\scriptsize] at (0.45,-0.55) {damped};
  \end{scope}
\end{tikzpicture}
\caption{\footnotesize \textbf{Damped.} $\langle g_a, g_e\rangle_F \in (-\|g_a\|_F^2, 0)$. Descent rate $< \eta\|g_a\|_F^2$ (typical ID regime).}
\end{subfigure}
\hfill
\begin{subfigure}[b]{0.29\textwidth}
\centering
\begin{tikzpicture}[scale=0.85, >={Stealth[scale=1.1]}]
  \draw[thin, gray!25] (-2.3,-1.1) rectangle (2.3,1.8);
  \begin{scope}[shift={(0.45,0.3)}]
    \draw[->, thick, blue!60!black, line width=1.2pt] (0,0) -- (1.6,0);
    \node[above, blue!60!black, font=\small] at (0.8,0.02) {$g_a$};
    \draw[->, thick, orange!80!black, line width=1.2pt] (0,0) -- (-2.2,0.7);
    \node[above, orange!80!black, font=\small] at (-2.2,0.7) {$g_e$};
    \draw[->, gray!60!black, dashed, line width=0.9pt] (0,0) -- (-0.6,0.7);
    \node[above, gray!70!black, font=\scriptsize] at (-0.6,0.7) {$g_a+g_e$};
    \draw[decorate, decoration={brace, amplitude=5pt}, black, line width=0.8pt] (0,-0.25) -- (-0.6,-0.25);
    \node[below, black, font=\scriptsize] at (-0.3,-0.55) {stall};
  \end{scope}
\end{tikzpicture}
\caption{\footnotesize \textbf{Stall.} $\langle g_a, g_e\rangle_F \le -\|g_a\|_F^2$. Descent rate $\le 0$ (extreme ID regime).}
\end{subfigure}
\caption{\textbf{The three regimes of Proposition~\ref{prop:first-step}.} The first-order anchor-loss decrement after one joint-loss step equals $\eta$ times the projection of $g_a + g_e$ (dashed) onto the direction of $g_a$ (blue). When $g_e$ (orange) makes an acute angle with $g_a$ \textbf{(left)}, descent is unimpeded or better; this is the OOD regime (Corollary~\ref{cor:feature-orth}). When $g_e$ makes an obtuse but bounded angle \textbf{(middle)}, descent is damped but still positive (the typical ID regime, Proposition~\ref{prop:id-interference}). When $g_e$ is strongly anti-aligned with $g_a$ \textbf{(right)}, the projection turns negative and anchor loss does not decrease. The detector's signal comes from the ID regime sitting in the middle panel and the OOD regime sitting approximately on the left.}
\label{fig:three-regimes}
\end{figure}
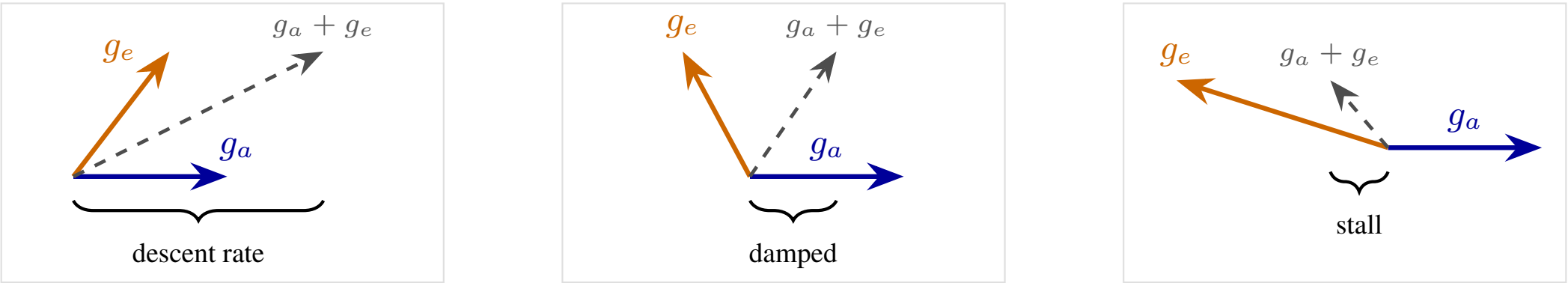

\paragraph{Interpretation: the cross-term controls the first-order descent rate.} The significance of Proposition~\ref{prop:first-step} is that the first-order anchor descent rate is \emph{entirely} controlled by the sign and magnitude of the cross-term $\langle g_a, g_e\rangle_F$. Fix $\eta$ small enough such that the remainder $R(\eta) = O(\eta^2)$ is negligible compared to the first-order terms of size $O(\eta)$. Then
\begin{equation*}
  \cL_{\mathrm{anc}}(W_1) - \cL_{\mathrm{anc}}(W_0) \approx -\eta\bigl(\|g_a\|_F^2 + \langle g_a, g_e\rangle_F\bigr).
\end{equation*}
Since $\|g_a\|_F^2 \ge 0$, the three qualitatively distinct regimes are (see Figure~\ref{fig:three-regimes}):
\begin{enumerate}[leftmargin=1.5em]
  \item \textbf{Alignment or orthogonality} $\langle g_a, g_e\rangle_F \ge 0$: anchor loss decreases at rate $\ge \eta \|g_a\|_F^2$. The disagree gradient either actively helps anchor descent (positive inner product) or is geometrically unrelated to it (zero inner product). \emph{This is the regime we will argue for OOD eval batches.}
  \item \textbf{Damped descent} $\langle g_a, g_e\rangle_F \in (-\|g_a\|_F^2,\, 0)$: anchor loss still decreases, but at a rate strictly smaller than $\eta\|g_a\|_F^2$. The disagree gradient partially cancels the anchor gradient. \emph{This is the typical ID regime.}
  \item \textbf{Stall or increase} $\langle g_a, g_e\rangle_F \le -\|g_a\|_F^2$: anchor loss stalls or increases. The disagree gradient more than cancels the anchor gradient. \emph{This is an extreme ID regime in which anchor and eval are particularly entangled.}
\end{enumerate}

The informal claim that the disagreement gradient `interferes with anchor agreement' in the ID regime but not in the OOD regime translates to this cross-term being negative under ID and near-zero under OOD. The next two subsections make each direction rigorous: Corollary~\ref{cor:feature-orth} upper-bounds $|\langle g_a, g_e\rangle_F|$ under an OOD feature-orthogonality assumption, while Proposition~\ref{prop:id-interference} lower-bounds it by a negative quantity under an ID confidence assumption.

\subsection{Feature-Orthogonality Corollary}
\label{sec:theory-feature-orth}

Corollary~\ref{cor:feature-orth} below gives an upper bound on $|\langle g_a, g_e\rangle_F|$ when anchor and eval features are approximately orthogonal. This happens whenever $\cos(\phi(x_a), \phi(x_e))$ is small.

\begin{assumption}[Feature orthogonality]
\label{assn:feat-orth}
There exists $\epsilon \ge 0$ such that, for every anchor-eval pair $(i, j)$,
\begin{equation}
  \bigl|\langle \phi(x_a^{(i)}),\, \phi(x_e^{(j)})\rangle\bigr| \leq \epsilon\, \|\phi(x_a^{(i)})\|\, \|\phi(x_e^{(j)})\|.
  \label{eq:feat-orth}
\end{equation}
\end{assumption}

The parameter $\epsilon \in [0, 1]$ is the worst-case cosine between anchor and eval features. When $\epsilon = 0$ the two feature sets lie in orthogonal subspaces; when $\epsilon = 1$ they are identical up to scale.

\begin{corollary}[Orthogonal features imply unimpeded anchor descent]
\label{cor:feature-orth}
Assume $\|\phi(x)\| \le B_\phi$ for all $x$ in $\Xanc \cup \Xeval$ and set $B_C := \max_{i,j} |C_{ij}| \le 2$ (from Lemma~\ref{lem:factorization}). Under Assumption~\ref{assn:feat-orth},
\begin{equation}
  \bigl|\langle g_a,\, g_e\rangle_F\bigr| \leq \epsilon\, B_\phi^2\, B_C.
  \label{eq:cross-term-bound}
\end{equation}
Consequently, when $\epsilon = o\bigl(\|g_a\|_F^2 / (B_\phi^2 B_C)\bigr)$, the first-step anchor descent in~\eqref{eq:first-step} reduces to $-\eta \|g_a\|_F^2 (1 + o(1))$, i.e., the disagreement gradient leaves anchor descent unimpeded at first order in the step size $\eta$.
\end{corollary}

\begin{proof}
Start from the factorization~\eqref{eq:factorization}. For each pair $(i, j)$, bound the summand by the product of term-wise bounds:
\begin{align*}
  |C_{ij} K_{ij}|
  &= |C_{ij}|\, \bigl|\langle \phi(x_a^{(i)}), \phi(x_e^{(j)})\rangle\bigr| \\
  &\le B_C \cdot \epsilon\, \|\phi(x_a^{(i)})\|\, \|\phi(x_e^{(j)})\| & \text{(Assumption~\ref{assn:feat-orth}; definition of $B_C$)} \\
  &\le B_C \cdot \epsilon\, B_\phi^2 & \text{(boundedness of $\phi$).}
\end{align*}
Averaging over the $n_a n_e$ pairs in~\eqref{eq:factorization} (the triangle inequality applied to the sum),
\begin{equation*}
  \bigl|\langle g_a, g_e\rangle_F\bigr|
  = \Bigl|\tfrac{1}{n_a n_e}\sum_{i, j} C_{ij} K_{ij}\Bigr|
  \le \tfrac{1}{n_a n_e}\sum_{i, j} |C_{ij} K_{ij}|
  \le \tfrac{1}{n_a n_e} \cdot n_a n_e \cdot \epsilon B_\phi^2 B_C
  = \epsilon B_\phi^2 B_C.
\end{equation*}
For the consequence, plug this bound into~\eqref{eq:first-step}:
\begin{equation*}
  \cL_{\mathrm{anc}}(W_1) - \cL_{\mathrm{anc}}(W_0) = -\eta\|g_a\|_F^2\Bigl(1 + \tfrac{\langle g_a, g_e\rangle_F}{\|g_a\|_F^2}\Bigr) + O(\eta^2),
\end{equation*}
and observe that the ratio $\langle g_a, g_e\rangle_F / \|g_a\|_F^2$ is $o(1)$ exactly when $\epsilon = o(\|g_a\|_F^2 / (B_\phi^2 B_C))$.
\end{proof}

\paragraph{Intuitive reading.} When anchor and eval features lie in approximately orthogonal subspaces of $\bbR^d$, moving the decoder in the direction $-g_e$ (which is a sum of rank-one outer products with eval features) has little effect on the decoder's behavior on anchor features. The disagreement gradient is, to first order, \emph{invisible} to the anchor gradient. Therefore, the joint update $-(g_a + g_e)$ acts on the anchor loss as if only $-g_a$ had been applied, and anchor loss descends at the usual rate $\eta\|g_a\|_F^2$.

\subsection{ID Interference}
\label{sec:theory-id-interference}

The OOD corollary of the previous subsection showed that $\langle g_a, g_e\rangle_F$ is \emph{small} when features are orthogonal. But to get a detector we need separation: the cross-term must not just be zero but \emph{negative} under ID, so anchor descent is visibly damped. This subsection establishes that separation.

\paragraph{Why this is not obvious.} A casual reader might expect that any gradient that is not orthogonal to $g_a$ contributes random noise. However, the claim here is stronger: under the mild ID conditions listed below, the sign of $\langle g_a, g_e\rangle_F$ is \emph{systematically} negative. This asymmetry between the OOD regime (cross-term $\approx 0$) and the ID regime (cross-term $< 0$) is the source of the detector's signal: $\bbE[\lambda_{\mathrm{OOD}}] - \bbE[\lambda_{\mathrm{ID}}]$ is bounded above by $-\eta\,\bbE_{p_0 \otimes p_0}[\langle g_a, g_e\rangle_F] < 0$ at first order, where $p_0 \otimes p_0$ denotes the product distribution under which the anchor pair $(x_a, y_a)$ and the eval pair $(x_e, \hat{y})$ are drawn independently, each from $p_0$.

\begin{assumption}[ID confidence, correctness, and class symmetry]
\label{assn:id-confidence}
Recall from Section~\ref{sec:theory-setup} that $p_x := q_{W_0}(\cdot \mid x) \in \Delta^{C-1}$ is the initial PFN prediction at query $x$, and the pseudolabel function is $\hat{y}(x) := \argmax_c p_x(c)$. Under sampling $(x_a, y_a), (x_e, \hat{y}) \sim p_0 \otimes p_0$, and for a constant $\tau < 1/(2(C-1))$:
\begin{enumerate}[leftmargin=3.5em]
  \item[(i)] The initial model is non-trivially correct on the anchor: $\bbE_{p_0}[p_{x_a}(y_a)] \ge 1/2$.
  \item[(ii)] The initial model is non-trivially confident on eval: $\bbE_{p_0}[p_{x_e}(\hat{y}(x_e))] \ge 1 - \tau$.
  \item[(iii)] Backbone features are $p_0$-centered and non-degenerate: $\bbE_{p_0}[\phi(x)] = 0$ and $\bbE_{p_0}[\|\phi(x)\|^2] > 0$.
  \item[(iv)] (Class symmetry) The class-conditional feature means $\mu_c := \bbE_{p_0}[\phi(x) \mid y(x) = c]$ span $C$ distinct directions, and the predictor satisfies $p_x(c) - 1/C \propto \langle \phi(x), \mu_c \rangle$ up to first order.
  \item[(v)] (Accuracy) The initial classifier is accurate on ID: $\bbP_{p_0}(\hat y(x) = y(x)) \ge 1 - \tau$.
\end{enumerate}
\end{assumption}

Condition (iii) can be arranged by subtracting the $p_0$-mean from $\phi$; we revisit this when connecting to Nagler's limiting feature in Section~\ref{sec:theory-nagler}. Condition (iv) is the class-symmetry assumption that makes the label-disagreement case reverse the sign of the feature covariance; it is standard in linearized-network analyses. We state the proposition for general $C$ but prove it first in the illustrative binary case, then give the multiclass extension.

\begin{proposition}[ID interference]
\label{prop:id-interference}
Under Assumption~\ref{assn:id-confidence},
\begin{equation}
  \bbE_{p_0 \otimes p_0}\bigl[C_{ij}\, K_{ij}\bigr] < 0,
\end{equation}
so in expectation $\langle g_a, g_e\rangle_F < 0$, and Proposition~\ref{prop:first-step} gives damped anchor descent.
\end{proposition}

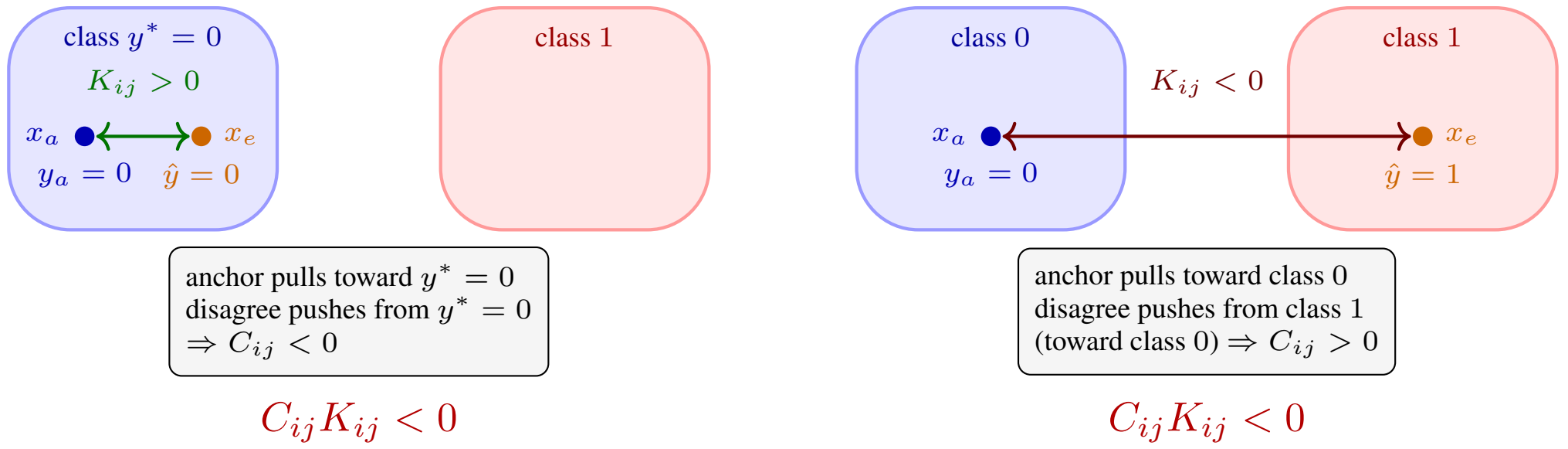
\begin{figure}[t]
\centering
\begin{subfigure}[b]{0.48\textwidth}
\centering
\begin{tikzpicture}[scale=1.0]
  \draw[fill=blue!10, draw=blue!40, thick, rounded corners=15pt] (-3.0, 0.5) rectangle (-0.7, 2.4);
  \node[blue!60!black, font=\scriptsize] at (-1.85, 2.15) {class $y^* = 0$};
  \draw[fill=red!10, draw=red!40, thick, rounded corners=15pt] (0.7, 0.5) rectangle (3.0, 2.4);
  \node[red!60!black, font=\scriptsize] at (1.85, 2.15) {class $1$};
  \filldraw[blue!70!black] (-2.35, 1.3) circle (2.2pt);
  \node[left=2pt, blue!70!black, font=\scriptsize] at (-2.35, 1.3) {$x_a$};
  \node[below=3pt, blue!70!black, font=\scriptsize] at (-2.35, 1.3) {$y_a = 0$};
  \filldraw[orange!80!black] (-1.35, 1.3) circle (2.2pt);
  \node[right=2pt, orange!80!black, font=\scriptsize] at (-1.35, 1.3) {$x_e$};
  \node[below=3pt, orange!80!black, font=\scriptsize] at (-1.35, 1.3) {$\hat y = 0$};
  \draw[<->, green!45!black, thick] (-2.25, 1.3) -- (-1.45, 1.3);
  \node[green!45!black, font=\scriptsize] at (-1.85, 1.75) {$K_{ij} > 0$};
  \node[draw, rounded corners=3pt, fill=gray!8, font=\scriptsize, align=left, inner sep=4pt] at (0.0, -0.2) {anchor pulls toward $y^*=0$\\ disagree pushes from $y^*=0$\\ $\Rightarrow C_{ij} < 0$};
  \node[font=\bfseries, red!70!black] at (0, -1.15) {$C_{ij} K_{ij} < 0$};
\end{tikzpicture}
\caption{\textbf{Label agreement} ($y_a = \hat y$). Anchor and eval lie in the same class region, giving $K_{ij} > 0$. The anchor gradient pulls toward that class while the disagreement gradient pushes away, giving $C_{ij} < 0$. The product $C_{ij} K_{ij}$ is therefore negative.}
\end{subfigure}
\hfill
\begin{subfigure}[b]{0.48\textwidth}
\centering
\begin{tikzpicture}[scale=1.0]
  \draw[fill=blue!10, draw=blue!40, thick, rounded corners=15pt] (-3.0, 0.5) rectangle (-0.7, 2.4);
  \node[blue!60!black, font=\scriptsize] at (-1.85, 2.15) {class $0$};
  \draw[fill=red!10, draw=red!40, thick, rounded corners=15pt] (0.7, 0.5) rectangle (3.0, 2.4);
  \node[red!60!black, font=\scriptsize] at (1.85, 2.15) {class $1$};
  \filldraw[blue!70!black] (-1.85, 1.3) circle (2.2pt);
  \node[left=2pt, blue!70!black, font=\scriptsize] at (-1.85, 1.3) {$x_a$};
  \node[below=3pt, blue!70!black, font=\scriptsize] at (-1.85, 1.3) {$y_a = 0$};
  \filldraw[orange!80!black] (1.85, 1.3) circle (2.2pt);
  \node[right=2pt, orange!80!black, font=\scriptsize] at (1.85, 1.3) {$x_e$};
  \node[below=3pt, orange!80!black, font=\scriptsize] at (1.85, 1.3) {$\hat y = 1$};
  \draw[<->, red!45!black, thick] (-1.75, 1.3) -- (1.75, 1.3);
  \node[red!45!black, font=\scriptsize] at (0, 1.75) {$K_{ij} < 0$};
  \node[draw, rounded corners=3pt, fill=gray!8, font=\scriptsize, align=left, inner sep=4pt] at (0.0, -0.2) {anchor pulls toward class $0$\\ disagree pushes from class $1$\\ (toward class $0$) $\Rightarrow C_{ij} > 0$};
  \node[font=\bfseries, red!70!black] at (0, -1.15) {$C_{ij} K_{ij} < 0$};
\end{tikzpicture}
\caption{\textbf{Label disagreement} ($y_a \neq \hat y$). Anchor and eval lie in different class regions, giving $K_{ij} < 0$. The anchor gradient and the disagreement gradient now both pull toward the same class, giving $C_{ij} > 0$. The product $C_{ij} K_{ij}$ is again negative.}
\end{subfigure}
\caption{\textbf{The ``double negative'' mechanism of Proposition~\ref{prop:id-interference}.} In both possible label configurations, the product $C_{ij} K_{ij}$ ends up negative: in the first case because $K > 0$ multiplies $C < 0$, in the second case because $K < 0$ multiplies $C > 0$. Since these are the only two cases, the expectation $\bbE[C_{ij} K_{ij}]$ is negative regardless of the label marginal. This is the mechanism that forces ID anchor descent into the ``damped'' regime of Figure~\ref{fig:three-regimes} (middle panel).}
\label{fig:double-negative}
\end{figure}

\begin{proof}
We prove the binary ($C = 2$) case in full and then indicate the multiclass extension.

\emph{Step 1 (binary simplification of residuals).} When $C = 2$ and the pseudolabel is $\hat{y}$, the uniform-on-non-pseudolabel distribution is $u_{\hat{y}} = e_{1-\hat{y}}$ (a one-hot on the flipped class). Hence $r_e^{(j)} = p_{x_e^{(j)}} - e_{1 - \hat{y}^{(j)}}$, which is the same as the CE residual against the flipped label. For the anchor, $r_a^{(i)} = p_{x_a^{(i)}} - e_{y_a^{(i)}}$ as before.

\emph{Step 2 (binary residual inner product).} Write $p_x = (1 - p_x^+, p_x^+)^\top$ with $p_x^+ := p_x(1)$, and identify $p_x^0 = 1 - p_x^+$, $p_x^1 = p_x^+$. A direct calculation gives
\begin{equation*}
  p_x - e_0 = (-p_x^+,\, p_x^+)^\top = -p_x^1\, (1, -1)^\top, \qquad p_x - e_1 = (1 - p_x^+,\, -(1 - p_x^+))^\top = p_x^0\, (1, -1)^\top.
\end{equation*}
Both residuals are scalar multiples of $e := (1, -1)^\top$. We can write this uniformly as
\begin{equation*}
  p_x - e_c = (-1)^{c+1} (1 - p_x^c)\, e.
\end{equation*}
Define $\alpha_x^c := (-1)^{c+1} (1 - p_x^c)$; note $|\alpha_x^c| = 1 - p_x^c$, and the sign depends on $c$. Then $r_a = \alpha_{x_a}^{y_a}\, e$ and $r_e = p_{x_e} - u_{\hat y} = p_{x_e} - e_{1-\hat y} = \alpha_{x_e}^{1-\hat y}\, e$. Hence
\begin{equation*}
  \langle r_a, r_e\rangle = \alpha_{x_a}^{y_a}\, \alpha_{x_e}^{1 - \hat{y}}\, \langle e, e\rangle
  = 2\, \alpha_{x_a}^{y_a}\, \alpha_{x_e}^{1 - \hat{y}}.
\end{equation*}

\emph{Step 3 (sign of each factor under Assumption~\ref{assn:id-confidence}).} Condition on $y_a$ and $\hat{y}(x_e)$. By the definition of $\alpha$:
\begin{itemize}[leftmargin=1.5em]
  \item $\alpha_{x_a}^{y_a} = (-1)^{y_a + 1}\,(1 - p_{x_a}^{y_a})$. Since $p_{x_a}^{y_a} \le 1$, the magnitude is $1 - p_{x_a}^{y_a} \ge 0$. The \emph{sign} is $(-1)^{y_a+1}$.
  \item $\alpha_{x_e}^{1 - \hat{y}} = (-1)^{(1-\hat{y})+1}\,(1 - p_{x_e}^{1 - \hat{y}}) = (-1)^{\hat{y}}\, p_{x_e}^{\hat{y}}$ (using $1 - p_{x_e}^{1-\hat{y}} = p_{x_e}^{\hat{y}}$). The magnitude is $p_{x_e}^{\hat{y}} = p_{x_e}(\hat{y}(x_e))$, which by Assumption~\ref{assn:id-confidence}(ii) is $\ge 1 - \tau$. The \emph{sign} is $(-1)^{\hat{y}}$.
\end{itemize}
Multiplying the signs:
\begin{equation*}
  \mathrm{sign}\bigl(\langle r_a, r_e\rangle\bigr) = (-1)^{y_a+1} \cdot (-1)^{\hat{y}} = -(-1)^{y_a + \hat{y}}.
\end{equation*}
When $y_a = \hat{y}$ (label agreement): sign $= -1$. When $y_a \neq \hat{y}$ (label disagreement): sign $= +1$.

\emph{Step 4 (sign of the feature kernel under class symmetry).} Under Assumption~\ref{assn:id-confidence}(iv) with $C = 2$, the class-conditional feature means are $\mu_0 = -\mu_1 =: -\mu$ (centering forces $\mu_0 + \mu_1 = 0$). Since the anchor draw $x_a$ and the eval draw $x_e$ are independent under $p_0 \otimes p_0$, conditioning on $(y_a, \hat{y}(x_e))$ makes $\phi(x_a)$ and $\phi(x_e)$ conditionally independent, so
\begin{equation*}
  \bbE\bigl[K_{ij} \mid y_a, \hat{y}\bigr]
  = \langle \bbE[\phi(x_a) \mid y_a],\, \bbE[\phi(x_e) \mid \hat{y}(x_e)]\rangle
  = \langle \mu_{y_a},\, m_{\hat y}\rangle,
\end{equation*}
where we write $m_c := \bbE[\phi(x_e) \mid \hat y(x_e) = c]$ for the \emph{predicted}-class conditional mean. By Assumption~\ref{assn:id-confidence}(v), $\bbP_{p_0}(\hat y(x) \neq y(x)) \le \tau$, and a standard total-probability decomposition gives
\begin{equation*}
  m_c = \mu_c \,+\, O(\tau)
\end{equation*}
with the $O(\tau)$ constant controlled by the uniform backbone bound $B_\phi$ and the posterior $\bbP(y = c \mid \hat y = c) \ge 1 - O(\tau)$ implied by (v) and the law of total probability. Hence $\bbE[K_{ij} \mid y_a, \hat y] = \langle \mu_{y_a}, \mu_{\hat y}\rangle + O(\tau B_\phi^2)$, with $\langle \mu_{y_a}, \mu_{\hat y}\rangle = +\|\mu\|^2$ under label agreement and $-\|\mu\|^2$ under disagreement. For $\tau B_\phi^2 < \|\mu\|^2$ (denoting the regime of any useful ID classifier) the conditional feature kernel is strictly positive under label agreement and strictly negative under label disagreement.

\emph{Step 5 (combining Steps 3 and 4).} The product $C_{ij} K_{ij}$ has sign equal to the product of the signs computed in Steps 3 and 4:
\begin{itemize}[leftmargin=1.5em]
  \item Label agreement ($y_a = \hat{y}$): $C_{ij} < 0$ and $\bbE[K_{ij}\,|\,y_a = \hat{y}] > 0$, so the contribution to $\bbE[C_{ij} K_{ij}]$ is \emph{negative}.
  \item Label disagreement ($y_a \neq \hat{y}$): $C_{ij} > 0$ and $\bbE[K_{ij}\,|\,y_a \neq \hat{y}] < 0$, so the contribution to $\bbE[C_{ij} K_{ij}]$ is also \emph{negative}.
\end{itemize}
Both events contribute negatively, so $\bbE[C_{ij} K_{ij}] < 0$ (Figure~\ref{fig:double-negative}). This proves the binary case.

\emph{Step 6 (multiclass extension).} For general $C$, we again condition on the label pair $(y_a, \hat{y})$. Conditional independence of $\phi(x_a)$ and $\phi(x_e)$ gives
\begin{equation*}
  \bbE[K_{ij}\,|\,y_a = c, \hat{y} = c'] = \langle \mu_c, \mu_{c'}\rangle.
\end{equation*}
Centering ($\sum_c \mu_c = 0$) and the class-symmetry condition (iv) force $\langle \mu_c, \mu_{c'}\rangle$ to be positive when $c = c'$ and to have average value $-\|\mu\|^2/(C-1) < 0$ when $c \neq c'$. The residual inner product $\bbE[C_{ij}\,|\,y_a = c, \hat{y} = c']$ generalizes to a label-pair-indexed quantity that, under Assumption~\ref{assn:id-confidence}(i)--(ii), is negative for $c = c'$ (the generalization of the binary `$-p_{x_e}(\hat{y})$' term) and positive for $c \neq c'$. The product $\bbE[C_{ij} K_{ij}\,|\,c, c']$ is therefore negative for both $c = c'$ and $c \neq c'$, and summing over $(c, c')$ weighted by their joint probability completes the argument.
\end{proof}

\paragraph{Intuitive reading.} Under label agreement, anchor and eval points tend to lie in the same class region, so their features are positively correlated ($K_{ij} > 0$), but the residual inner product is negative ($C_{ij} < 0$) because the anchor wants to push toward the correct class while the disagreement term wants to push away from it. Under label disagreement, features are negatively correlated ($K_{ij} < 0$), but the residual inner product is positive ($C_{ij} > 0$) for the mirror reason. The product $C_{ij} K_{ij}$ is negative in \emph{both} cases, and these are the only two cases, so the expectation is negative regardless of how the labels mix. This is why the ID regime gives a strictly damped anchor descent rather than a random-sign perturbation: the mechanism aligns the two factors of the product so that the sign is consistent across both label configurations, $y_a = \hat y$ and $y_a \neq \hat y$ alike.

\paragraph{Contrast with OOD.} Under Corollary~\ref{cor:feature-orth}, the feature kernel $K_{ij}$ shrinks to size $O(\epsilon)$ in the far-OOD regime, so the product $C_{ij} K_{ij}$ also shrinks to $O(\epsilon)$ regardless of the label configuration. The sign of $C_{ij}$ no longer has a $K_{ij}$ of matching sign to multiply with; the whole product goes to zero. Combining with Proposition~\ref{prop:id-interference}, we have \emph{in the far-OOD regime (where $\epsilon B_\phi^2 B_C < \gamma$)}
\begin{equation*}
  \bbE\bigl[\lambda_{\mathrm{OOD}}\bigr] - \bbE\bigl[\lambda_{\mathrm{ID}}\bigr] \approx \eta\, \bbE\bigl[\langle g_a, g_e\rangle_F^{\mathrm{ID}}\bigr] - \eta\, \bbE\bigl[\langle g_a, g_e\rangle_F^{\mathrm{OOD}}\bigr] < 0,
\end{equation*}
where the first expectation is $-\gamma < 0$ by Proposition~\ref{prop:id-interference} and the second is $O(\epsilon)$ by Corollary~\ref{cor:feature-orth}. We note that for near-OOD, where $\epsilon$ is not small, the signed inequality is not rigorously established.

\subsection{Why Feature-Orthogonality Holds Under OOD: Nagler's Tilting}
\label{sec:theory-nagler}

Assumption~\ref{assn:feat-orth} ($\epsilon$ small in the OOD regime) is the key empirical claim of this paper. We now argue that it is a structural consequence of the attention mechanism in trained transformer PFNs, using the framework of \citet{nagler2023statistical}. The goal of this subsection is to give a mechanistic reason why the feature vector $\phi(x_e)$ collapses to a prediction-uninformative direction when $x_e$ is OOD.

Nagler's argument as stated concludes pointwise feature collapse: every far-OOD query maps to the same direction $\phi_\infty$. That conclusion depends on a few idealized assumptions: a one-layer attention mechanism (\citet{nagler2023statistical} assume a single layer; PFNs in practice are multi-layer) and the strict total-variation limit $\|p_1 - p_0\|_{\mathrm{TV}} \to 1$. Hence, this behavior is best read as a limit point instead of a literal description of any specific experiment. We use the argument here to motivate Assumption~\ref{assn:feat-orth}, which is much weaker than full collapse: it asks only that anchor and eval features be on average near-perpendicular. Note that the weaker condition can hold even when full collapse fails (OOD features fanning out within a subspace orthogonal to the span of the ID features give orthogonality without collapse), and the separation theorem (Theorem~\ref{thm:main-separation}) only needs the weaker condition.

\paragraph{What attention computes (informally).} In a PFN's forward pass, the query $x_e$ attends to the support set and produces a feature as a weighted average of transformed support tokens. The weights are softmax-normalized query-key similarities: $\alpha_j \propto \exp(v(x_e)^\top W_q\, s_j)$, where $s_j$ is the $j$-th support token and $v(x_e)$ is the query's encoding. When the query resembles some support samples strongly, those attention weights dominate, and the feature is determined by a \emph{local neighborhood} of $x_e$ in the support set. When no support sample resembles the query, the weights are approximately uniform, and the feature becomes a \emph{global average} over the support set. The rest of this subsection formalizes this explanation of OOD feature collapse.

\paragraph{Locality of trained PFNs.} \citet[Theorem 5.4]{nagler2023statistical} establishes that any PFN with vanishing bias on a sufficiently rich class $\cP$ of ID distributions must be \emph{local}: the prediction at a query $x$ depends only on support samples $(X_i, Y_i)$ with $\|X_i - x\| \le \epsilon_n$, where $\epsilon_n \to 0$. Labels of samples far from $x$ can be flipped arbitrarily without changing the prediction. A well-trained TabPFN approximates this property empirically.

\paragraph{Attention as exponential tilting.} \citet[Theorem 6.3]{nagler2023statistical} promotes the informal picture above to an exact statement about the \emph{limiting} feature (in the regime where the bias converges to a deterministic function $\bar{q}_\theta$). The limiting feature takes the form
\begin{equation}
  \bar{u}(x) = \sum_{h=1}^H W_v^{(h)}\, \bbE_{V \sim g_h(\cdot \mid x)}[V],
  \label{eq:tilted-feature}
\end{equation}
where the attention head $h$ induces the exponentially tilted measure
\begin{equation}
  g_h(s \mid x) = \frac{\exp\bigl(v(x)^\top W_q^{(h)}\, s\bigr)}{\bbE_{V \sim p_0}\bigl[\exp(v(x)^\top W_q^{(h)} V)\bigr]}\, p_0(s)
  \label{eq:tilt}
\end{equation}
on the support marginal $p_0$. The vector $v(x) \in \bbR^{d+1}$ is the query's encoding and $W_q^{(h)}, W_v^{(h)}$ are the learned query and value projection matrices for head $h$.

\paragraph{Interpreting Equation~\eqref{eq:tilt}.} The tilted measure $g_h(\cdot \mid x)$ is $p_0$ reweighted by the exponential factor $\exp(v(x)^\top W_q^{(h)} s)$, normalized to sum to one. Two extreme cases help build intuition:
\begin{itemize}[leftmargin=1.5em]
  \item If the tilt $\exp(v(x)^\top W_q^{(h)} s)$ has a sharp peak at some $s^* \in \mathrm{supp}(p_0)$, then $g_h(\cdot \mid x)$ concentrates on a small neighborhood of $s^*$. The resulting feature $\bbE_{V \sim g_h}[V]$ is close to $s^*$, and depends strongly on $x$ (which determines where the peak is).
  \item If the tilt is approximately constant over $s \sim p_0$, the normalized measure $g_h(\cdot \mid x)$ is indistinguishable from $p_0$. The feature is $\bbE_{V \sim p_0}[V]$, the $p_0$-mean, and depends on $x$ only through higher-order perturbations.
\end{itemize}

\begin{figure}[t]
\centering
\begin{subfigure}[b]{0.48\textwidth}
\centering
\begin{tikzpicture}[scale=0.95]
  \draw[->, gray!60] (-0.2,0) -- (5.6,0) node[right, font=\scriptsize, gray!60] {$s$};
  \draw[->, gray!60] (0,-0.1) -- (0,2.5);
  \draw[thick, blue!55!black, smooth, samples=100, domain=0:5.4]
    plot ({\x}, {1.5*exp(-(\x-2.5)*(\x-2.5)/0.8)});
  \draw[thick, orange!75!black, dashed, smooth, samples=100, domain=0:5.4]
    plot ({\x}, {0.2 + 1.5*exp(-(\x-1.8)*(\x-1.8)/0.18)});
  \draw[very thick, green!50!black, smooth, samples=100, domain=0:5.4]
    plot ({\x}, {2.1*exp(-(\x-2.05)*(\x-2.05)/0.18)});
  \node[blue!55!black, font=\scriptsize] at (3.9, 0.65) {$p_0(s)$};
  \node[orange!75!black, font=\scriptsize] at (0.65, 0.6) {$\psi(s; x_e)$};
  \node[green!50!black, font=\scriptsize] at (2.85, 2.25) {$g_h(s \mid x_e)$};
  \draw[->, red, line width=0.9pt] (1.8, -0.55) -- (1.8, -0.08);
  \node[below, red, font=\scriptsize] at (1.8, -0.55) {$x_e$ (ID)};
\end{tikzpicture}
\caption{\textbf{ID query.} The tilt has a sharp peak near $x_e$, so the tilted measure $g_h(\cdot \mid x_e)$ concentrates on a small neighborhood of $x_e$ within the support. The mean feature depends strongly on where $x_e$ is.}
\end{subfigure}
\hfill
\begin{subfigure}[b]{0.48\textwidth}
\centering
\begin{tikzpicture}[scale=0.95]
  \draw[->, gray!60] (-0.2,0) -- (5.6,0) node[right, font=\scriptsize, gray!60] {$s$};
  \draw[->, gray!60] (0,-0.1) -- (0,2.5);
  \draw[thick, blue!55!black, smooth, samples=100, domain=0:5.4]
    plot ({\x}, {1.5*exp(-(\x-2.5)*(\x-2.5)/0.8)});
  \draw[thick, orange!75!black, dashed, smooth, samples=100, domain=0:5.4]
    plot ({\x}, {0.85 + 0.015*\x});
  \draw[very thick, green!50!black, smooth, samples=100, domain=0:5.4]
    plot ({\x}, {1.32*exp(-(\x-2.5)*(\x-2.5)/0.8)});
  \node[blue!55!black, font=\scriptsize] at (1.4, 1.55) {$p_0(s)$};
  \node[orange!75!black, font=\scriptsize] at (4.0, 1.2) {$\psi(s; x_e)$};
  \node[green!50!black, font=\scriptsize] at (2.55, 0.45) {$g_h \approx p_0$};
  \draw[->, red, line width=0.9pt] (4.8, -0.55) -- (4.8, -0.08);
  \node[below, red, font=\scriptsize] at (4.8, -0.55) {$x_e$ (far-OOD)};
\end{tikzpicture}
\caption{\textbf{OOD query.} The tilt is nearly flat over the support of $p_0$, so $g_h(\cdot \mid x_e) \approx p_0$. The mean feature degenerates to $\bbE_{V \sim p_0}[V]$, the same input-independent direction for every OOD query.}
\end{subfigure}
\caption{\textbf{Exponential tilting of the support measure (Nagler Theorem 6.3).} Each attention head reweights the support distribution $p_0$ (solid blue) by a tilt factor $\psi(s; x_e) = \exp(v(x_e)^\top W_q s)$ (dashed orange) to produce the head's effective sampling distribution $g_h(s \mid x_e)$ (thick green). When the query $x_e$ is close to the support, the tilt concentrates the measure on a small neighborhood of $x_e$ and the resulting feature is query-specific. When $x_e$ is far from the support, the tilt is approximately constant and the measure collapses back to $p_0$, giving the same feature $\phi_\infty$ for every far-OOD query.}
\label{fig:tilting}
\end{figure}
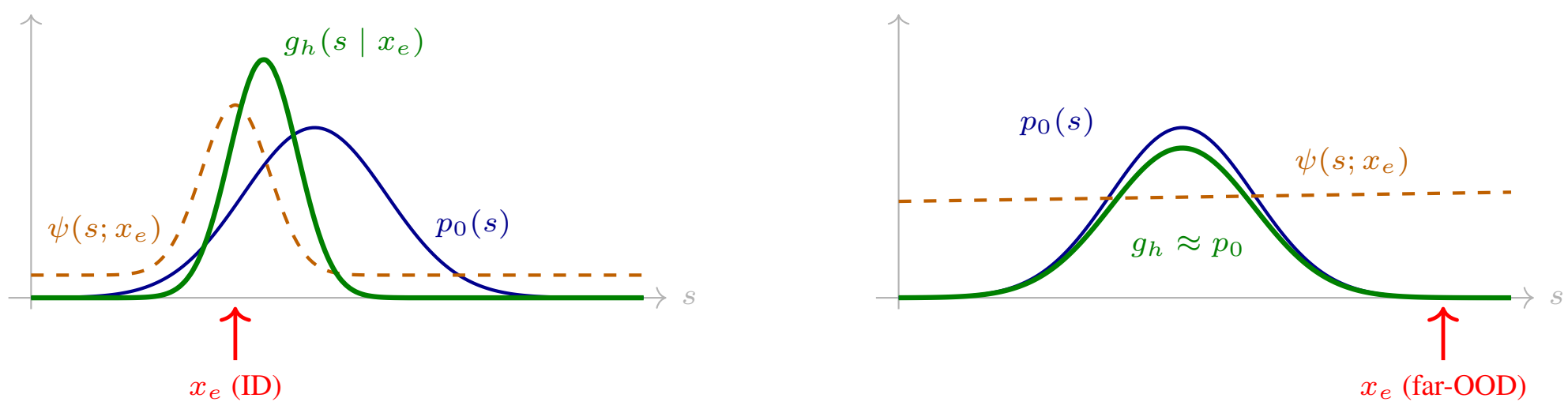

\begin{figure}[t]
\centering
\begin{subfigure}[b]{0.48\textwidth}
\centering
\begin{tikzpicture}[scale=1.1, >={Stealth[scale=1.0]}]
  \draw[->, gray!50] (-0.3,0) -- (4.7,0) node[right, gray!50, font=\scriptsize] {$\phi_1$};
  \draw[->, gray!50] (0,-0.3) -- (0,2.8) node[above, gray!50, font=\scriptsize] {$\phi_2$};
  \draw[fill=blue!8, draw=blue!30, thick] plot [smooth cycle, tension=0.7] coordinates {(1.2, 0.8) (2.3, 1.0) (2.7, 1.8) (2.0, 2.2) (1.0, 1.8) (0.7, 1.2)};
  \node[font=\scriptsize] at (3.2, 0.95) {ID features};
  \foreach \x/\y in {1.3/1.1, 1.5/1.6, 2.0/1.8, 2.1/1.3, 2.4/1.7} {
    \filldraw[blue!70!black] (\x, \y) circle (1.8pt);
  }
  \foreach \x/\y in {1.5/1.3, 1.8/1.1, 2.2/1.5, 1.7/1.9, 1.4/1.45} {
    \draw[orange!80!black, line width=1pt] (\x-0.08,\y-0.08) -- (\x+0.08,\y+0.08);
    \draw[orange!80!black, line width=1pt] (\x-0.08,\y+0.08) -- (\x+0.08,\y-0.08);
  }
  \draw[gray!30, fill=white, rounded corners=2pt] (3.05, 1.85) rectangle (4.2, 2.5);
  \filldraw[blue!70!black] (3.25, 2.35) circle (1.8pt);
  \node[right, font=\scriptsize] at (3.35, 2.35) {anchor};
  \draw[orange!80!black, line width=1pt] (3.17, 1.97) -- (3.33, 2.13);
  \draw[orange!80!black, line width=1pt] (3.17, 2.13) -- (3.33, 1.97);
  \node[right, font=\scriptsize] at (3.35, 2.05) {eval};
\end{tikzpicture}
\caption{\textbf{ID eval.} Anchor features (blue dots) and eval features (orange crosses) both lie in the region occupied by ID features and intermix. Their inner products $K_{ij} = \langle \phi(x_a^{(i)}), \phi(x_e^{(j)})\rangle$ are on average positive (Assumption~\ref{assn:id-confidence}(iv) $\rightarrow$ Proposition~\ref{prop:id-interference}).}
\end{subfigure}
\hfill
\begin{subfigure}[b]{0.48\textwidth}
\centering
\begin{tikzpicture}[scale=1.1]
  \draw[->, gray!50] (-0.3,0) -- (4.7,0) node[right, gray!50, font=\scriptsize] {$\phi_1$};
  \draw[->, gray!50] (0,-0.3) -- (0,2.8) node[above, gray!50, font=\scriptsize] {$\phi_2$};
  \draw[fill=blue!8, draw=blue!30, thick] plot [smooth cycle, tension=0.7] coordinates {(1.2, 0.8) (2.3, 1.0) (2.7, 1.8) (2.0, 2.2) (1.0, 1.8) (0.7, 1.2)};
  \node[font=\scriptsize] at (3.2, 0.95) {ID features};
  \foreach \x/\y in {1.3/1.1, 1.5/1.6, 2.0/1.8, 2.1/1.3, 2.4/1.7} {
    \filldraw[blue!70!black] (\x, \y) circle (1.8pt);
  }
  \filldraw[red!60!black] (0.6, 0.45) circle (2pt);
  \node[above left=1pt, red!60!black, font=\scriptsize] at (1.6, 0.15) {$\phi_\infty$};
  \foreach \x/\y in {0.5/0.4, 0.65/0.5, 0.55/0.35, 0.7/0.42, 0.48/0.52} {
    \draw[orange!80!black, line width=1pt] (\x-0.06,\y-0.06) -- (\x+0.06,\y+0.06);
    \draw[orange!80!black, line width=1pt] (\x-0.06,\y+0.06) -- (\x+0.06,\y-0.06);
  }
  \draw[orange!50!black, dashed, thin] (0.6, 0.45) circle (0.3);
  \draw[<->, gray!60!black, line width=0.7pt] (0.7, 0.55) -- (1.25, 1.05);
  \node[gray!70!black, font=\scriptsize, rotate=47] at (1.02, 0.82) {};
  \draw[gray!30, fill=white, rounded corners=2pt] (3.05, 1.85) rectangle (4.2, 2.5);
  \filldraw[blue!70!black] (3.25, 2.35) circle (1.8pt);
  \node[right, font=\scriptsize] at (3.35, 2.35) {anchor};
  \draw[orange!80!black, line width=1pt] (3.17, 1.97) -- (3.33, 2.13);
  \draw[orange!80!black, line width=1pt] (3.17, 2.13) -- (3.33, 1.97);
  \node[right, font=\scriptsize] at (3.35, 2.05) {eval};
\end{tikzpicture}
\caption{\textbf{OOD eval.} By Nagler's tilting (Figure~\ref{fig:tilting}), eval features (orange crosses) collapse onto the single input-independent direction $\phi_\infty$ (red dot), far from the ID features. After centering by $\phi_\infty$, anchor--eval inner products shrink to $O(\epsilon)$.}
\end{subfigure}
\caption{\textbf{Feature-space picture of Corollary~\ref{cor:feature-orth}.} Under ID \textbf{(left)}, anchor and eval features coexist in the ID feature region and the kernel $K_{ij}$ is typically positive. Under OOD \textbf{(right)}, eval features are forced by the attention mechanism onto a single input-independent direction $\phi_\infty$, so the centered kernel $\langle \tilde\phi(x_a), \tilde\phi(x_e)\rangle$ becomes small (Assumption~\ref{assn:feat-orth}). This is what switches the gradient geometry from Figure~\ref{fig:three-regimes} (middle) to Figure~\ref{fig:three-regimes} (left).}
\label{fig:feature-collapse}
\end{figure}
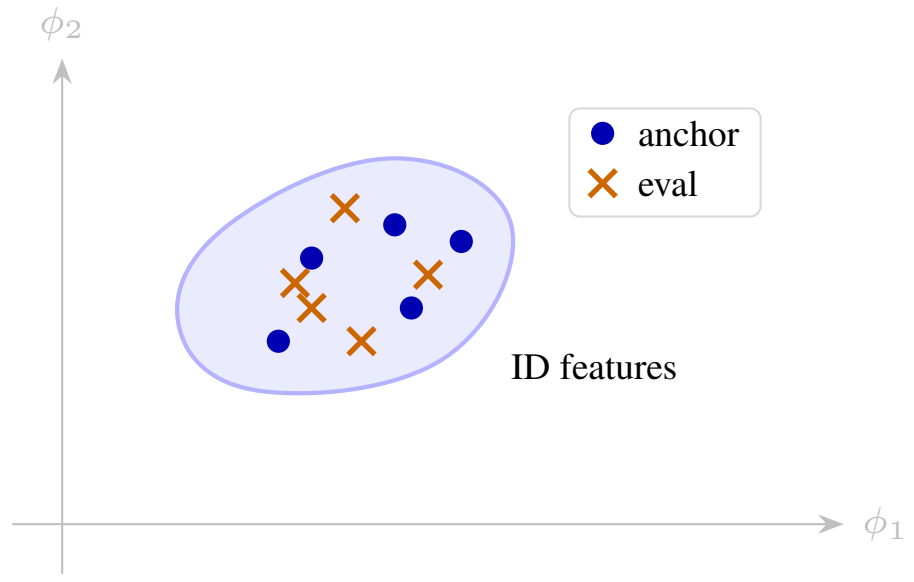
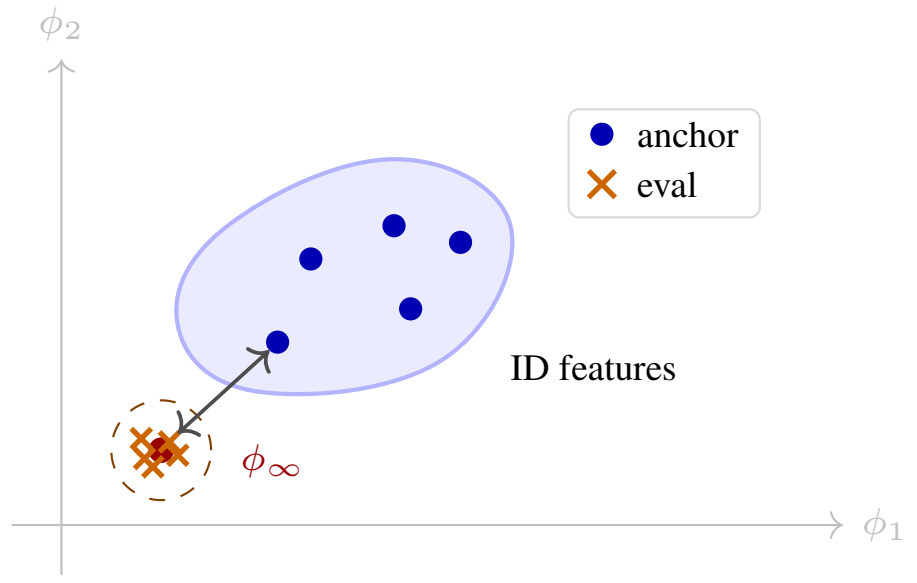

\paragraph{Implication for OOD queries.} Consider a query $x_e$ drawn from a shifted distribution $p_1 \neq p_0$. The tilt~\eqref{eq:tilt} assigns high mass to support samples $s$ that are similar to $x_e$ in the metric induced by $W_q^{(h)}$. Two regimes emerge, corresponding to the two extreme cases above (Figure~\ref{fig:tilting}):

\begin{itemize}[leftmargin=1.5em]
  \item \emph{ID query} ($x_e \sim p_0$, close to the support): $v(x_e)^\top W_q^{(h)} s$ has a sharp maximum over $s \sim p_0$, so $g_h(\cdot \mid x_e)$ concentrates on a localized neighborhood of $x_e$ within the support. The feature $\bar{u}(x_e)$ is then a kernel-smoothed average over that neighborhood and depends strongly on $x_e$; distinct ID queries produce distinct features.
  \item \emph{Far-OOD query} ($x_e$ outside the support of $p_0$): no support sample $s$ is especially similar to $x_e$, so $v(x_e)^\top W_q^{(h)} s$ is approximately constant over $s \sim p_0$. The tilt degenerates, $g_h(\cdot \mid x_e) \to p_0$, and the feature collapses to
  \begin{equation}
    \bar{u}(x_e) \to \phi_\infty := \sum_{h=1}^H W_v^{(h)}\, \bbE_{V \sim p_0}[V],
    \label{eq:phi-infinity}
  \end{equation}
  a single input-\emph{independent} vector. Figure~\ref{fig:feature-collapse} visualizes the resulting ID-vs-OOD feature geometry: under ID, features occupy a rich, input-dependent region and anchor-eval inner products are $\Theta(1)$; under far-OOD they collapse to $\phi_\infty$ and centered inner products shrink to $O(\epsilon)$.
\end{itemize}

\paragraph{From feature collapse to feature orthogonality.} Define the centered feature $\tilde{\phi}(x) := \phi(x) - \phi_\infty$. For far-OOD $x_e$, $\tilde{\phi}(x_e) \to 0$. For ID $x_a$, $\tilde{\phi}(x_a)$ retains input-specific structure inherited from the local tilt. Consequently,
\begin{equation}
  \bigl|\langle \tilde{\phi}(x_a),\, \tilde{\phi}(x_e)\rangle\bigr| = \|\tilde{\phi}(x_a)\|\, \|\tilde{\phi}(x_e)\|\, \bigl|\cos(\tilde{\phi}(x_a), \tilde{\phi}(x_e))\bigr| \longrightarrow 0
\end{equation}
as $x_e$ recedes from $\mathrm{supp}(p_0)$: the factor $\|\tilde{\phi}(x_e)\| \to 0$ forces the product to zero regardless of the cosine. This verifies Assumption~\ref{assn:feat-orth} in the far-OOD limit, with $\epsilon$ controlled by $\|\tilde{\phi}(x_e)\|$.

\paragraph{Note on centering.} Assumption~\ref{assn:id-confidence}(iii) asked for feature centering $\bbE_{p_0}[\phi(x)] = 0$. This is either automatic (under the tilting picture with a centering layer in the architecture) or can be enforced by working with the empirically centered feature $\tilde{\phi}$; either way, all preceding statements apply verbatim.

The feature-collapse picture above gives $\epsilon \to 0$ in Assumption~\ref{assn:feat-orth} only in the far-OOD limit $\|p_1 - p_0\|_{\mathrm{TV}} \to 1$, via Proposition~\ref{prop:prediction-collapse}. For near-OOD, OOD features stay close to ID features and the anchor-eval cosine remains $\Theta(1)$, so Theorem~\ref{thm:main-separation}'s signed lower bound is vacuous. 

\subsection{Multi-Step Extension}
\label{sec:theory-tstep}

Proposition~\ref{prop:first-step} analyzes the first gradient step. Algorithm~\ref{alg:method} uses $T = 50$ steps, so we need a multi-step statement. The main idea is that in a small-step regime, the $T$-step decrement is the sum of $T$ first-step decrements plus a telescoping error of size $O(T^2 \eta^2 L)$. The sign of the cross-term at the first step therefore determines the sign of the cumulative descent, provided that the cross-term does not flip sign along the trajectory.

\begin{assumption}[Small-step regime]
\label{assn:small-step}
The joint loss $\cL$ has $L$-Lipschitz gradient on the trajectory $\{W_t\}_{t=0}^T$, and $T \eta L \le 1/2$.
\end{assumption}

\paragraph{The smoothness constant.} For the last-layer parametrization $q_W(y \mid x) = \mathrm{softmax}(W \phi(x))_y$, the Hessian of the per-sample CE loss with respect to $W$ is the Kronecker product
\begin{equation*}
  \nabla_W^2\, \ell_{\mathrm{CE}}(W; x, y) = J(q_W) \otimes \phi(x) \phi(x)^\top,
\end{equation*}
where $J(q) := \mathrm{diag}(q) - qq^\top$ is the Jacobian of the softmax at $q = q_W(\cdot \mid x)$. For any unit vector $v \in \bbR^C$, $v^\top J(q) v = \sum_c q_c v_c^2 - (\sum_c q_c v_c)^2 = \mathrm{Var}_{c \sim q}(v_c) \le 1/2$, with the maximum achieved by a two-class concentrated $q$ and $v$ pointing along the corresponding axis. Hence $\|J(q)\|_{\mathrm{op}} \le 1/2$. The second Kronecker factor has operator norm $\|\phi(x)\|^2 \le B_\phi^2$, and the operator norm of a Kronecker product is the product of operator norms, so
\begin{equation}
  L \le B_\phi^2 / 2.
  \label{eq:smoothness-bound}
\end{equation}
The DCE Hessian admits the same bound: DCE is itself a cross-entropy against $u_{\hat y}$, so its Hessian has the same $J(q) \otimes \phi\phi^\top$ form. Averaging over the batch preserves the bound. 

\begin{proposition}[$T$-step anchor descent]
\label{prop:T-step}
Under Assumption~\ref{assn:small-step}, let $\bar{g}_a := \tfrac{1}{T} \sum_{t=0}^{T-1} \nabla \cL_{\mathrm{anc}}(W_t)$ and $\bar{g}_e := \tfrac{1}{T} \sum_{t=0}^{T-1} \nabla \cL_{\mathrm{dis}}(W_t)$ denote the time-averaged gradients along the trajectory. Then
\begin{equation}
  \cL_{\mathrm{anc}}(W_T) - \cL_{\mathrm{anc}}(W_0) = -T \eta \bigl(\|\bar{g}_a\|_F^2 + \langle \bar{g}_a,\, \bar{g}_e\rangle_F\bigr) + O(T^2 \eta^2 L\, B_g^2),
  \label{eq:T-step}
\end{equation}
where $B_g := \max_t \|\nabla \cL(W_t)\|_F$ is a trajectory-uniform gradient bound. Furthermore, if $\langle g_a^{(t)},\, g_a^{(0)}\rangle_F \ge (1 - \kappa)\, \|g_a^{(t)}\|_F\, \|g_a^{(0)}\|_F$ for all $t \le T$ (gradients do not rotate substantially), then $\|\bar{g}_a\|_F^2 = \|g_a^{(0)}\|_F^2\, (1 + O(\kappa))$.
\end{proposition}

\begin{proof}
We apply Proposition~\ref{prop:first-step} at each step of the trajectory and sum, controlling the accumulated error from the Hessian terms.

\emph{Step 1 (first-step expansion at each $t$).} For each $t = 0, 1, \ldots, T-1$, apply Taylor's theorem to $\cL_{\mathrm{anc}}$ on the segment $[W_t, W_{t+1}]$ with $W_{t+1} = W_t - \eta(g_a^{(t)} + g_e^{(t)})$ and $g_\star^{(t)} := \nabla \cL_\star(W_t)$ for $\star \in \{\mathrm{anc}, \mathrm{dis}\}$:
\begin{equation*}
  \cL_{\mathrm{anc}}(W_{t+1}) - \cL_{\mathrm{anc}}(W_t) = -\eta\, \|g_a^{(t)}\|_F^2 - \eta\, \langle g_a^{(t)}, g_e^{(t)}\rangle_F + R_t,
\end{equation*}
with $|R_t| \le \tfrac{1}{2} L \eta^2 \|g_a^{(t)} + g_e^{(t)}\|_F^2 \le \tfrac{1}{2} L \eta^2\, B_g^2$.

\emph{Step 2 (telescoping).} Summing over $t = 0, \ldots, T-1$,
\begin{equation*}
  \cL_{\mathrm{anc}}(W_T) - \cL_{\mathrm{anc}}(W_0) = -\eta \sum_{t=0}^{T-1} \|g_a^{(t)}\|_F^2 - \eta \sum_{t=0}^{T-1} \langle g_a^{(t)}, g_e^{(t)}\rangle_F + \sum_{t=0}^{T-1} R_t,
\end{equation*}
with the Hessian remainder bounded by $\sum_t |R_t| \le \tfrac{1}{2} T L \eta^2 B_g^2$.

\emph{Step 3 (replacing sums by time-averaged gradients).} The trajectory displacement is $\|W_t - W_0\| \le t \eta B_g \le T\eta B_g$, and $L$-Lipschitzness of $\nabla \cL$ gives $\|g_\star^{(t)} - g_\star^{(0)}\|_F \le L T \eta B_g$ for $\star \in \{\mathrm{anc}, \mathrm{dis}\}$ at every $t$. Hence each $g_\star^{(t)}$ lies within an $L T\eta B_g$-ball of $\bar g_\star$, and
\begin{equation*}
  \bigl|\tfrac{1}{T}\sum_{t} \|g_a^{(t)}\|_F^2 - \|\bar g_a\|_F^2\bigr| \le (L T\eta B_g)^2, \qquad
  \bigl|\tfrac{1}{T}\sum_{t} \langle g_a^{(t)}, g_e^{(t)}\rangle_F - \langle \bar g_a, \bar g_e\rangle_F\bigr| \le (L T\eta B_g)^2,
\end{equation*}
each by direct expansion and Cauchy--Schwarz. Multiplying by $T\eta$ converts this to an error of size $T^3 \eta^3 L^2 B_g^2$ in the anchor-loss decrement. Under Assumption~\ref{assn:small-step} ($T\eta L \le 1/2$), this is dominated by the Hessian remainder $T^2 \eta^2 L B_g^2$, so the total remainder in~\eqref{eq:T-step} is $O(T^2\eta^2 L B_g^2)$.

\emph{Step 4 (small-rotation corollary).} If $g_a^{(t)}$ stays within angle $\arccos(1 - \kappa)$ of $g_a^{(0)}$ for all $t$, then $\|\bar{g}_a\|_F^2$ is within a $(1 - O(\kappa))$ factor of $\|g_a^{(0)}\|_F^2$ by direct calculation: decompose each $g_a^{(t)} = \alpha_t g_a^{(0)} + \beta_t h_t$ with $\|h_t\| = 1$, $h_t \perp g_a^{(0)}$, and $\alpha_t \|g_a^{(0)}\|_F \ge (1 - \kappa) \|g_a^{(t)}\|_F$, then average.
\end{proof}

Proposition~\ref{prop:T-step} inherits the sign structure of the one-step result. If the OOD feature-orthogonality of Corollary~\ref{cor:feature-orth} holds \emph{uniformly} along the trajectory (which requires only that the backbone is frozen, not that $W$ stays close to $W_0$: the feature-space kernel $K_{ij}$ does not depend on $W$ at all), then OOD anchor descent at the time-averaged rate is $\approx -T\eta \|\bar{g}_a\|_F^2$. By Proposition~\ref{prop:id-interference}, ID anchor descent is strictly damped to $-T\eta(\|\bar{g}_a\|_F^2 + \langle \bar{g}_a, \bar{g}_e\rangle_F)$ with $\langle \bar{g}_a, \bar{g}_e\rangle_F < 0$.

\subsection{Main Separation Theorem}
\label{sec:theory-main-separation}

The preceding subsections have established: (i) anchor descent is controlled by the gradient cross-term (Proposition~\ref{prop:first-step}); (ii) under OOD feature orthogonality the cross-term is $O(\epsilon)$ (Corollary~\ref{cor:feature-orth}); (iii) under ID confidence the cross-term has a strictly negative expectation $-\gamma$ for some $\gamma > 0$ (Proposition~\ref{prop:id-interference}); (iv) the first-step analysis lifts to $T$ steps in the small-step regime (Prop~\ref{prop:T-step}). We now compose these into a separation theorem that underpins the detector.

\begin{theorem}[Main separation]
\label{thm:main-separation}
Let $p_1$ be an eval distribution and write $\bbE[\lambda_{\mathrm{ID}}]$, $\bbE[\lambda_{\mathrm{OOD}}]$ for the expected post-finetuning anchor losses when $\Xeval$ is drawn i.i.d.\ from $p_0$ and from $p_1$ respectively. Suppose the core gradient-geometry assumptions hold: A1--A3 of Table~\ref{tab:assumptions} (design and boundedness), together with Assumption~\ref{assn:small-step} (small-step regime), Assumption~\ref{assn:feat-orth} (with parameter $\epsilon$ under $p_1$), and Assumption~\ref{assn:id-confidence} (under $p_0$). Let $\gamma := -\bbE_{p_0 \otimes p_0}[C_{ij} K_{ij}] > 0$ denote the ID interference strength from Proposition~\ref{prop:id-interference}. Then
\begin{equation}
  \bbE[\lambda_{\mathrm{ID}}] - \bbE[\lambda_{\mathrm{OOD}}]
  \ge T\eta\bigl(\gamma - \epsilon B_\phi^2 B_C\bigr) - O(T^2 \eta^2 L\, B_g^2),
  \label{eq:main-separation}
\end{equation}
where $B_\phi$, $B_C$, $L$, $B_g$ are the constants defined earlier. In particular, whenever $\epsilon < \gamma / (B_\phi^2 B_C)$ and $T\eta$ is in the small-step regime of Assumption~\ref{assn:small-step}, the right-hand side is strictly positive and $\bbE[\lambda_{\mathrm{OOD}}] < \bbE[\lambda_{\mathrm{ID}}]$.
\end{theorem}

\begin{proof}
Apply Proposition~\ref{prop:T-step} separately under $p_0$ and $p_1$. Let $\bar g_a^{\mathrm{ID}}, \bar g_e^{\mathrm{ID}}$ and $\bar g_a^{\mathrm{OOD}}, \bar g_e^{\mathrm{OOD}}$ denote the respective time-averaged gradients, and let $\lambda(W_0) = \cL_{\mathrm{anc}}(W_0)$ be the pre-finetuning anchor loss (the same in both regimes since it depends only on the anchor, not on the eval batch). Then
\begin{align*}
  \lambda_{\mathrm{ID}} - \lambda(W_0)
  &= -T\eta\bigl(\|\bar g_a^{\mathrm{ID}}\|_F^2 + \langle \bar g_a^{\mathrm{ID}}, \bar g_e^{\mathrm{ID}}\rangle_F\bigr) + O(T^2 \eta^2 L B_g^2), \\
  \lambda_{\mathrm{OOD}} - \lambda(W_0)
  &= -T\eta\bigl(\|\bar g_a^{\mathrm{OOD}}\|_F^2 + \langle \bar g_a^{\mathrm{OOD}}, \bar g_e^{\mathrm{OOD}}\rangle_F\bigr) + O(T^2 \eta^2 L B_g^2).
\end{align*}
Subtract:
\begin{equation}
\begin{aligned}
  \bbE[\lambda_{\mathrm{ID}} - \lambda_{\mathrm{OOD}}]
  &= T\eta\,\Bigl[\bbE\|\bar g_a^{\mathrm{OOD}}\|_F^2 - \bbE\|\bar g_a^{\mathrm{ID}}\|_F^2 \\
  &\qquad + \bbE\langle \bar g_a^{\mathrm{OOD}}, \bar g_e^{\mathrm{OOD}}\rangle_F - \bbE\langle \bar g_a^{\mathrm{ID}}, \bar g_e^{\mathrm{ID}}\rangle_F\Bigr] + O(T^2\eta^2 L B_g^2).
\end{aligned}
\label{eq:separation-decomp}
\end{equation}

The anchor gradient $\bar g_a$ depends only on the anchor set and the classifier at $W_t$; it is the same in expectation at $t = 0$ whether the eval batch is from $p_0$ or $p_1$, because the finetuning trajectory is anchored at $W_0$ and the leading term is the same first-step gradient. Along the trajectory, $W_t^\mathrm{ID}$ and $W_t^\mathrm{OOD}$ diverge by at most $O(T\eta B_g)$ (both trajectories have per-step displacement $\le \eta B_g$), so by $L$-Lipschitz continuity of $\nabla\cL_\mathrm{anc}$ and Cauchy--Schwarz, $\bbE\|\bar g_a^{\mathrm{OOD}}\|_F^2 - \bbE\|\bar g_a^{\mathrm{ID}}\|_F^2 = O(T\eta L B_g^2)$. Multiplied by the outer $T\eta$ factor, this is $O(T^2 \eta^2 L B_g^2)$ and is subsumed in the second-order remainder. Hence
\begin{equation*}
  \bbE[\lambda_{\mathrm{ID}} - \lambda_{\mathrm{OOD}}]
  = T\eta\,\bigl(\bbE\langle \bar g_a, \bar g_e^{\mathrm{OOD}}\rangle_F - \bbE\langle \bar g_a, \bar g_e^{\mathrm{ID}}\rangle_F\bigr) + O(T^2\eta^2 L B_g^2),
\end{equation*}
where we have identified $\bar g_a^{\mathrm{ID}} \approx \bar g_a^{\mathrm{OOD}} =: \bar g_a$ to leading order.

Apply Corollary~\ref{cor:feature-orth} to the OOD term:
\begin{equation*}
  \bigl|\bbE\langle \bar g_a, \bar g_e^{\mathrm{OOD}}\rangle_F\bigr| \le \epsilon\, B_\phi^2\, B_C.
\end{equation*}
Apply Proposition~\ref{prop:id-interference} to the ID term:
\begin{equation*}
  \bbE\langle \bar g_a, \bar g_e^{\mathrm{ID}}\rangle_F \le -\gamma.
\end{equation*}
Substituting and using $|a| \le \epsilon B_\phi^2 B_C \Rightarrow -\epsilon B_\phi^2 B_C \le a$,
\begin{align*}
  \bbE[\lambda_{\mathrm{ID}} - \lambda_{\mathrm{OOD}}]
  &\ge T\eta\,\bigl(-\epsilon B_\phi^2 B_C - (-\gamma)\bigr) + O(T^2\eta^2 L B_g^2) \\
  &= T\eta\,\bigl(\gamma - \epsilon B_\phi^2 B_C\bigr) + O(T^2\eta^2 L B_g^2).
\end{align*}
\end{proof}

\paragraph{What the theorem says.} The separation between expected ID and OOD anchor losses is at least $T\eta$ times an architecture-independent gap $\gamma - \epsilon B_\phi^2 B_C$, which is positive whenever the OOD feature orthogonality is strong enough to overwhelm the ID interference constant. The factor $T\eta$ means longer finetuning amplifies the separation linearly (within the small-step regime), while the constant $\gamma$ is a structural property of the ID data and the pretrained classifier and is independent of the shift's geometry. Severity dependence enters through $\epsilon$: Proposition~\ref{prop:prediction-collapse}'s feature collapse drives $\tilde\phi(x_e) \to 0$ (and hence $\epsilon \to 0$) in the far-OOD regime $\|p_1 - p_0\|_{\mathrm{TV}} \to 1$, but $\epsilon$ can be $\Theta(1)$ for near-OOD. The theorem's signed lower bound is therefore only applicable in the far-OOD regime.

\paragraph{When the theorem fails.} Theorem~\ref{thm:main-separation}'s guarantee becomes vacuous when $\epsilon B_\phi^2 B_C \ge \gamma$, i.e., when OOD features retain too much ID-like structure relative to the ID interference strength. In this regime the theorem imposes no signed constraint on the separation, and the observed direction is an empirical matter: the $\mathbf{!}$ flags in Table~\ref{tab:detection} mark the dataset--model pairs where it reverses ($\lambda_{\mathrm{OOD}} > \lambda_{\mathrm{ID}}$ for the majority of seeds). All four such cells are benign shifts, and all four coincide with the cells where the learning-rate stability rule falls back to $\eta = 10^{-4}$.

\subsection{Statistical Validity of the Hypothesis Test}
\label{sec:theory-conformal}

Theorem~\ref{thm:main-separation} establishes a separation between expected anchor losses under ID and OOD distributions. The practical detector of Algorithm~\ref{alg:method} does not observe these expectations directly; it observes $K$ calibration samples $\lambda^{(1)}, \ldots, \lambda^{(K)}$ and a single test statistic $\lambda^*$, and rejects $H_0$ when $\lambda^*$ falls below the empirical $\alpha$-quantile of the calibration set. We now verify that this decision rule controls Type-I error at level $\alpha$ (up to a finite-$K$ correction), via a standard conformal prediction argument.

We first make precise what ``in-distribution'' means for this test. In a continuous feature space the data need not occupy a sharply bounded set, and density typically decays gradually rather than defining a natural in/out boundary.

\begin{definition}[In-distribution evaluation batch]
\label{def:id-batch}
Let $p_0$ denote the data-generating distribution on inputs and labels, and $p_0^X$ its covariate marginal. The support set, the anchor set $(\Xanc, \yanc)$, and the $K$ calibration batches are sampled from $p_0$. Conditional on the fixed support and anchor sets, a deployment evaluation batch is \emph{in-distribution} when
\begin{equation*}
  H_0 :\; \Xeval^* \sim (p_0^X)^{\otimes n_e},
\end{equation*}
that is, when it is drawn from the same covariate distribution as --- and is therefore exchangeable with --- the calibration batches.
\end{definition}

This is a distribution-level definition. One could instead posit a pointwise ID region, for example a density level set $\{x : \mathrm{d}p_0^X/\mathrm{d}\mu\,(x) > c\}$ or a support neighborhood $\{x : d(x, \operatorname{supp} p_0^X) \le r\}$, but both require an arbitrary choice of $c$ or $r$. Note that \cald{} uses neither: the detector thresholds the calibrated statistic $\lambda$, and the validity of Proposition~\ref{prop:conformal} below follows from the exchangeability in Definition~\ref{def:id-batch} alone, independent of dimension, support geometry, or the existence of a density. The feature-separation parameter $\epsilon$ of Theorem~\ref{thm:main-separation} plays a different role: it governs the \emph{power} analysis, not the definition of ID versus OOD --- in the far-OOD limit $\|p_1 - p_0\|_{\mathrm{TV}} \to 1$ the tilting argument predicts $\epsilon \to 0$, while for near-OOD shifts $\epsilon$ may remain $\Theta(1)$, making the power bound vacuous even where the detector has empirical power.

\begin{proposition}[Finite-sample Type-I error control]
\label{prop:conformal}
Consider the calibrated-disagreement-detection test defined in Algorithm~\ref{alg:method}. Assume that the anchor set $(\Xanc, \yanc)$, the support set $\cD^{\mathrm{sup}}$, and all finetuning hyperparameters $(T, \eta)$ are fixed; that the eval batches $\Xeval^{(1)}, \ldots, \Xeval^{(K)}, \Xeval^*$ across rounds are i.i.d.\ draws; and that the finetuning inner loop is deterministic given its inputs. Under the null hypothesis $H_0 : \Xeval^* \sim p_0$ (same distribution as the calibration eval batches), the one-sided test
\begin{equation}
  \mathrm{Reject}\ H_0 \iff \lambda^* < \lambda^{(\lceil \alpha(K+1) \rceil)}
  \label{eq:conformal-test}
\end{equation}
(where $\lambda^{(k)}$ denote the calibration statistics in ascending order) satisfies
\begin{equation}
  \bbP_{H_0}(\mathrm{Reject}) \le \frac{\lceil \alpha(K+1) \rceil}{K+1} \le \alpha + \frac{1}{K+1}.
  \label{eq:conformal-bound}
\end{equation}
\end{proposition}

\begin{proof}
Under $H_0$ and the stated i.i.d.\ conditions, the $K+1$ random variables
\begin{equation*}
  \lambda^{(1)}, \ldots, \lambda^{(K)}, \lambda^*
\end{equation*}
are i.i.d.\ because each $\lambda^{(k)}$ and $\lambda^*$ is a deterministic function of a fresh eval batch $\Xeval^{(k)}$ (resp.\ $\Xeval^*$) applied through the same fixed anchor, support, and finetuning procedure. In particular they are \emph{exchangeable}: the joint law is invariant under any permutation of the indices.

Let $R$ denote the rank of $\lambda^*$ among the $K+1$ statistics when they are sorted in ascending order (breaking ties uniformly at random). By exchangeability, $R$ is uniformly distributed on $\{1, 2, \ldots, K+1\}$. The test rejects exactly when $\lambda^* < \lambda^{(\lceil \alpha(K+1) \rceil)}$, i.e., when $\lambda^*$ is among the $\lceil \alpha(K+1)\rceil - 1$ smallest of the $K + 1$ statistics, i.e., when $R \le \lceil \alpha(K+1) \rceil - 1$. Hence
\begin{equation*}
  \bbP_{H_0}(\mathrm{Reject})
  = \bbP\bigl(R \le \lceil \alpha(K+1) \rceil - 1\bigr)
  = \frac{\lceil \alpha(K+1) \rceil - 1}{K+1}
  \le \frac{\alpha(K+1)}{K+1}
  = \alpha,
\end{equation*}
using $\lceil \alpha(K+1)\rceil \le \alpha(K+1) + 1$ in the inequality. This gives the tight bound $\bbP_{H_0}(\mathrm{Reject}) \le \alpha$. The weaker (but standard) form quoted in~\eqref{eq:conformal-bound}, $\bbP_{H_0}(\mathrm{Reject}) \le \lceil \alpha(K+1)\rceil/(K+1) \le \alpha + 1/(K+1)$, is the discretization version obtained by dropping the $-1$: its slack $1/(K+1)$ quantifies the cost of using a finite calibration set.
\end{proof}

\paragraph{Power analysis.} Proposition~\ref{prop:conformal} controls Type-I error only; power depends on the separation established in Theorem~\ref{thm:main-separation}. 
A power statement requires two ingredients. First, a mean separation: $\bbE[\lambda_{\mathrm{OOD}}] \le \bbE[\lambda_{\mathrm{ID}}] - \Delta_\lambda$ for some $\Delta_\lambda > 0$. Theorem~\ref{thm:main-separation} supplies one, with $\Delta_\lambda \ge T\eta(\gamma - \epsilon B_\phi^2 B_C)$. Second, concentration: $\lambda_{\mathrm{ID}}$ and $\lambda_{\mathrm{OOD}}$ must be sufficiently tight around their means. The batch-gradient structure gives an $O(1/\sqrt{n_e})$ standard deviation under the small-step regime, but a full Hoeffding-type bound would need to be proved separately. Given both ingredients, the detector's power approaches $1$ as $K \to \infty$ for any $\Delta_\lambda$ exceeding the concentration noise. We leave a detailed power analysis as a direction for future work.

\paragraph{Assumption satisfied in the experimental setup.} The assumptions of Proposition~\ref{prop:conformal} match Algorithm~\ref{alg:method}'s protocol: anchor and support are fixed across calibration rounds (only eval batches are resampled), the finetuning inner loop is deterministic given its inputs (full-batch gradient descent with no randomness beyond the initial draw), and the eval batches are i.i.d.\ from $p_0$ during calibration. 

\subsection{Multiclass Case}
\label{sec:theory-multiclass}

All statements in Lemma~\ref{lem:factorization}, Proposition~\ref{prop:first-step}, Corollary~\ref{cor:feature-orth}, and Proposition~\ref{prop:T-step} hold for arbitrary $C \geq 2$: the outer-product structure $\nabla_W \ell = (q - \mathrm{target})\, \phi^\top$ produces a $C$-dimensional first factor but carries through unchanged. The ID-interference argument in Proposition~\ref{prop:id-interference} uses binary structure for concreteness; the extension to $C > 2$ replaces the two-component sign analysis with a joint distribution over label pairs $(y_a, \hat{y}(x_e)) \in [C]^2$. Under a class-symmetry condition ($p_0$ marginally balanced over classes, or $\phi$ covariance block-diagonal across class regions), the label-agreement contribution dominates and gives the same negative expected sign for the cross-term.

\subsection{Distributional Difference vs.\ Predictive Harm}
\label{sec:theory-harm}

The theory developed so far controls the gradient cross-term $\langle g_a, g_e\rangle_F$ via the TV distance $\|p_1 - p_0\|_{\mathrm{TV}}$: an \emph{unsigned} measure of distributional difference. But a detector of \emph{performance deterioration} should fire only on shifts that actually hurt predictive accuracy. As the benign-shift experiments (Adult Income, Diabetes Readmission) show, these notions are not equivalent.

This subsection bridges the gap. We:
\begin{enumerate}[leftmargin=1.5em]
  \item Formally decompose the accuracy gap into a density-ratio integral that makes the sign structure of harm explicit.
  \item Extend Nagler's tilting from features to \emph{predictions}: at far-OOD inputs, not only does the feature collapse, but so does the classifier's output, converging to a constant distribution $q_\infty$.
  \item Propose a second, accuracy-sensitive signal $\hat\mu$ that estimates the accuracy gap $\Delta$ directly (Proposition~\ref{prop:harm-theorem}), with $\hat\lambda$ serving as a shift-confirmation gate.
\end{enumerate}
The additional assumptions required by Proposition~\ref{prop:harm-theorem} are catalogued in Section~\ref{sec:theory-assumptions}.

\subsubsection{Accuracy gap as a signed density-ratio integral}

Let $\ell_{01}(q, x, y) := \indic\{y \neq \arg\max_c q(c \mid x)\}$ be the 0-1 loss and $\mathrm{acc}_q(p) := 1 - \bbE_{(x, y) \sim p}[\ell_{01}(q, x, y)]$. The accuracy gap between $p_0$ and $p_1$ under the fixed (pretrained) classifier $q = q_{W_0}$ is
\begin{equation}
  \Delta(q, p_0, p_1) := \mathrm{acc}_q(p_0) - \mathrm{acc}_q(p_1).
  \label{eq:accuracy-gap-def}
\end{equation}

\begin{lemma}[Density-ratio decomposition of the accuracy gap]
\label{lem:density-ratio}
Suppose $p_0$ and $p_1$ have densities with respect to a common reference measure on $\cX \times \cY$, with $p_0(x, y) > 0$ wherever $p_1(x, y) > 0$, and let $r(x, y) := p_1(x, y)/p_0(x, y)$ denote the density ratio. Then
\begin{equation}
  \Delta(q, p_0, p_1) = \bbE_{p_0}\bigl[(r - 1)\, \ell_{01}\bigr].
  \label{eq:density-ratio-decomp}
\end{equation}
\end{lemma}

\begin{proof}
\emph{Step 1 (rewrite $\Delta$ in loss form).} Since $\mathrm{acc}_q(p) = 1 - \bbE_p[\ell_{01}(q, x, y)]$,
\begin{equation*}
  \Delta = \mathrm{acc}_q(p_0) - \mathrm{acc}_q(p_1)
  = (1 - \bbE_{p_0}[\ell_{01}]) - (1 - \bbE_{p_1}[\ell_{01}])
  = \bbE_{p_1}[\ell_{01}] - \bbE_{p_0}[\ell_{01}].
\end{equation*}

\emph{Step 2 (change of measure).} The support condition $\{p_1 > 0\} \subseteq \{p_0 > 0\}$ ensures the density ratio $r = p_1/p_0$ is well-defined and finite $p_0$-almost everywhere. For any bounded $f: \cX \times \cY \to \bbR$,
\begin{equation*}
  \bbE_{p_1}[f]
  = \int f\, p_1
  = \int f \cdot \frac{p_1}{p_0} \cdot p_0
  = \int f\, r\, p_0
  = \bbE_{p_0}[r\, f].
\end{equation*}
Applied to $f = \ell_{01}$: $\bbE_{p_1}[\ell_{01}] = \bbE_{p_0}[r\, \ell_{01}]$.

\emph{Step 3 (subtract and combine).}
\begin{equation*}
  \Delta
  = \bbE_{p_0}[r\, \ell_{01}] - \bbE_{p_0}[\ell_{01}]
  = \bbE_{p_0}[(r - 1)\, \ell_{01}].
\end{equation*}
\end{proof}

\paragraph{Reading the decomposition.} The gap is the $p_0$-expectation of the 0-1 loss weighted by the \emph{signed} excess density $r - 1$:
\begin{itemize}[leftmargin=1.5em]
  \item On the mistake region $M := \{(x, y) : \ell_{01}(q, x, y) = 1\}$: if $r > 1$ (shift puts more mass there), the integrand $(r-1)\ell_{01}$ is positive, so the contribution to $\Delta$ is positive (i.e., $\mathrm{acc}_{p_1} < \mathrm{acc}_{p_0}$; harmful). Conversely, if $r < 1$ on $M$, the shift pulls mass \emph{out} of the mistake region, the contribution is negative, and $\Delta < 0$ (OOD accuracy exceeds ID accuracy; benign).
  \item On the correct region $M^c$: the integrand vanishes because $\ell_{01} = 0$, so changes in mass on correctly-classified regions do not affect $\Delta$ directly.
\end{itemize}

Thus $\Delta$ is controlled \emph{entirely} by how the shift re-weights the mistake region. In contrast, $\|p_1 - p_0\|_{\mathrm{TV}}$ is controlled by re-weighting of the \emph{entire} domain. Two shifts with the same TV distance can produce very different $\Delta$ values depending on whether they transfer mass into or out of the mistake region.

\begin{corollary}[TV upper bound is loose]
\label{cor:tv-loose}
$|\Delta(q, p_0, p_1)| \le 2\,\|p_1 - p_0\|_{\mathrm{TV}}$, with equality if and only if $\{|r - 1| > 0\} \subseteq M$ up to $p_0$-null sets and $(r-1)\ell_{01}$ has constant sign on $M$ (i.e., the shift is supported on the mistake region and the density ratio is monotone in one direction).
\end{corollary}

\begin{proof}
\emph{Step 1 (absolute-value bound inside expectation).} By Lemma~\ref{lem:density-ratio} and the triangle inequality,
\begin{equation*}
  |\Delta|
  = \bigl|\bbE_{p_0}[(r - 1)\, \ell_{01}]\bigr|
  \le \bbE_{p_0}\bigl[|r - 1|\, \ell_{01}\bigr].
\end{equation*}
Since $\ell_{01}(q, x, y) \in \{0, 1\}$, $|r - 1|\, \ell_{01} \le |r - 1|$ pointwise, hence
\begin{equation*}
  \bbE_{p_0}\bigl[|r - 1|\, \ell_{01}\bigr] \le \bbE_{p_0}[|r - 1|].
\end{equation*}

\emph{Step 2 (identify $\bbE_{p_0}[|r - 1|]$ with TV distance).} The standard density-based formula for total variation distance reads
\begin{equation*}
  \|p_1 - p_0\|_{\mathrm{TV}}
  = \tfrac{1}{2} \int |p_1 - p_0|
  = \tfrac{1}{2} \int |r - 1|\, p_0
  = \tfrac{1}{2}\, \bbE_{p_0}[|r - 1|].
\end{equation*}
Therefore $\bbE_{p_0}[|r - 1|] = 2\, \|p_1 - p_0\|_{\mathrm{TV}}$. Combining with Step 1 gives $|\Delta| \le 2\, \|p_1 - p_0\|_{\mathrm{TV}}$.

\emph{Step 3 (equality conditions).} The two inequalities used above are tight under the following conditions:
\begin{itemize}[leftmargin=1.5em]
  \item $|\bbE[g]| = \bbE[|g|]$ requires $g = (r-1)\ell_{01}$ to have constant sign $p_0$-almost surely.
  \item $|r - 1|\, \ell_{01} = |r - 1|$ $p_0$-almost surely requires $\ell_{01}(q, x, y) = 1$ on $\{|r - 1| > 0\}$, i.e., the shift is supported on the mistake region.
\end{itemize}
Both must hold simultaneously for equality in the TV bound.
\end{proof}

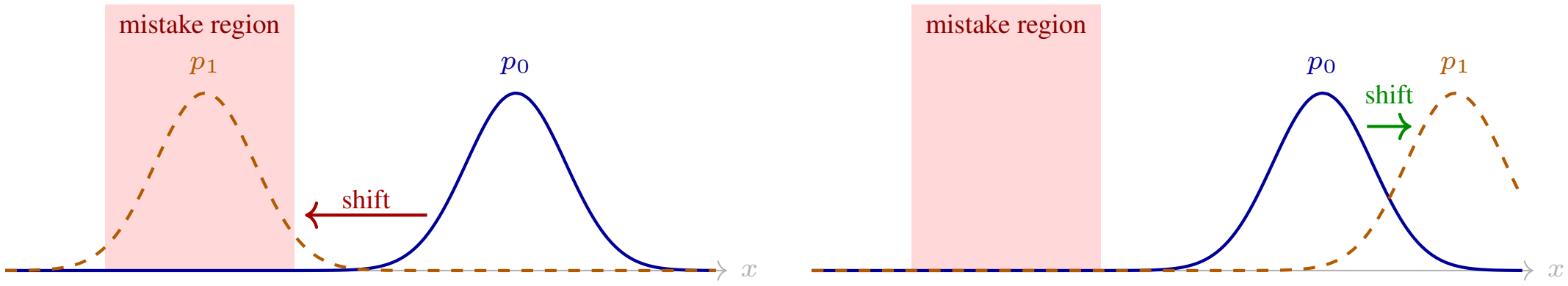
\begin{figure}[t]
\centering
\begin{subfigure}[b]{0.48\textwidth}
\centering
\begin{tikzpicture}[scale=1.0]
  \draw[->, gray!60] (-3.3, 0) -- (3.2, 0) node[right, gray!60, font=\scriptsize] {$x$};
  \fill[red!15] (-2.4, 0) rectangle (-0.7, 2.4);
  \node[red!55!black, font=\scriptsize] at (-1.55, 2.2) {mistake region};
  \draw[thick, blue!60!black, smooth, samples=100, domain=-3.3:3.1]
    plot ({\x}, {1.6 * exp(-(\x-1.3)*(\x-1.3)/0.4)});
  \node[blue!60!black, font=\scriptsize] at (1.3, 1.85) {$p_0$};
  \draw[thick, orange!70!black, dashed, smooth, samples=100, domain=-3.3:3.1]
    plot ({\x}, {1.6 * exp(-(\x-(-1.5))*(\x-(-1.5))/0.4)});
  \node[orange!70!black, font=\scriptsize] at (-1.5, 1.85) {$p_1$};
  \draw[->, thick, red!65!black] (0.5, 0.5) -- (-0.6, 0.5);
  \node[font=\scriptsize, red!65!black, anchor=north] at (-0.05, 0.85) {shift};
\end{tikzpicture}
\caption{\textbf{Harmful shift.} $p_1$ moves mass into the classifier's mistake region (shaded), so the accuracy gap $\Delta > 0$ and accuracy drops on the shifted distribution.}
\end{subfigure}
\hfill
\begin{subfigure}[b]{0.48\textwidth}
\centering
\begin{tikzpicture}[scale=1.0]
  \draw[->, gray!60] (-3.3, 0) -- (3.2, 0) node[right, gray!60, font=\scriptsize] {$x$};
  \fill[red!15] (-2.4, 0) rectangle (-0.7, 2.4);
  \node[red!55!black, font=\scriptsize] at (-1.55, 2.2) {mistake region};
  \draw[thick, blue!60!black, smooth, samples=100, domain=-3.3:3.1]
    plot ({\x}, {1.6 * exp(-(\x-1.3)*(\x-1.3)/0.4)});
  \node[blue!60!black, font=\scriptsize] at (1.3, 1.85) {$p_0$};
  \draw[thick, orange!70!black, dashed, smooth, samples=100, domain=-3.3:3.1]
    plot ({\x}, {1.6 * exp(-(\x-2.5)*(\x-2.5)/0.4)});
  \node[orange!70!black, font=\scriptsize] at (2.5, 1.85) {$p_1$};
  \draw[->, thick, green!55!black] (1.7, 1.3) -- (2.1, 1.3);
  \node[font=\scriptsize, green!55!black, anchor=north] at (1.9, 1.8) {shift};
\end{tikzpicture}
\caption{\textbf{Benign shift.} $p_1$ moves mass away from the mistake region (shaded) into the correct region, so $\Delta \le 0$ and accuracy is preserved or improved.}
\end{subfigure}
\caption{\textbf{Same TV distance, opposite signs of $\Delta$.} Both panels show a shift $p_0 \rightarrow p_1$ of the same magnitude (same $\|p_1 - p_0\|_{\mathrm{TV}}$) under the same classifier, which has a mistake region on the left of the axis (shaded red). In \textbf{(a)}, the shift transfers mass into the mistake region, so accuracy drops. In \textbf{(b)}, the shift transfers mass away from the mistake region, so accuracy is preserved or improved. The anchor-loss detector $\hat\lambda$ (Section~\ref{sec:theory-nagler}) has the same value in both cases; it is insensitive to the sign of $\Delta$. An additional signal is needed.}
\label{fig:harmful-vs-benign}
\end{figure}

Corollary~\ref{cor:tv-loose} formalizes the central tension: TV distance upper-bounds $|\Delta|$ but the bound is rarely tight on benchmark data, because shifts on real-world data rarely align perfectly with the classifier's mistake region. Figure~\ref{fig:harmful-vs-benign} shows two shifts with similar TV distances but opposite signs of $\Delta$.

\subsubsection{Far-OOD prediction collapse}

Nagler's tilting argument (Section~\ref{sec:theory-nagler}) established that far-OOD features collapse to a single input-independent direction $\phi_\infty$. We now observe that this extends to the classifier's output: all far-OOD queries receive the \emph{same} prediction $q_\infty$.

\begin{assumption}[Far-OOD regime]
\label{assn:far-ood}
The eval distribution $p_1$ satisfies $\|p_1 - p_0\|_{\mathrm{TV}} \to 1$ in the sense that, for $p_1$-almost every $x_e$, the tilt $\exp(v(x_e)^\top W_q^{(h)} s)$ is approximately constant over $s \sim p_0$ for every head $h$.
\end{assumption}

We acknowledge that this is the most aggressive assumption in this derivation and it is therefore best read as an idealization; it is defensible only for one-layer attention (\citet{nagler2023statistical}'s tilting theorem assumes a single layer, while TabPFN and TabICL are multi-layer), and only in the strict limit $\|p_1 - p_0\|_{\mathrm{TV}} \to 1$, while typical benchmark shifts have TV distance bounded well away from $1$. The conclusion of Proposition~\ref{prop:prediction-collapse} below, that \emph{all} far-OOD queries receive the same prediction $q_\infty$, should accordingly be read as the limit point of a trend rather than a description of any specific experiment. We note that the subsequent separation theorem (Theorem~\ref{thm:main-separation}) does not depend on this assumption; only the harm theorem (Proposition~\ref{prop:harm-theorem}) does.

\begin{proposition}[Far-OOD prediction collapse]
\label{prop:prediction-collapse}
Under Assumption~\ref{assn:far-ood}, the PFN's prediction collapses pointwise: for $p_1$-almost every $x_e$,
\begin{equation}
  q_{W_0}(\cdot \mid x_e) \longrightarrow q_\infty := \mathrm{softmax}(W_0 \phi_\infty)
  \label{eq:prediction-collapse}
\end{equation}
in total variation, where $\phi_\infty := \sum_h W_v^{(h)} \bbE_{V \sim p_0}[V]$ is the input-independent direction defined in~\eqref{eq:phi-infinity}; equivalently, $q_\infty$ is the same constant prediction for every far-OOD query.
\end{proposition}

\begin{proof}
We show successively that (i) the tilted measures at each head approach $p_0$ in total variation, (ii) the tilted mean features approach $\bbE_{p_0}[V]$ in norm, (iii) the backbone feature approaches $\phi_\infty$, and (iv) the softmax output approaches $q_\infty$ in total variation.

\emph{Step 1 (tilted measure converges to $p_0$).} Assumption~\ref{assn:far-ood} posits a sequence $\epsilon_n \to 0$ and a set of $x_e$'s of $p_1$-mass $\to 1$ such that, for every head $h$,
\begin{equation*}
  \sup_{s \in \mathrm{supp}(p_0)} \bigl|\psi_h(s; x_e) - Z_h(x_e)\bigr| \le \epsilon_n,
\end{equation*}
where $\psi_h(s; x) := \exp(v(x)^\top W_q^{(h)} s)$ and $Z_h(x) := \bbE_{V \sim p_0}[\psi_h(V; x)]$ is the partition function. Using this uniform bound,
\begin{equation*}
  \bigl\|g_h(\cdot \mid x_e) - p_0\bigr\|_{\mathrm{TV}}
  = \tfrac{1}{2}\, \bbE_{V \sim p_0}\left[\left|\frac{\psi_h(V; x_e)}{Z_h(x_e)} - 1\right|\right]
  \le \frac{\epsilon_n}{2 Z_h(x_e)}.
\end{equation*}

\emph{Step 2 (tilted mean feature converges).} For any bounded $V$ (say $\|V\| \le B_s$),
\begin{equation*}
  \bigl\|\bbE_{V \sim g_h(\cdot \mid x_e)}[V] \,-\, \bbE_{V \sim p_0}[V]\bigr\|
  \le 2 B_s \cdot \bigl\|g_h(\cdot \mid x_e) - p_0\bigr\|_{\mathrm{TV}}
  \le \frac{B_s\, \epsilon_n}{Z_h(x_e)}.
\end{equation*}

\emph{Step 3 (feature converges to $\phi_\infty$).} Substituting Step 2 into the limiting-feature expression~\eqref{eq:tilted-feature},
\begin{equation*}
  \bigl\|\bar{u}(x_e) - \phi_\infty\bigr\|
  = \Bigl\|\sum_{h=1}^H W_v^{(h)}\bigl(\bbE_{g_h}[V] - \bbE_{p_0}[V]\bigr)\Bigr\|
  \le \sum_{h=1}^H \|W_v^{(h)}\|\, \frac{B_s\, \epsilon_n}{Z_h(x_e)}
  =: C_\phi(x_e)\, \epsilon_n,
\end{equation*}
where $C_\phi(x_e) := B_s \sum_h \|W_v^{(h)}\| / Z_h(x_e) < \infty$ for $p_1$-a.e. $x_e$ (the partition functions are bounded away from zero on the assumed far-OOD set).

\emph{Step 4 (softmax is Lipschitz).} The softmax map $z \mapsto \mathrm{softmax}(z)$ is $1$-Lipschitz from $(\bbR^C, \|\cdot\|_2)$ to the probability simplex under $\|\cdot\|_2$ (a standard consequence of the Jacobian bound $\|J(q)\|_{\mathrm{op}} \le 1/2$ used in Section~\ref{sec:theory-tstep}, which gives a tighter constant $1/2$ for differentiable paths; the $1$-Lipschitz statement suffices here):
\begin{equation*}
  \bigl\|\mathrm{softmax}(z) - \mathrm{softmax}(z')\bigr\|_2 \le \|z - z'\|_2.
\end{equation*}
Converting to the $\ell_1$ norm used by TV distance via $\|\cdot\|_1 \le \sqrt{C}\,\|\cdot\|_2$,
\begin{equation*}
  \bigl\|\mathrm{softmax}(z) - \mathrm{softmax}(z')\bigr\|_1 \le \sqrt{C}\,\|z - z'\|_2.
\end{equation*}
Applied to $z = W_0 \bar u(x_e)$ and $z' = W_0 \phi_\infty$, and using $\|z - z'\|_2 \le \|W_0\|_{\mathrm{op}} \|\bar u(x_e) - \phi_\infty\|_2$,
\begin{align*}
  \bigl\|q_{W_0}(\cdot \mid x_e) - q_\infty\bigr\|_{\mathrm{TV}}
  &\;=\; \tfrac{1}{2}\,\bigl\|\mathrm{softmax}(W_0 \bar u(x_e)) - \mathrm{softmax}(W_0 \phi_\infty)\bigr\|_1 \\
  &\;\le\; \tfrac{\sqrt{C}}{2}\,\|W_0\|_{\mathrm{op}}\, C_\phi(x_e)\, \epsilon_n.
\end{align*}

As $\epsilon_n \to 0$ (Assumption~\ref{assn:far-ood}), the right-hand side vanishes $p_1$-almost surely, proving $q_{W_0}(\cdot \mid x_e) \to q_\infty$ in total variation.
\end{proof}

\begin{corollary}[Far-OOD accuracy is constant-predictor accuracy]
\label{cor:ood-acc}
Under Assumption~\ref{assn:far-ood},
\begin{equation}
  \mathrm{acc}_{q_{W_0}}(p_1) \longrightarrow \mathrm{acc}_{q_\infty}(p_1) = p_{Y \mid p_1}(\hat y_\infty),
\end{equation}
where $\hat y_\infty := \arg\max_c q_\infty(c)$ is the constant prediction at $\phi_\infty$ and $p_{Y \mid p_1}$ is the label marginal under $p_1$.
\end{corollary}

\begin{proof}
Write the accuracy as an expectation:
\begin{equation*}
  \mathrm{acc}_q(p_1) = \bbP_{(x, y) \sim p_1}\left(\arg\max_c q(c \mid x) = y\right).
\end{equation*}

\emph{Step 1 (arg-max converges).} Proposition~\ref{prop:prediction-collapse} gives $q_{W_0}(\cdot \mid x_e) \to q_\infty$ in TV for $p_1$-almost every $x_e$. We assume that $q_\infty$ has a unique maximizer, so the arg-max function is continuous at $q_\infty$; ties, if they occur, can be broken by an arbitrary fixed rule without changing the statement. Hence
\begin{equation*}
  \arg\max_c q_{W_0}(c \mid x_e) \longrightarrow \arg\max_c q_\infty(c) = \hat y_\infty
\end{equation*}
for $p_1$-almost every $x_e$.

\emph{Step 2 (accuracy difference is bounded by mismatch probability).} By the triangle inequality applied to $\indic\{\arg\max q_{W_0}(\cdot \mid x) = y\}$ versus $\indic\{\hat y_\infty = y\}$:
\begin{align*}
  \bigl|\mathrm{acc}_{q_{W_0}}(p_1) - \mathrm{acc}_{q_\infty}(p_1)\bigr|
  &= \Bigl|\bbP_{p_1}(\arg\max q_{W_0} = y) - \bbP_{p_1}(\hat y_\infty = y)\Bigr| \\
  &\le \bbP_{p_1}\left(\arg\max_c q_{W_0}(c \mid x) \neq \hat y_\infty\right).
\end{align*}
By Step 1, the right-hand side tends to zero.

\emph{Step 3 (evaluate the limit).} Since $\hat y_\infty$ is a \emph{constant} independent of $x$,
\begin{equation*}
  \mathrm{acc}_{q_\infty}(p_1)
  = \bbP_{(x, y) \sim p_1}(\hat y_\infty = y)
  = \bbP_{y \sim p_{Y \mid p_1}}(y = \hat y_\infty)
  = p_{Y \mid p_1}(\hat y_\infty).
\end{equation*}
\end{proof}

Far-OOD accuracy equals the probability that the constant prediction $\hat y_\infty$ matches the true OOD label. Whether this is harmful depends on the alignment between $\hat y_\infty$ and the OOD label marginal:
\begin{itemize}[leftmargin=1.5em]
  \item \emph{Harmful shift}: $\hat y_\infty$ poorly aligned with $p_{Y \mid p_1}$, so $\mathrm{acc}_{q_\infty}(p_1)$ is low (e.g., Covertype: the classifier predicts a near-uniform distribution over forest types, but OOD elevation concentrates on subalpine species; mismatch).
  \item \emph{Benign shift}: $\hat y_\infty$ happens to align with $p_{Y \mid p_1}$, so $\mathrm{acc}_{q_\infty}(p_1)$ equals or exceeds ID accuracy (e.g., Adult Income: female subgroup has lower base rate of high income, so predicting `low income' as a constant happens to be more accurate than the ID-calibrated conditional prediction).
\end{itemize}

This gives us a handle on harm that is invisible to the TV-distance-based bound of Corollary~\ref{cor:tv-loose}.

\subsubsection{The two-signal decomposition}

We combine the feature-collapse signal (our original $\lambda$) with a prediction-collapse signal that exploits Proposition~\ref{prop:prediction-collapse}.

\begin{definition}[Two-signal quantities]
\label{def:two-signal}
Define the population quantities
\begin{align*}
  \lambda &:= \bbE[\lambda_{\mathrm{ID}}] - \bbE[\lambda_{\mathrm{OOD}}], \qquad \text{(anchor-loss drop under eval batches; Proposition~\ref{prop:first-step})} \\
  \mu &:= \bbE_{(x, y) \sim p_0}[p_x(y)] \,-\, \bbE_{x \sim p_1}[p_x(\hat y(x))], \qquad \text{(mean confidence drop from anchor to eval).}
\end{align*}
The sample estimators $\hat\lambda$, $\hat\mu$ replace each population expectation with its empirical average over the relevant batch. Recall $\hat y(x) = \arg\max_c p_x(c)$, so $p_x(\hat y(x)) = \max_c p_x(c)$ is the maximum softmax probability (MSP) at $x$.
\end{definition}

\paragraph{Roles of the two signals.} $\lambda$ tracks distributional difference (Proposition~\ref{prop:first-step}, Corollary~\ref{cor:feature-orth}): it is large when features shift, regardless of whether the shift hurts. $\mu$ tracks predictive confusion: when the shift pushes eval points onto $\phi_\infty$, each eval MSP collapses to the constant $q_\infty(\hat y_\infty)$, and $\mu$ grows proportionally to the gap between ID accuracy and this constant.

\begin{assumption}[ID calibration]
\label{assn:calibration}
The initial classifier is approximately calibrated on $p_0$:
\begin{equation*}
  \bigl|\bbE_{(x, y) \sim p_0}[p_x(y)] - \mathrm{acc}_{q_{W_0}}(p_0)\bigr| \le \mathrm{cal}_0,
  \qquad
  \bigl|\bbE_{x \sim p_0}[\max_c p_x(c)] - \mathrm{acc}_{q_{W_0}}(p_0)\bigr| \le \mathrm{cal}_0',
\end{equation*}
for small constants $\mathrm{cal}_0, \mathrm{cal}_0' \ll 1$.
\end{assumption}

The first inequality is standard expected-calibration-error on the true-label posterior; the second says the classifier's mean MSP on ID is close to its ID accuracy. Both hold for well-calibrated classifiers.

\begin{assumption}[OOD calibration at the collapse point]
\label{assn:ood-calibration}
The limiting prediction $q_\infty$ and the OOD label marginal $p_{Y \mid p_1}$ agree at the collapse class $\hat y_\infty$:
\begin{equation*}
  \bigl|q_\infty(\hat y_\infty) - p_{Y \mid p_1}(\hat y_\infty)\bigr| \le \mathrm{cal}_\infty.
\end{equation*}
\end{assumption}

This is weaker than full calibration on $p_1$ (which is unavailable without OOD labels): it only constrains the classifier's confidence in its constant prediction to match the frequency of that class under $p_1$. We treat $\mathrm{cal}_\infty$ as a structural miscalibration that is dataset-dependent but bounded.

\begin{assumption}[Label marginal on OOD]
\label{assn:ood-marginal}
The OOD label marginal is bounded away from a point mass: $\max_c p_{Y \mid p_1}(c) < 1$.
\end{assumption}

\paragraph{Mixture model for partial shifts.} To state the theorem, we model partial shifts as two-component mixtures. Let $p_1 = (1-\alpha)\, p_0 + \alpha\, p_\star$, where $\alpha \in [0, 1]$ is the OOD fraction and $p_\star$ is an auxiliary far-OOD distribution satisfying Assumption~\ref{assn:far-ood}. Pure ID is $\alpha = 0$, pure OOD is $\alpha = 1$, and partial shifts are $\alpha \in (0, 1)$. This parametrization captures a large class of distribution shifts and is convenient for tracking linear dependence on shift magnitude.

\begin{proposition}[Harm characterization]
\label{prop:harm-theorem}
Under Assumptions~\ref{assn:far-ood}--\ref{assn:ood-marginal} and the mixture model $p_1 = (1-\alpha) p_0 + \alpha p_\star$, the mean confidence drop tracks the accuracy gap:
\begin{equation}
  \mu = \Delta(q_{W_0}, p_0, p_1) + O\bigl(\alpha\, \mathrm{cal}_\infty \,+\, \mathrm{cal}_0 + \mathrm{cal}_0' \,+\, o(1)\bigr).
  \label{eq:harm-theorem}
\end{equation}
In particular, if $\mathrm{cal}_0, \mathrm{cal}_0', \mathrm{cal}_\infty \to 0$, then $\mu \to \Delta$.
\end{proposition}

\begin{proof}
We compute $\mu$ and $\Delta$ separately for the mixture and compare.

\emph{Step 1 (anchor term of $\mu$).} By Assumption~\ref{assn:calibration} (first inequality),
\begin{equation}
  \bbE_{(x, y) \sim p_0}[p_x(y)] = \mathrm{acc}_{q_{W_0}}(p_0) + O(\mathrm{cal}_0).
  \label{eq:anchor-term}
\end{equation}

\emph{Step 2 (eval term of $\mu$, ID component).} Since $p_x(\hat y(x)) = \max_c p_x(c)$, the mixture gives
\begin{equation*}
  \bbE_{x \sim p_1}[\max_c p_x(c)]
  = (1-\alpha)\, \bbE_{x \sim p_0}[\max_c p_x(c)]
  + \alpha\, \bbE_{x \sim p_\star}[\max_c p_x(c)].
\end{equation*}
By Assumption~\ref{assn:calibration} (second inequality),
\begin{equation}
  \bbE_{x \sim p_0}[\max_c p_x(c)] = \mathrm{acc}_{q_{W_0}}(p_0) + O(\mathrm{cal}_0').
  \label{eq:eval-id-term}
\end{equation}

\emph{Step 3 (eval term of $\mu$, OOD component).} Under Assumption~\ref{assn:far-ood}, Proposition~\ref{prop:prediction-collapse} gives $q_{W_0}(\cdot \mid x) \to q_\infty$ in total variation for $p_\star$-a.e. $x$, so $\max_c q_{W_0}(c \mid x) \to q_\infty(\hat y_\infty)$. Since $\max_c q_{W_0}(c \mid x) \in [0, 1]$ is uniformly bounded, dominated convergence yields
\begin{equation}
  \bbE_{x \sim p_\star}[\max_c p_x(c)] = q_\infty(\hat y_\infty) + o(1).
  \label{eq:eval-ood-term}
\end{equation}

\emph{Step 4 (combine eval terms).} Substituting~\eqref{eq:eval-id-term} and~\eqref{eq:eval-ood-term},
\begin{equation*}
  \bbE_{x \sim p_1}[\max_c p_x(c)]
  = (1 - \alpha)\bigl[\mathrm{acc}(p_0) + O(\mathrm{cal}_0')\bigr]
  + \alpha\bigl[q_\infty(\hat y_\infty) + o(1)\bigr].
\end{equation*}

\emph{Step 5 (compute $\mu$).} Subtracting from~\eqref{eq:anchor-term},
\begin{align}
  \mu
  &= \bbE_{p_0}[p_x(y)] - \bbE_{p_1}[\max_c p_x(c)] \nonumber \\
  &= \bigl[\mathrm{acc}(p_0) + O(\mathrm{cal}_0)\bigr] - \bigl[(1-\alpha)(\mathrm{acc}(p_0) + O(\mathrm{cal}_0')) + \alpha(q_\infty(\hat y_\infty) + o(1))\bigr] \nonumber \\
  &= \alpha\,\bigl(\mathrm{acc}(p_0) - q_\infty(\hat y_\infty)\bigr) + O(\mathrm{cal}_0 + \mathrm{cal}_0') + \alpha\, o(1).
  \label{eq:mu-explicit}
\end{align}

\emph{Step 6 (compute $\Delta$).} Corollary~\ref{cor:ood-acc} applied to $p_\star$ gives $\mathrm{acc}_{q_{W_0}}(p_\star) = p_{Y \mid p_\star}(\hat y_\infty) + o(1)$. For the mixture,
\begin{equation*}
  \mathrm{acc}_{q_{W_0}}(p_1)
  = (1 - \alpha)\, \mathrm{acc}_{q_{W_0}}(p_0) + \alpha\, \mathrm{acc}_{q_{W_0}}(p_\star)
  = (1 - \alpha)\, \mathrm{acc}(p_0) + \alpha\, p_{Y \mid p_\star}(\hat y_\infty) + o(1),
\end{equation*}
hence
\begin{equation}
  \Delta = \mathrm{acc}(p_0) - \mathrm{acc}_{q_{W_0}}(p_1) = \alpha\bigl(\mathrm{acc}(p_0) - p_{Y \mid p_\star}(\hat y_\infty)\bigr) + o(1).
  \label{eq:delta-explicit}
\end{equation}

\emph{Step 7 (compare).} Subtract~\eqref{eq:delta-explicit} from~\eqref{eq:mu-explicit}:
\begin{equation*}
  \mu - \Delta
  = \alpha\,\bigl(p_{Y \mid p_\star}(\hat y_\infty) - q_\infty(\hat y_\infty)\bigr)
  + O(\mathrm{cal}_0 + \mathrm{cal}_0') + o(1).
\end{equation*}
By Assumption~\ref{assn:ood-calibration} applied to $p_\star$, $|p_{Y \mid p_\star}(\hat y_\infty) - q_\infty(\hat y_\infty)| \le \mathrm{cal}_\infty$, so
\begin{equation*}
  |\mu - \Delta| \le \alpha\, \mathrm{cal}_\infty + O(\mathrm{cal}_0 + \mathrm{cal}_0') + o(1),
\end{equation*}
which is~\eqref{eq:harm-theorem}.
\end{proof}

\paragraph{Sample-version corollary.} The sample estimator $\hat\mu$ differs from the population $\mu$ by a Hoeffding-style concentration term (both summands take values in $[0, 1]$):
\begin{equation*}
  |\hat\mu - \mu| = O\bigl(1/\sqrt{n_a} + 1/\sqrt{n_e}\bigr)
\end{equation*}
with high probability. Combining with Proposition~\ref{prop:harm-theorem},
\begin{equation}
  \hat\mu = \Delta + O\bigl(\alpha\, \mathrm{cal}_\infty \,+\, \mathrm{cal}_0 + \mathrm{cal}_0' \,+\, 1/\sqrt{n}\bigr),
\end{equation}
where $n = \min(n_a, n_e)$.

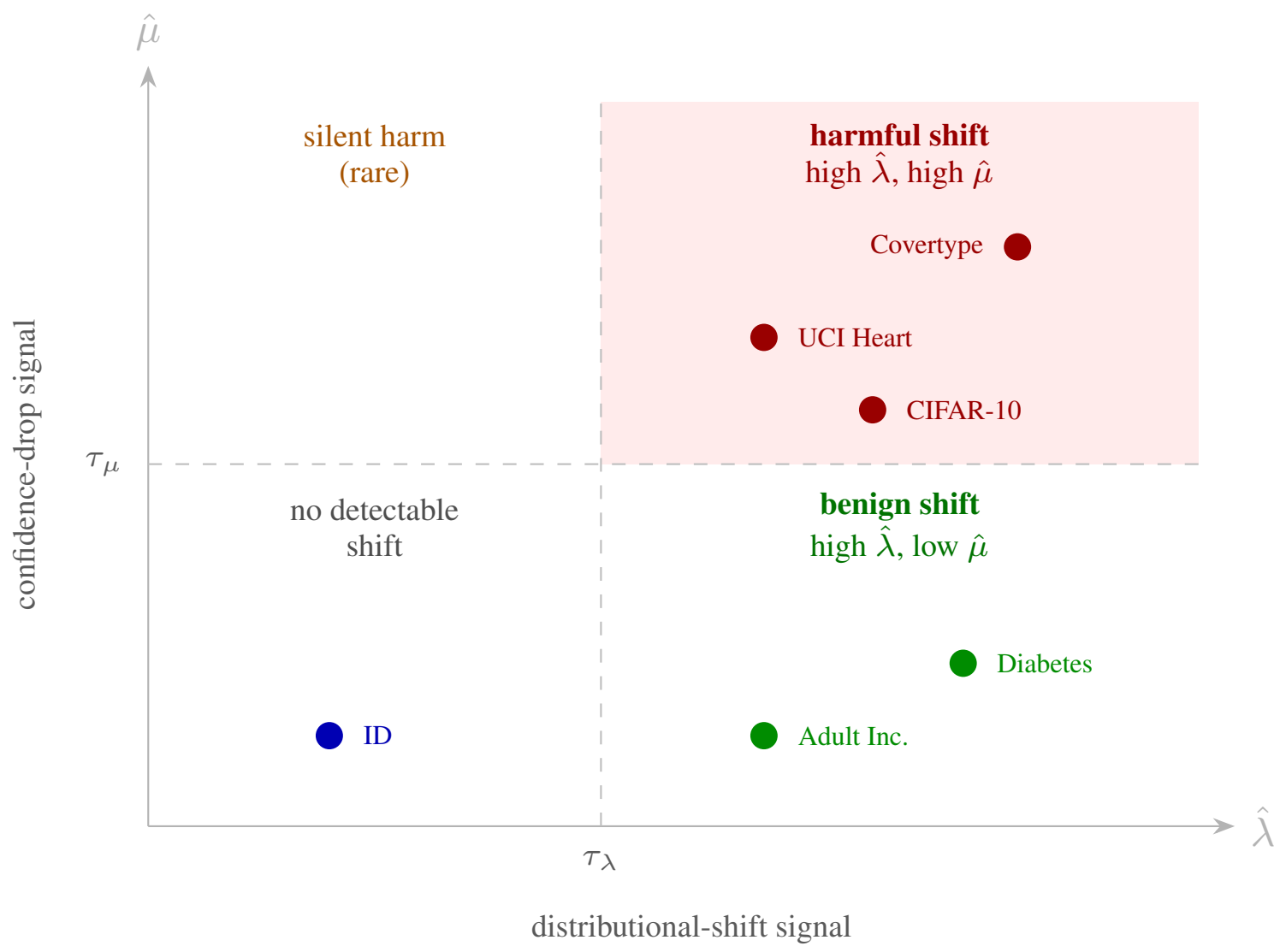
\begin{figure}[t]
\centering
\begin{tikzpicture}[scale=1.4, >={Stealth[scale=1.0]}]
  \draw[->, gray!60] (0, 0) -- (6.0, 0) node[right, font=\small] {$\hat\lambda$};
  \draw[->, gray!60] (0, 0) -- (0, 4.2) node[above, font=\small] {$\hat\mu$};
  \node[font=\scriptsize, gray!70!black, anchor=north] at (3.0, -0.4) {distributional-shift signal};
  \node[font=\scriptsize, gray!70!black, rotate=90, anchor=south] at (-0.5, 2.0) {confidence-drop signal};
  \fill[red!8] (2.5, 2.0) rectangle (5.8, 4.0);
  \draw[gray!50, dashed] (2.5, 0) -- (2.5, 4.0);
  \draw[gray!50, dashed] (0, 2.0) -- (5.8, 2.0);
  \node[font=\scriptsize, gray!70!black, anchor=north] at (2.5, -0.05) {$\tau_\lambda$};
  \node[font=\scriptsize, gray!70!black, anchor=east] at (-0.05, 2.0) {$\tau_\mu$};
  \node[font=\scriptsize, align=center, text=gray!60!black] at (1.25, 1.65) {no detectable\\ shift};
  \node[font=\scriptsize, align=center, text=green!45!black] at (4.15, 1.65) {\textbf{benign shift}\\ high $\hat\lambda$, low $\hat\mu$};
  \node[font=\scriptsize, align=center, text=red!60!black] at (4.15, 3.7) {\textbf{harmful shift}\\ high $\hat\lambda$, high $\hat\mu$};
  \node[font=\scriptsize, align=center, text=orange!65!black] at (1.25, 3.7) {silent harm\\ (rare)};
  \filldraw[blue!70!black] (1.0, 0.5) circle (2pt);
  \node[font=\tiny, blue!70!black, anchor=west] at (1.1, 0.5) {ID};
  \filldraw[green!55!black] (3.4, 0.5) circle (2pt);
  \node[font=\tiny, green!55!black, anchor=west] at (3.5, 0.5) {Adult Inc.};
  \filldraw[green!55!black] (4.5, 0.9) circle (2pt);
  \node[font=\tiny, green!55!black, anchor=west] at (4.6, 0.9) {Diabetes};
  \filldraw[red!60!black] (4.8, 3.2) circle (2pt);
  \node[font=\tiny, red!60!black, anchor=east] at (4.7, 3.2) {Covertype};
  \filldraw[red!60!black] (3.4, 2.7) circle (2pt);
  \node[font=\tiny, red!60!black, anchor=west] at (3.5, 2.7) {UCI Heart};
  \filldraw[red!60!black] (4.0, 2.3) circle (2pt);
  \node[font=\tiny, red!60!black, anchor=west] at (4.1, 2.3) {CIFAR-10};
\end{tikzpicture}
\caption{\textbf{Two-signal detector space.} Each dataset is a point in the $(\hat\lambda, \hat\mu)$ plane. The current $\hat\lambda$-only detector uses only the horizontal axis and cannot distinguish benign shifts (right-lower) from harmful shifts (upper-right). By Proposition~\ref{prop:harm-theorem}, $\hat\mu$ alone estimates $\Delta$ up to calibration errors, so the vertical axis is the harm-magnitude estimator. Gating on $\hat\lambda > \tau_\lambda$ ensures that a distributional shift is actually present before trusting $\hat\mu$ (robustness gate); the conjunction $\hat\lambda > \tau_\lambda \wedge \hat\mu > \tau_\mu$ selects the upper-right region (red box). The upper-left ``silent harm'' quadrant is theoretically possible but rare in practice, because Nagler's tilting links feature shift to prediction confusion.}
\label{fig:two-signal-detector}
\end{figure}

\paragraph{The role of $\hat\lambda$.} Proposition~\ref{prop:harm-theorem} establishes that $\hat\mu$ \emph{alone} estimates $\Delta$ up to calibration and sampling errors. The anchor-loss signal $\hat\lambda$ does not appear in the statement. Its role in the paper is auxiliary but practically important:
\begin{itemize}[leftmargin=1.5em]
  \item \emph{Robustness gate.} When $\hat\mu$ is small, it could mean either no harm ($\Delta \approx 0$) or severe miscalibration (the confidence-based estimator is unreliable). Checking $\hat\lambda > \tau_\lambda$ confirms that a distributional shift is present and therefore that $\hat\mu$ is being estimated in a non-trivial regime.
  \item \emph{Calibration-free fallback.} $\hat\lambda$ requires no calibration assumptions; it fires under Assumption~\ref{assn:feat-orth} alone. If Assumptions~\ref{assn:calibration}--\ref{assn:ood-calibration} are violated, $\hat\lambda$ can still detect distributional shift, though not necessarily harm.
  \item \emph{Two-signal operating point.} A practical harm detector declares ``harmful shift'' when both $\hat\lambda > \tau_\lambda$ and $\hat\mu > \tau_\mu$ (Figure~\ref{fig:two-signal-detector}). The upper-right quadrant of the $(\hat\lambda, \hat\mu)$ plane is the high-precision region; the lower-right contains benign shifts; the upper-left corresponds to miscalibrated estimators or rare silent-harm cases.
\end{itemize}

\begin{definition}[Composed two-signal detector]
\label{def:composed-detector}
The composed detector alerts if and only if the conformal $\lambda$-test of Proposition~\ref{prop:conformal} rejects \emph{and} $\hat\mu > \tau_\mu$.
\end{definition}

\begin{remark}[False-alarm containment under the null]
\label{rem:fpr-containment}
Under $H_0$ the composed detector alerts only when both tests alert, so its false-alarm rate is at most the smaller of the two individual rates --- for arbitrary dependence between $\hat\lambda$ and $\hat\mu$, and with no multiple-testing correction. In particular, the conformal bound $\alpha + 1/(K{+}1)$ of Proposition~\ref{prop:conformal} carries over to the composed detector unchanged.
\end{remark}

\begin{remark}[The gate never adds alarms]
\label{rem:gate-monotone}
Under a shifted distribution the deployment and calibration rounds are no longer exchangeable, so no absolute false-alarm bound is possible. However, the $\hat\mu$ gate can only remove alerts produced by the $\lambda$-test, never add them; the composed detector's benign false-alarm rate is therefore bounded by that of the $\lambda$-test under \emph{every} distribution.
\end{remark}

We evaluate the composed detector in Appendix~\ref{sec:abl-twosignal}; $\lambda$ remains the primary, conformally certified statistic.

\paragraph{Visualizing the two-signal plane.} Proposition~\ref{prop:harm-theorem} predicts that $\hat\mu \approx \Delta$ up to calibration and sampling errors, and that harmful shifts populate a specific region of the $(\hat\lambda, \hat\mu)$ plane (Figure~\ref{fig:two-signal-detector}).

\begin{table}[t]
\centering
\small
\caption{Summary of assumptions used in the theoretical argument.}
\label{tab:assumptions}
\begin{tabular}{p{0.18\linewidth} p{0.47\linewidth} p{0.27\linewidth}}
\toprule
Label & Statement & Source \\
\midrule
\multicolumn{3}{l}{\emph{Core gradient-geometry assumptions (Theorem~\ref{thm:main-separation})}} \\
\midrule
A1 (design) & Only the last-layer weight $W$ is updated; backbone $\phi$ is frozen. & Analysis restriction; the deployed detector updates the full decoder (Appendix~\ref{sec:appendix-design}). \\
A2 (design) & Pseudolabels $\hat{y}$ are fixed at $W_0$ and do not change during finetuning. & Pseudolabel step of Algorithm~\ref{alg:method}. \\
A3 (standard) & $\|\phi(x)\| \le B_\phi$ and softmax residuals bounded. & Standard for trained models. \\
Assn.~\ref{assn:feat-orth} (key) & $|\langle \phi(x_a), \phi(x_e)\rangle| \le \epsilon \|\phi(x_a)\|\|\phi(x_e)\|$ for $x_e \sim p_1$. & Consequence of \citet[Thm.~6.3]{nagler2023statistical} in the far-OOD limit; genuine empirical claim for near-OOD. \\
Assn.~\ref{assn:id-confidence} & Anchor correctness + eval confidence + centered, class-symmetric features + ID accuracy $\ge 1 - \tau$. & Holds for any useful PFN on ID data. \\
Assn.~\ref{assn:small-step} & $T \eta L \le 1/2$ for $L$ the smoothness constant of $\cL$. & Verified numerically in Section~\ref{sec:theory-tstep}. \\
\midrule
\multicolumn{3}{l}{\emph{Harm-characterization assumptions (Proposition~\ref{prop:harm-theorem})}} \\
\midrule
Assn.~\ref{assn:far-ood} & Far-OOD regime: the tilt $\psi_h(s; x_e)$ is approximately constant over $s \sim p_0$ for $p_1$-a.e.\ $x_e$. & Consequence of Nagler's tilting in the TV-limit $\|p_1 - p_0\|_{\mathrm{TV}} \to 1$. \\
Assn.~\ref{assn:calibration} & ID calibration: mean true-label confidence $\bbE_{p_0}[p_x(y)]$ and mean MSP $\bbE_{p_0}[\max_c p_x(c)]$ both track ID accuracy up to small errors $\mathrm{cal}_0, \mathrm{cal}_0'$. & Holds for well-calibrated classifiers; measurable via held-out split. \\
Assn.~\ref{assn:ood-calibration} (key) & OOD calibration at the collapse point: $|q_\infty(\hat y_\infty) - p_{Y\mid p_1}(\hat y_\infty)| \le \mathrm{cal}_\infty$. & Genuine empirical claim. Not directly verifiable without OOD labels. \\
Assn.~\ref{assn:ood-marginal} & OOD label marginal is bounded away from a point mass. & Holds on all real benchmarks. \\
\bottomrule
\end{tabular}
\end{table}

\subsection{Summary of Assumptions}
\label{sec:theory-assumptions}

Table~\ref{tab:assumptions} collects all assumptions used in the appendix. Of the ten, only Assumption~\ref{assn:feat-orth} and Assumption~\ref{assn:ood-calibration} are genuinely empirical (i.e., not provable from the setup alone). The remainder are design choices, standard regularity, or consequences derived from Nagler's tilting.

\paragraph{Key assumptions.} Two assumptions are inherently empirical:
\begin{itemize}[leftmargin=1.5em]
  \item \textbf{Assumption~\ref{assn:feat-orth} (feature orthogonality)} is the scope boundary of the \emph{separation} argument (Theorem~\ref{thm:main-separation}). Where it fails, when OOD features retain substantial ID-like structure along the directions relevant for classification, the theorem does not guarantee $\bbE[\lambda_{\mathrm{OOD}}] < \bbE[\lambda_{\mathrm{ID}}]$. %
  \item \textbf{Assumption~\ref{assn:ood-calibration} (OOD calibration at the collapse point)} is the scope boundary of the \emph{harm} argument (Proposition~\ref{prop:harm-theorem}). Where it fails, $\hat\mu$ no longer tracks $\Delta$ and can systematically over- or under-estimate harm. It is also the most delicate assumption: it is the only one that constrains behavior on the shifted distribution, and without OOD labels we cannot directly verify it. %
\end{itemize}

\paragraph{Plausibility of the remaining harm-specific assumptions.} Beyond the key pair, four further conditions are relevant to the harm theorem (Proposition~\ref{prop:harm-theorem}):
\begin{itemize}[leftmargin=1.5em]
  \item Assumption~\ref{assn:calibration} (ID calibration) is routinely violated by modern deep networks (since they are often overconfident). A rigorous version of the theorem can work with (temperature-)scaled predictions using a held-out calibration set.
  \item Assumption~\ref{assn:far-ood} (far-OOD regime) is an idealized limit. For near-OOD, Proposition~\ref{prop:prediction-collapse} holds only approximately, so Proposition~\ref{prop:harm-theorem} degrades correspondingly; a formal near-OOD rate is left to future work.
  \item The mixture model $p_1 = (1-\alpha) p_0 + \alpha p_\star$ used to state Proposition~\ref{prop:harm-theorem} is a modeling choice. Real shifts may not decompose this way, but for general $p_1$ the same argument applies with $\alpha$ interpreted as the effective OOD mass $\int \indic\{\text{$x$ is far-OOD}\} \, dp_1$.
  \item Nagler's tilting theorem is proved only for one-layer transformers. Extending the argument to multilayer transformers (which TabPFN and TabICL use) is a nontrivial open question. Until this is resolved, this part of the argument is a \emph{mechanism hypothesis} rather than a proven property.
\end{itemize}

\paragraph{What holds unconditionally.} The density-ratio decomposition of Lemma~\ref{lem:density-ratio} and the TV upper bound of Corollary~\ref{cor:tv-loose} are both exact and require no assumptions. They already suffice to rebut the framing ``$\hat\lambda$ measures harm'' and to motivate the need for a second signal beyond $\hat\lambda$.

\newpage
\section{Scope and failure modes}
\label{sec:scope}

\subsection{Catalog of failure modes}
\label{sec:scope-catalog}

A PFN at deployment can exhibit many distinct kinds of failure, and \cald is targeted at a specific subset of them. We describe the failure modes a PFN may encounter, and indicate for each one whether \cald is empirically evaluated against it and whether the theory of Appendix~\ref{sec:theory} covers it. The space decomposes along three axes (see Table~\ref{tab:scope}):
\begin{itemize}[leftmargin=1.5em]
    \item \emph{Distribution shift} can be \textbf{covariate} ($p_0(x) \neq p_1(x)$ with $p(y\mid x)$ unchanged), \textbf{label} ($p_0(y) \neq p_1(y)$ with $p(x\mid y)$ unchanged), \textbf{concept} ($p(y\mid x)$ itself changes), or structured variants such as \textbf{subpopulation} (minority group reweighted) and \textbf{open-set} (novel classes appear at test time).
    \item Within distribution shift, the shift may be \textbf{harmful} (predictive accuracy drops) or \textbf{benign} (accuracy preserved or improved). \cald is calibrated to \emph{harm} rather than to \emph{distance}.
    \item \emph{Non-distributional failure modes} sit outside the i.i.d.\ framing of conformal calibration entirely: schema mismatch (different feature columns or dtypes), adversarial inputs, insufficient support set, numerical issues (NaNs, out-of-range values), and miscalibration without distributional change.
\end{itemize}

\begin{table}[h]
\centering
\small
\caption{\textbf{Scope of \cald.} Failure modes a PFN may encounter at deployment, with whether \cald is empirically evaluated against it and whether the theory of Appendix~\ref{sec:theory} covers it. \checkmark indicates full empirical coverage; \emph{partial} indicates a heuristic check (e.g., a severity sweep without a formal guarantee); $\times$ indicates absence.}
\vspace{2pt}
\label{tab:scope}
\resizebox{\linewidth}{!}{%
\begin{tabular}{l c c l}
\toprule
\textbf{Failure mode} & \textbf{Tested} & \textbf{Theory} & \textbf{Notes} \\
\midrule
\multicolumn{4}{l}{\emph{Distribution shift -- harmful (\cald should fire)}}\\
Covariate shift            & \checkmark & Thm.~\ref{thm:main-separation} (far-OOD) & main use case \\
Subpopulation shift        & \checkmark & severity sweep (heuristic)               & spurious attributes \\
Domain shift               & \checkmark & Thm.~\ref{thm:main-separation} (far-OOD) & subset of covariate shift \\
Near-OOD (overlapping support) & partial & vacuous in Thm.~\ref{thm:main-separation} & empirical only \\
\midrule
\multicolumn{4}{l}{\emph{Distribution shift -- benign by accuracy criterion}}\\
Covariate shift, $\Delta$ small & \checkmark & Lem.~\ref{lem:density-ratio} & \cald \emph{fires} via gradient signature; see \S\ref{sec:fp-empirical} \\
\midrule
\multicolumn{4}{l}{\emph{Out of scope}}\\
Pure label shift ($p(y)$ only) & $\times$ & none & test statistic is feature-driven \\
Concept shift ($p(y\mid x)$) & $\times$ & none & undetectable without test labels \\
Open-set / novel class     & $\times$ & partial (handled as far-OOD) & special case of label shift \\
Adversarial inputs         & $\times$ & none & breaks i.i.d.\ calibration \\
Schema / engineering errors & $\times$ & none & pre-test sanity check, not a monitor \\
Predictive miscalibration  & $\times$ & none & orthogonal to distributional shift \\
\bottomrule
\end{tabular}%
}%
\end{table}

\subsection{Empirical investigation: covariate shift without an accuracy drop}
\label{sec:fp-empirical}

The benign-by-accuracy row of Table~\ref{tab:scope} deserves a closer look. Among the seven benign datasets of Table~\ref{tab:baselines} (below the Hoeffding harm threshold $\Delta^{\star} \approx 4.84$\,pp at $\delta = 0.05$, $N_{\mathrm{exp2}} = 1280$, defined in Appendix~\ref{sec:deterioration-survey}), four deserve a closer look:
\begin{equation*}
  \cD_{\mathrm{benign}} = \{\textsf{diabetes\_readmission},\ \textsf{bank\_marketing},\ \textsf{credit\_default},\ \textsf{folktables}\}.
\end{equation*}
On these four datasets, the post-finetune anchor-loss test of \cald nevertheless rejects $H_0\!:\!\Xeval^* \sim p_0$ at high power for at least one of the two methods (cald-tabpfn, cald-tabicl) on every dataset. The full numbers are reproduced in Table~\ref{tab:fp-cald-aurocs}.

\begin{table}[h]
\centering
\small
\caption{\textbf{\cald and standalone shift detectors on the four datasets where $\Delta < \Delta^{\star}$.} AUROC computed on the same calibration protocol as Table~\ref{tab:detection} (10 seeds, 50 calibration rounds, support and anchor sets fixed within each seed). Every distribution-shift detector that does not condition on harm (KNN, MMD, Mahalanobis, Energy, MSP) also fires on these datasets at AUROC $\ge 0.9$, indicating real distributional shift in feature space.}
\label{tab:fp-cald-aurocs}
\vspace{2pt}
\begin{tabular}{l c c c c c c c c}
\toprule
& $\Delta_{\mathrm{acc}}$ (pp) & cald-tabpfn & cald-tabicl & KNN & MMD & Mahal. & Energy & MSP \\
\midrule
\textsf{diabetes\_readmission} & $-2.34$ & $0.976$ & $1.000$ & $1.000$ & $1.000$ & $1.000$ & $1.000$ & $0.869$ \\
\textsf{bank\_marketing}       & $+4.37$ & $0.997$ & $1.000$ & $1.000$ & $1.000$ & $0.989$ & $0.934$ & $1.000$ \\
\textsf{credit\_default}       & $+2.81$ & $1.000$ & $0.698$ & $1.000$ & $1.000$ & $1.000$ & $1.000$ & $0.764$ \\
\textsf{folktables}            & $+3.12$ & $0.624^{\dagger}$ & $0.899$ & $0.926$ & $1.000$ & $0.968$ & $0.722$ & $0.833$ \\
\bottomrule
\end{tabular}
\vspace{2pt}\\
{\footnotesize $\Delta_{\mathrm{acc}} := \mathrm{acc}(p_0) - \mathrm{acc}(p_1)$ (positive = OOD accuracy drop). $^{\dagger}$ The LR-stability heuristic (Algorithm~\ref{alg:method}) flags this cell as \emph{unstable} and falls back to $\eta = 10^{-4}$, where the signal collapses; see \S\ref{sec:fp-self-correct}.}
\end{table}

\paragraph{Reading the table.} The AUROC values for KNN, MMD, Mahalanobis, Energy, and MSP show that the four datasets are not edge cases: their ID and OOD distributions are easily distinguishable in feature space by every standalone shift detector we evaluate. The benign label is purely a property of the predictive accuracy gap $\Delta(q, p_0, p_1)$ as decomposed in Lemma~\ref{lem:density-ratio}: it places strong constraints on how mass moves on the mistake region $M = \{(x, y) : \indic\{y \neq \arg\max_c q_{W_0}(c \mid x)\} = 1\}$, but no constraints whatsoever on how mass moves on $M^c$. In all four datasets we observe heavy reweighting of $M^c$ (which is most of the input domain when the pretrained classifier $q_{W_0}$ is accurate on $p_0$): \cald, like every shift detector that uses information about $M^c$, picks this up.

\subsection{Two sub-mechanisms within Theorem~\ref{thm:main-separation}}
\label{sec:fp-decomposition}

The above empirical observation is consistent with, and indeed predicted by, the multi-step main separation theorem. Recall the central inequality (Theorem~\ref{thm:main-separation}):
\begin{equation}
  \bbE[\lambda_{\mathrm{ID}}] - \bbE[\lambda_{\mathrm{OOD}}]
    \;\ge\; T\eta\,\bigl(\gamma - \epsilon\, B_\phi^2\, B_C\bigr) - O(T^2 \eta^2 L\, B_g^2),
  \label{eq:fp-main-sep-restated}
\end{equation}
where $\gamma := -\bbE_{p_0 \otimes p_0}[C_{ij} K_{ij}] > 0$ is the ID interference strength (Proposition~\ref{prop:id-interference}) and $\epsilon$ is the OOD feature-orthogonality parameter (Assumption~\ref{assn:feat-orth} and Corollary~\ref{cor:feature-orth}). The detector fires when the right-hand side is positive, i.e.\ when
\begin{equation}
  \gamma - \epsilon\, B_\phi^2\, B_C > 0.
  \label{eq:fp-firing-condition}
\end{equation}

The proof of Theorem~\ref{thm:main-separation} (Appendix~\ref{sec:theory-main-separation}) makes the firing condition slightly richer than~\eqref{eq:fp-firing-condition} suggests. Specializing equation~\eqref{eq:separation-decomp} to the $T$-step regime and decomposing the cross-term in polar form $\langle \bar g_a, \bar g_e\rangle_F = \|\bar g_a\|_F \|\bar g_e\|_F \cos\theta$, where $\theta$ is the angle between the time-averaged anchor and disagree gradients, yields
\begin{align}
  \bbE[\lambda_{\mathrm{ID}} - \lambda_{\mathrm{OOD}}]
  &= T\eta\,\Bigl[\bbE\bigl(\|\bar g_a\|_F \|\bar g_e^{\mathrm{ID}}\|_F \cos\theta_{\mathrm{ID}}\bigr) \nonumber\\
  &\hspace{4em} - \bbE\bigl(\|\bar g_a\|_F \|\bar g_e^{\mathrm{OOD}}\|_F \cos\theta_{\mathrm{OOD}}\bigr)\Bigr] \nonumber\\
  &\hspace{2em} + O(T^2 \eta^2 L\, B_g^2),
  \label{eq:fp-polar-decomposition}
\end{align}
under the leading-order identification $\bar g_a^{\mathrm{ID}} \approx \bar g_a^{\mathrm{OOD}} =: \bar g_a$ established in the theorem's proof. The integrated separation is therefore positive whenever
\begin{equation}
  \underbrace{\bbE\bigl(\|\bar g_a\|_F \|\bar g_e^{\mathrm{ID}}\|_F \cos\theta_{\mathrm{ID}}\bigr)}_{\text{ID conflict, large negative}}
  \;<\;
  \underbrace{\bbE\bigl(\|\bar g_a\|_F \|\bar g_e^{\mathrm{OOD}}\|_F \cos\theta_{\mathrm{OOD}}\bigr)}_{\text{OOD signed cross-term}}.
  \label{eq:fp-separation-cond}
\end{equation}
The right-hand side can be made larger than the left-hand side by either of the following two mutually compatible mechanisms:

\begin{description}[leftmargin=2.5em, style=nextline]
  \item[\textbf{(M1) Cosine shift toward less conflict.}] $\cos\theta_{\mathrm{OOD}} > \cos\theta_{\mathrm{ID}}$ with $\|\bar g_e^{\mathrm{OOD}}\|_F \approx \|\bar g_e^{\mathrm{ID}}\|_F$. Even a small positive shift $\Delta\!\cos := \cos\theta_{\mathrm{OOD}} - \cos\theta_{\mathrm{ID}}$ is enough to make~\eqref{eq:fp-separation-cond} positive when $\theta_{\mathrm{ID}}$ is in a strongly-conflicting regime ($\cos\theta_{\mathrm{ID}} \to -1$).
  \item[\textbf{(M2) Eval-gradient magnitude shrinkage.}] $\|\bar g_e^{\mathrm{OOD}}\|_F < \|\bar g_e^{\mathrm{ID}}\|_F$ with $\cos\theta_{\mathrm{OOD}} \approx \cos\theta_{\mathrm{ID}}$ (or even $\cos\theta_{\mathrm{OOD}} < \cos\theta_{\mathrm{ID}}$, provided the magnitude attenuation dominates). This is the canonical Nagler-tilting picture of Appendix~\ref{sec:theory-nagler}: under far-OOD inputs the prediction collapses, the disagree loss saturates, and $\|\bar g_e\|_F \to 0$.
\end{description}

The original informal argument in the body of the paper (Section~\ref{sec:method-intuition}) emphasized M2; it presented OOD as a regime where ``the PFN is uncertain on its own pseudolabels, $\nabla \cL_{\mathrm{eval}}$ is small in magnitude and approximately orthogonal to $\nabla \cL_{\mathrm{anc}}$''. Theorem~\ref{thm:main-separation}'s actual statement is more permissive: the firing condition~\eqref{eq:fp-separation-cond} holds whenever \emph{either} mechanism produces enough integrated effect. M1 is sufficient on its own. M2 is sufficient on its own. The two can also reinforce or partially cancel.

\paragraph{Why benign datasets satisfy~\eqref{eq:fp-separation-cond}.} On the benign-by-accuracy datasets, $q_{W_0}$ classifies the OOD batch with accuracy comparable to or better than ID, so the prediction-collapse picture of Proposition~\ref{prop:prediction-collapse} does not apply: the model is confident on OOD, and $\|\bar g_e^{\mathrm{OOD}}\|_F$ is comparable to $\|\bar g_e^{\mathrm{ID}}\|_F$. M2 is muted. But the geometry of the OOD batch in feature space is still different: a non-trivial fraction of OOD points fall in regions where the kernel $K_{ij} = \langle \tilde\phi(x_a^i), \tilde\phi(x_e^j)\rangle$ has a distribution that is no longer centered exactly where it sits under $p_0$. This perturbs $\cos\theta$ by a small amount, which integrated over $T$ steps suffices for~\eqref{eq:fp-separation-cond} to hold.

The benign-firing cells therefore correspond to the $\gamma$-driven branch of~\eqref{eq:fp-firing-condition}: $\gamma$ remains close to its ID value ($p_0$ is unchanged on the anchor side) and $\epsilon\, B_\phi^2\, B_C$ does not need to vanish for the bound to be active: a small $\Delta\!\cos$ produced by partial feature reorganization is enough.

\subsection{Gradient-probe diagnostics}
\label{sec:fp-probe}

To verify the M1/M2 picture empirically, we instrument the inner finetune loop of \cald to log the per-step gradient quantities of equation~\eqref{eq:fp-polar-decomposition}. For each of the four benign datasets and two harmful controls (\textsf{camelyon17}, \textsf{covertype}), and for each of the two methods (cald-tabpfn, cald-tabicl), we run $n_{\mathrm{rounds}} = 10$ probe rounds at three seeds with the LR selected by the same heuristic as in Table~\ref{tab:detection}. At every step $t = 1, \ldots, T = 50$ within every round we record $\|g_a^{(t)}\|_F^2$, $\|g_e^{(t)}\|_F^2$, $\langle g_a^{(t)}, g_e^{(t)}\rangle_F$, and $\cL_{\mathrm{anc}}(W_t)$. We additionally log the pre-finetune cosine similarity $\cos\theta_0 = \langle g_a^{(0)}, g_e^{(0)}\rangle_F / (\|g_a^{(0)}\|_F\, \|g_e^{(0)}\|_F)$ on each round.

The probe code (\texttt{detection/viz/cald\_gradient\_probe.py}) calls into the reference implementation \texttt{detection/core/finetune.py:run\_one\_iteration} for the cald-tabpfn arm and uses a self-contained instrumented loop for cald-tabicl. Aggregating across seeds and rounds for each (dataset, method) cell, we report:
\begin{itemize}[leftmargin=1.5em]
  \item $\lambda_{\mathrm{gap}} := \bbE[\lambda_{\mathrm{ID}}] - \bbE[\lambda_{\mathrm{OOD}}]$, the directly observed integrand of the conformal test (Appendix~\ref{sec:theory-conformal});
  \item $\rho_e^{(50)} := \bbE\|\bar g_e^{\mathrm{OOD}}\|_F^2 / \bbE\|\bar g_e^{\mathrm{ID}}\|_F^2$, averaged over all 50 finetune steps and across rounds and seeds;
  \item $\Delta\!\cos := \bbE[\cos\theta_0^{\mathrm{OOD}}] - \bbE[\cos\theta_0^{\mathrm{ID}}]$, the pre-finetune cosine shift;
  \item $H_{\mathrm{ID}}, H_{\mathrm{OOD}}$, the mean entropy of $q_{W_0}(\cdot \mid x_e)$ on the eval batch (a check on whether the prediction-collapse picture of Proposition~\ref{prop:prediction-collapse} is active).
\end{itemize}

Table~\ref{tab:fp-probe-summary} reports these four quantities together with the conformal AUROC for the (dataset, method) cell.

\begin{table}[h]
\centering
\small
\caption{\textbf{Gradient-probe summary across the eight benign cells and four harmful-control cells.} $\rho_e^{(50)} < 1$ indicates eval-gradient magnitude shrinkage on OOD (mechanism M2). $\Delta\!\cos > 0$ indicates a cosine shift toward less conflict on OOD (mechanism M1). Cells with AUROC $\ge 0.99$ exhibit at least one of M1, M2 with the correct sign (and often both); cells with AUROC $< 0.7$ exhibit neither.}
\label{tab:fp-probe-summary}
\vspace{2pt}
\resizebox{\linewidth}{!}{%
\begin{tabular}{l l c c r r r r r r}
\toprule
\textbf{dataset} & \textbf{method} & \textbf{kind} & \textbf{AUROC} & $\Delta_{\mathrm{acc}}$ (pp) & $\lambda_{\mathrm{gap}}$ & $\rho_e^{(50)}$ & $\cos\theta_0^{\mathrm{ID}}$ & $\cos\theta_0^{\mathrm{OOD}}$ & $\Delta\!\cos$ \\
\midrule
Diab.    & cald-tabpfn  & benign  & $0.976$ & $-2.34$ & $+0.352$ & $0.421$ & $-1.000$ & $-0.973$ & $+0.027$ \\
Bank     & cald-tabpfn  & benign  & $0.997$ & $+4.37$ & $+0.058$ & $0.909$ & $-0.336$ & $-0.335$ & $+0.001$ \\
Credit   & cald-tabpfn  & benign  & $1.000$ & $+2.81$ & $+0.269$ & $0.155$ & $-0.336$ & $-0.300$ & $+0.036$ \\
Folktbl. & cald-tabpfn  & benign  & $0.624$ & $+3.12$ & $-0.004$ & $0.878$ & $+0.992$ & $+0.990$ & $-0.002$ \\
Diab.    & cald-tabicl  & benign  & $1.000$ & $-2.34$ & $+0.085$ & $0.982$ & $-0.253$ & $-0.240$ & $+0.013$ \\
Bank     & cald-tabicl  & benign  & $1.000$ & $+4.37$ & $+0.155$ & $0.749$ & $+0.969$ & $+0.985$ & $+0.016$ \\
Credit   & cald-tabicl  & benign  & $0.698$ & $+2.81$ & $+0.012$ & $0.948$ & $+0.442$ & $+0.440$ & $-0.002$ \\
Folktbl. & cald-tabicl  & benign  & $0.899$ & $+3.12$ & $+0.025$ & $0.909$ & $+0.483$ & $+0.376$ & $-0.108$ \\
\midrule
Camel.   & cald-tabpfn  & harmful & $0.675$ & $+8.91$ & $+0.069$ & $0.718$ & $+0.059$ & $-0.231$ & $-0.289$ \\
Camel.   & cald-tabicl  & harmful & $1.000$ & $+8.91$ & $+0.180$ & $0.719$ & $-0.189$ & $-0.611$ & $-0.422$ \\
Covert.  & cald-tabpfn  & harmful & $1.000$ & $+83.52$ & $+0.163$ & $0.269$ & $-0.697$ & $+0.240$ & $+0.937$ \\
Covert.  & cald-tabicl  & harmful & $1.000$ & $+83.52$ & $+0.468$ & $0.060$ & $+0.075$ & $+0.087$ & $+0.011$ \\
\bottomrule
\end{tabular}%
}
\end{table}

\paragraph{Reading Table~\ref{tab:fp-probe-summary}.} The eight benign cells fall into a consistent pattern with respect to mechanism M1: in five of eight cells, $\Delta\!\cos > 0$ at magnitudes between $+0.001$ and $+0.036$. The exceptions are folktables (both methods, $\Delta\!\cos$ negative or zero) and credit\_default/cald-tabicl ($\Delta\!\cos = -0.002$). On those exception cells, M2 carries the signal: $\rho_e^{(50)}$ is well below $1$, indicating the magnitude attenuation. The only cell where neither mechanism is active is folktables/cald-tabpfn at the unstable LR, with $\rho_e^{(50)} = 0.878$ and $\Delta\!\cos = -0.002$ both close to neutral; the $\lambda_{\mathrm{gap}}$ is correspondingly $-0.004$, and the conformal test does not reject (\S\ref{sec:fp-self-correct}).

The four harmful controls are consistent with M2 being the dominant mechanism: $\rho_e^{(50)}$ ranges from $0.060$ to $0.719$, all well below $1$, while $\Delta\!\cos$ takes both signs depending on dataset. The covertype/cald-tabpfn cell is the cleanest example of M1 and M2 firing together, with the cosine flipping sign ($\cos\theta_0^{\mathrm{ID}} = -0.697 \to \cos\theta_0^{\mathrm{OOD}} = +0.240$, $\Delta\!\cos = +0.937$) and the magnitude collapsing by a factor of $\rho_e^{(50)} = 0.27$.

\paragraph{Per-seed consistency.} The aggregate $\Delta\!\cos$ values reported above are seed-means; the underlying distribution of $\cos\theta_0$ is sensitive to the random init of the decoder, which can flip its sign across seeds. To guard against the pathology of two seeds with opposite signs averaging to zero, we also computed $\Delta\!\cos$ per-seed for the eight benign cells (24 seed--cell pairs). On diabetes\_readmission and bank\_marketing, $\Delta\!\cos$ is positive in all six seeds for both methods; on credit\_default/cald-tabpfn, two of three seeds have $\Delta\!\cos > 0$ and the third is small negative ($-0.020$). On folktables/cald-tabpfn, $\Delta\!\cos$ is essentially zero in all three seeds. The aggregate signs in Table~\ref{tab:fp-probe-summary} are not seed-mean artifacts.

\subsection{Case studies}
\label{sec:fp-cases}

We highlight three (dataset, method) cells that span the qualitative regimes.

\paragraph{Case 1: \textsf{diabetes\_readmission}/cald-tabpfn (M1, benign).} Despite $\Delta_{\mathrm{acc}} = -2.34$\,pp (model is more accurate on OOD than ID), the gradient probe records a $\lambda_{\mathrm{gap}}$ of $+0.352$, larger than every harmful control except covertype/cald-tabicl. The mechanism is purely M1: $\rho_e^{(50)} = 0.42$ shows mild magnitude attenuation but $\cos\theta_0^{\mathrm{ID}} = -1.000$ vs $\cos\theta_0^{\mathrm{OOD}} = -0.973$ shows the cosine moving very slightly out of the strict-conflict regime. Integrated over $T=50$ steps with $\eta = 10^{-2}$, this $\Delta\!\cos = +0.027$ is enough to satisfy~\eqref{eq:fp-separation-cond}. The pre-finetune entropy $H_{\mathrm{OOD}} = 0.297$ is in fact slightly \emph{below} $H_{\mathrm{ID}} = 0.318$, ruling out the prediction-collapse picture for this dataset.

\begin{figure}[h]
  \centering
  \includegraphics[width=0.92\linewidth]{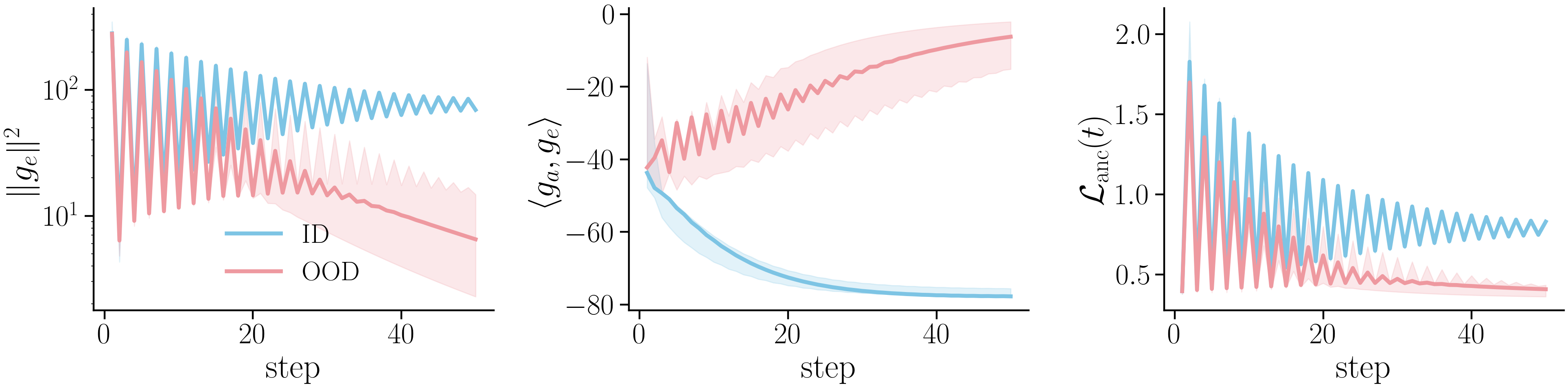}
  \caption{\textbf{Mechanism M1 on a benign dataset (\textsf{diabetes\_readmission}/cald-tabpfn).} Per-step trajectories of $\|g_e\|_F^2$ (left), $\langle g_a, g_e\rangle_F$ (middle), and $\cL_{\mathrm{anc}}(W_t)$ (right), median across rounds and seeds with interquartile shading. The cross-term diverges between ID (more negative) and OOD (less negative), driving the integrated $\lambda_{\mathrm{gap}} = +0.352$ even though $\Delta_{\mathrm{acc}} = -2.34$\,pp.}
  \label{fig:fp-mechanism-diabetes}
\end{figure}

Figure~\ref{fig:fp-mechanism-diabetes} plots $\|g_e^{(t)}\|_F^2$, $\langle g_a^{(t)}, g_e^{(t)}\rangle_F$, and $\cL_{\mathrm{anc}}(W_t)$ versus $t$ for ID and OOD on this cell. The middle panel is the diagnostic visual: the cross-term on ID descends to $\approx -80$ over the trajectory while the cross-term on OOD plateaus near $-10$. Anchor descent on OOD therefore proceeds with five-fold less integrated interference than on ID, and the right panel shows the consequence: $\cL_{\mathrm{anc}}(W_T)$ on OOD reaches $\approx 0.4$ while on ID it stays around $0.75$.

\begin{figure}[h]
  \centering
  \includegraphics[width=0.92\linewidth]{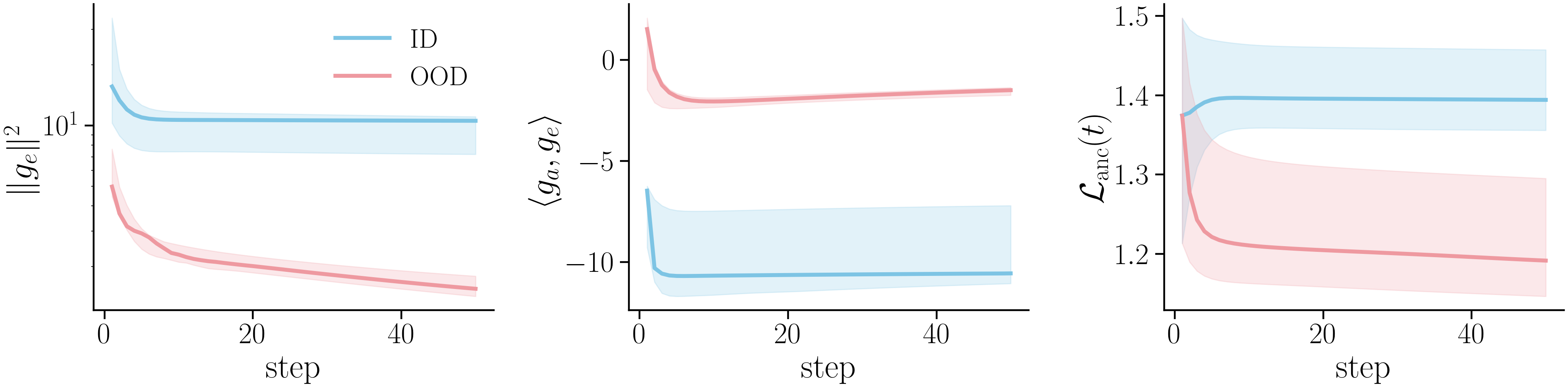}
  \caption{\textbf{Mechanisms M1+M2 firing together on a harmful dataset (\textsf{covertype}/cald-tabpfn).} The OOD eval-gradient magnitude is uniformly an order of magnitude smaller than ID throughout the trajectory (left), the cross-term sign on OOD is much closer to zero than on ID (middle), and the anchor loss on OOD descends to a clearly distinct level (right).}
  \label{fig:fp-mechanism-covertype}
\end{figure}

\paragraph{Case 2: \textsf{covertype}/cald-tabpfn (M1+M2, harmful).} The two mechanisms both fire and reinforce each other. $\rho_e^{(50)} = 0.269$ shows strong magnitude collapse (M2): the OOD prediction is near-uniform on $7$ classes (consistent with $H_{\mathrm{OOD}} = 1.479$ approaching the maximal entropy $\log 7 \approx 1.95$), and the disagree loss saturates. Simultaneously, $\Delta\!\cos = +0.937$ (M1): the cosine flips from anti-aligned ($\cos\theta_0^{\mathrm{ID}} = -0.697$) to mildly aligned ($\cos\theta_0^{\mathrm{OOD}} = +0.240$). The per-step trajectories (Figure~\ref{fig:fp-mechanism-covertype}) are textbook for both mechanisms.

\paragraph{Case 3: \textsf{camelyon17}/cald-tabicl (M2 only, harmful).} The cosine on this cell moves \emph{further} into the conflict regime ($\Delta\!\cos = -0.422$) but the AUROC is $1.000$. The mechanism is pure M2: $\rho_e^{(50)} = 0.719$ (eval gradient $\sim 30$\% smaller on OOD). Mechanism M1's contribution to~\eqref{eq:fp-separation-cond} is negative on this cell, but the integrated M2 contribution dominates: the magnitude attenuation reduces the absolute size of the OOD cross-term in~\eqref{eq:fp-polar-decomposition} below the ID cross-term, even though the OOD angle is more anti-aligned.

\begin{figure}[h]
  \centering
  \includegraphics[width=\linewidth]{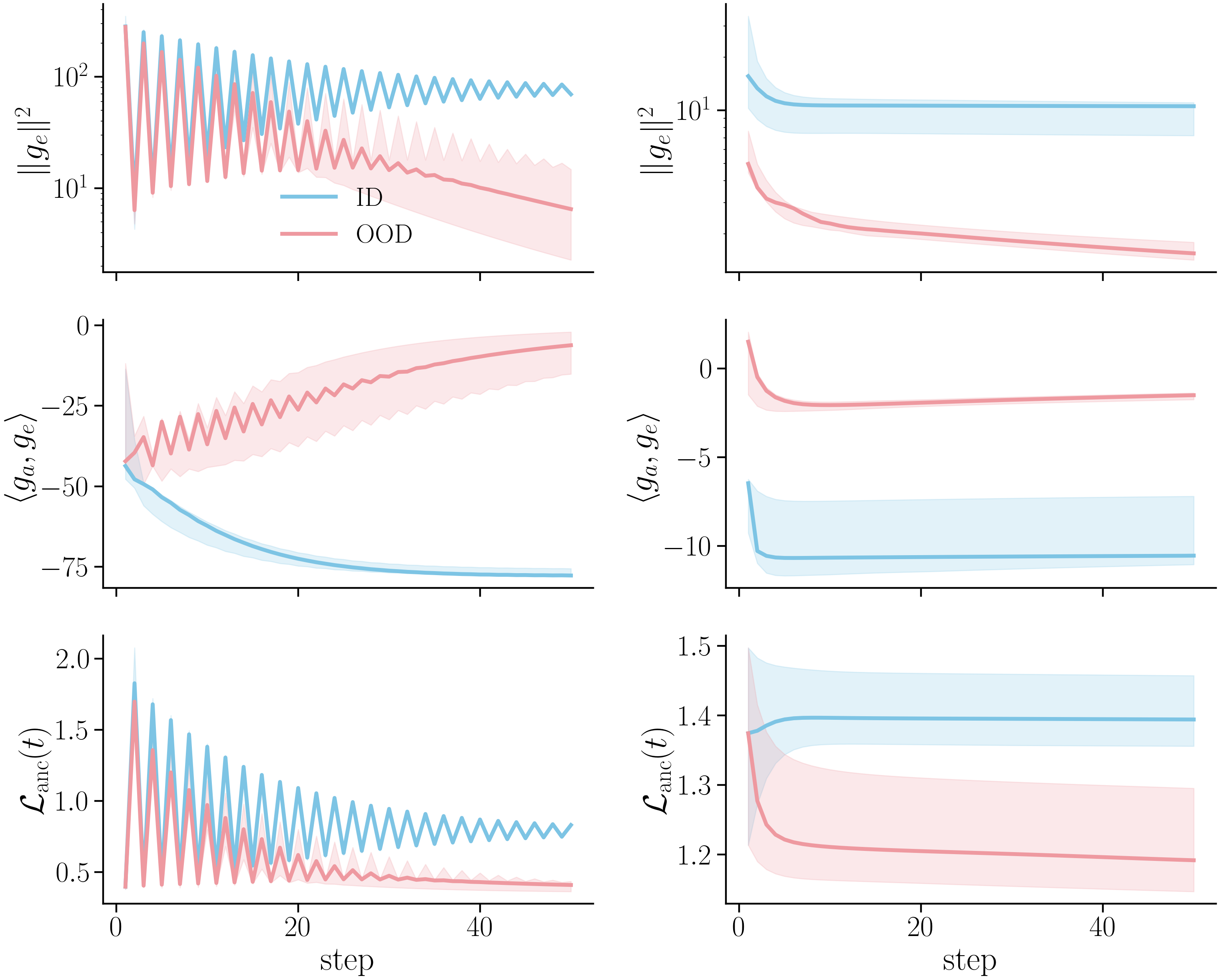}
  \caption{\textbf{Same detector outcome ($\lambda_{\mathrm{OOD}} < \lambda_{\mathrm{ID}}$), two different mechanisms.} Left column: \textsf{diabetes\_readmission} (benign, $\Delta_{\mathrm{acc}} = -2.34$\,pp, AUROC $0.976$) fires via cosine shift (M1). Right column: \textsf{covertype} (harmful, $\Delta_{\mathrm{acc}} = +83.5$\,pp, AUROC $1.000$) fires via the combined M1+M2 mechanism, with a much cleaner log-scale separation in $\|g_e\|_F^2$.}
  \label{fig:fp-contrast}
\end{figure}

\paragraph{Case 4: \textsf{diabetes\_readmission}/cald-tabpfn vs.\ \textsf{covertype}/cald-tabpfn (head-to-head).} Figure~\ref{fig:fp-contrast} places the cleanest M1 example side-by-side with the cleanest M1+M2 example. Both produce $\lambda_{\mathrm{OOD}} < \lambda_{\mathrm{ID}}$ with high confidence; the gradient signatures are visibly distinct.

\begin{figure}[h]
  \centering
  \includegraphics[width=0.78\linewidth]{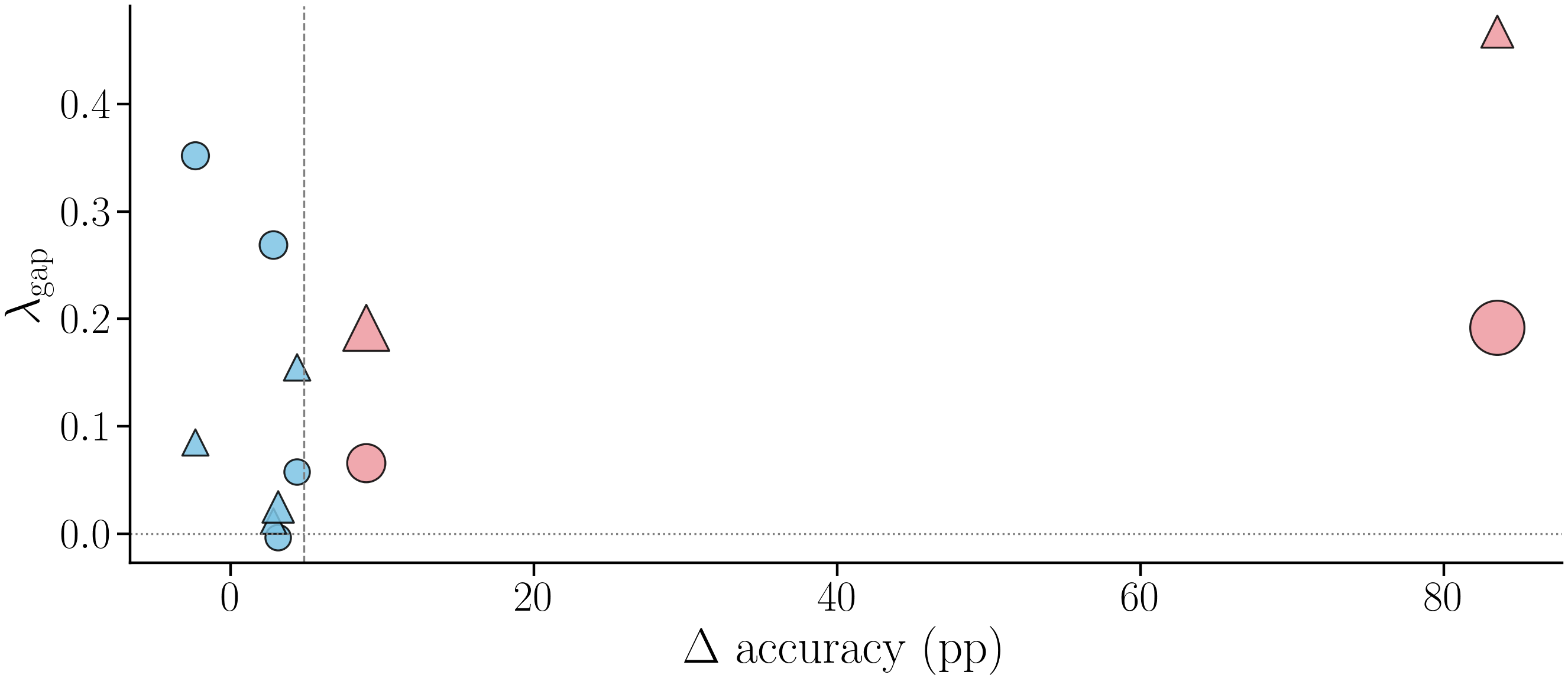}
  \caption{\textbf{$\lambda_{\mathrm{gap}}$ versus $\Delta_{\mathrm{acc}}$ across the twelve probe cells.} Marker size scales with $|\Delta\!\cos|$. The dashed vertical line is the Hoeffding harm threshold $\Delta^{\star} \approx 4.84$\,pp. Points to the left of the threshold are benign; points to the right are harmful. Both regions contain cells with $\lambda_{\mathrm{gap}} \ge 0.05$, confirming that the conformal test fires on both kinds of distributional shift.}
  \label{fig:fp-headline}
\end{figure}

Figure~\ref{fig:fp-headline} plots $\lambda_{\mathrm{gap}}$ against $\Delta_{\mathrm{acc}}$ for all twelve cells in Table~\ref{tab:fp-probe-summary}. The visualization makes the central claim concrete: the largest $\lambda_{\mathrm{gap}}$ in the benign region is $+0.352$ (\textsf{diabetes\_readmission}/cald-tabpfn), which exceeds the $\lambda_{\mathrm{gap}}$ of the harmful camelyon17 (both methods). The two axes are not monotonically related on this 12-cell sample.

\subsection{Fallback behavior of the LR-stability rule}
\label{sec:fp-self-correct}

The single benign cell in Table~\ref{tab:fp-probe-summary} where the conformal test does not reject is \textsf{folktables}/cald-tabpfn, with AUROC $0.624$, $\lambda_{\mathrm{gap}} = -0.004$, $\rho_e^{(50)} = 0.878$, and $\Delta\!\cos = -0.002$. Both mechanisms M1 and M2 are dormant here: the gradient signature of equation~\eqref{eq:fp-separation-cond} is essentially neutral, and the detector's reported AUROC is the only one below $0.7$ across the eight benign cells.

The low AUROC and the fallback have a common cause. Algorithm~\ref{alg:method} selects a finetuning learning rate per (dataset, method) cell with a tail-stability rule: it checks the per-step direction indicator $\indic\{\bbE\lambda_{\mathrm{OOD}}^{(t)} < \bbE\lambda_{\mathrm{ID}}^{(t)}\}$ for $t \in [30, 50]$ across calibration rounds, and rejects a candidate $\eta$ if the indicator flips more than twice or has tail-mean below $0.5$. On folktables/cald-tabpfn the rule rejects $\eta = 10^{-2}$ and falls back to $\eta = 10^{-4}$, flagging the cell \texttt{unstable}; by~\eqref{eq:fp-firing-condition}, when neither M1 nor M2 produces a coherent integrated effect, $\bar g_a$ and $\bar g_e$ are too noisy to define a reliable post-finetune $\lambda$, and the rule registers this as tail oscillation. We report the modest AUROC on this cell as a measured limitation of the detector in this regime: the stability rule localizes the unreliable cell, but it does not repair it.

A census over all 30 method $\times$ dataset cells shows the fallback is rare and confined to benign shifts: four cells fall back to $\eta = 10^{-4}$ (adult\_income on both models, at AUROC $1.000$ despite the fallback; celebA/cald-tabicl at $0.845$; and folktables/cald-tabpfn at $0.624$), while every severely shifted dataset selects its learning rate stably. The selection is also insensitive to the choice of tail window: re-running the rule with windows $25$--$45$, $30$--$50$, and $35$--$50$ changes the selected rate in 1 of 60 cell--window comparisons (folktables/cald-tabpfn under $25$--$45$, $\Delta$AUROC $-0.052$). Appendix~\ref{sec:abl-frozen} reports the full census together with the frozen-configuration comparison.

\subsection{Implications for the calibration-to-harm claim}
\label{sec:fp-implications}

Section~\ref{sec:method-empirical-calibration} and \S\ref{sec:abl-calibration} of the main text present the empirical observation that AUROC scales monotonically with shift severity (calibration to harm). The probe data above show that this monotonicity does not hold on the four-dataset benign group; in particular, the diabetes\_readmission/cald-tabpfn cell has $\lambda_{\mathrm{gap}} = +0.352$ at $\Delta_{\mathrm{acc}} = -2.34$\,pp, while camelyon17/cald-tabpfn has $\lambda_{\mathrm{gap}} = +0.069$ at $\Delta_{\mathrm{acc}} = +8.91$\,pp.

The right interpretation, supported by Lemma~\ref{lem:density-ratio}, is that $\lambda_{\mathrm{gap}}$ and $\Delta_{\mathrm{acc}}$ measure formally different objects:
\begin{itemize}[leftmargin=1.5em]
  \item $\Delta_{\mathrm{acc}}$ depends only on how the shift moves mass on the mistake region $M$ (Lemma~\ref{lem:density-ratio}). In particular, $\Delta_{\mathrm{acc}}$ is invariant under any reweighting of $M^c$ that preserves the marginal on $M$.
  \item $\lambda_{\mathrm{gap}}$ depends on the integrated cross-term~\eqref{eq:fp-polar-decomposition}, which is a feature-space quantity. Reweighting that occurs entirely on $M^c$ still perturbs the geometry of $\bar g_a, \bar g_e$ and therefore still affects $\lambda_{\mathrm{gap}}$.
\end{itemize}
Two shifts with the same $|\Delta_{\mathrm{acc}}|$ can therefore produce very different $\lambda_{\mathrm{gap}}$ values. The four benign-by-accuracy datasets are the regime where $|\Delta_{\mathrm{acc}}|$ is small but feature-space reweighting on $M^c$ is large; the harm decomposition of Lemma~\ref{lem:density-ratio} predicts exactly this kind of dissociation.

The accuracy-sensitive auxiliary signal $\hat\mu$ proposed in Proposition~\ref{prop:harm-theorem} is the formal vehicle for restoring calibration to harm: $\hat\mu$ is constructed to track the density-ratio integral of Lemma~\ref{lem:density-ratio} directly, rather than the unsigned TV distance that drives $\lambda$. A natural reading of the present appendix's empirical findings is that $\hat\lambda$ and $\hat\mu$ should be reported jointly: $\hat\lambda$ confirms shift, $\hat\mu$ adjudicates whether the shift is harmful. The dissociation observed on the four benign datasets is then not a defect of \cald but a quantitative reading of the same $\hat\lambda$/$\hat\mu$ separation that the harm theorem (Proposition~\ref{prop:harm-theorem}) makes formal.

\newpage
\section{Experimental Details}
\label{sec:experimental-details}

This appendix collects dataset-, baseline-, and protocol-level details deferred from Section~\ref{sec:experiments}, together with a survey of TabPFN/TabICL accuracy under the shifts considered. \S\ref{sec:dataset-description} describes the 14 datasets and their ID/OOD splits; \S\ref{sec:baselines-detail} catalogs the 15 post-hoc baselines we compare against; \S\ref{sec:deterioration-survey} surveys the per-dataset accuracy drops and applies a Hoeffding bound to classify each shift as benign or harmful; and \S\ref{sec:main-results-full} reports our main TPR@5\%\,FPR results with confidence intervals and explains the LR-stability protocol used for \cald.

\subsection{Dataset Description}
\label{sec:dataset-description}

We evaluate on 14 datasets covering tabular, vision-foundation-model, and text-foundation-model representations. For each dataset we describe the modality and the criterion that splits in-distribution (ID) from out-of-distribution (OOD); in many cases the OOD pool corresponds to the dataset's own test split, with the criterion baked in by the dataset designers. Sample-size and protocol details are deferred to \S\ref{sec:main-results-full}.

\begin{enumerate}[leftmargin=1.4em]
    \item \textbf{UCI Heart Disease.} Tabular medical features (13 covariates: age, sex, chest-pain type, cholesterol, ECG measurements, exercise-induced angina, etc.). \emph{ID/OOD:} Cleveland Clinic patients are ID, Hungarian Institute of Cardiology patients are OOD. Split is \emph{site} shift: different hospital, different population, different measurement conventions, with non-trivial covariate redistribution across the 13 fields.
    \item \textbf{Diabetes Readmission} (UCI \#296). Tabular hospital records (49 covariates: demographics, admission codes, lab counts, medication lists, length of stay). \emph{ID/OOD:} \texttt{admission\_type} = \emph{Emergency} is ID, \texttt{Elective} is OOD. Split is \emph{patient-acuity} shift: emergency admissions skew toward acute conditions and shorter stays, elective toward planned procedures and longer stays, redistributing nearly every covariate.
    \item \textbf{Adult Income} (UCI \#2). Tabular 1994 US Census records (14 covariates: age, work class, education, occupation, hours-per-week, etc.). \emph{ID/OOD:} \texttt{sex} = \emph{Male} is ID, \emph{Female} is OOD. Split is \emph{demographic} shift: in 1994 the labor-force composition for men and women differs sharply on hours, occupation, and capital gains, so the covariate distributions are far from identical.
    \item \textbf{Bank Marketing} (UCI \#222). Tabular Portuguese-bank telemarketing call records (20 covariates including age, job, balance, contact channel, prior-campaign outcome, and macroeconomic indicators). \emph{ID/OOD:} calls in \emph{May--August 2008} are ID, calls in \emph{October--November 2008} are OOD. Split is \emph{temporal/macroeconomic} shift bracketing the September 2008 Lehman collapse, after which euro-zone interest rates and consumer-confidence indicators in the dataset shift abruptly.
    \item \textbf{Credit Default} (UCI \#350). Tabular Taiwanese credit-card client records (23 covariates: limit balance, sex, age, education, six months of bill amounts and payment statuses). \emph{ID/OOD:} \texttt{EDUCATION} = \emph{University} is ID, \emph{Graduate school} is OOD. Split is \emph{education-stratum} shift: graduate-degree holders systematically have higher credit limits and different payment patterns.
    \item \textbf{Covertype} (sklearn \texttt{fetch\_covtype}). Tabular forest-cover records from Roosevelt National Forest (54 covariates: 10 quantitative including elevation/slope/aspect/distances, 4 binary wilderness areas, 40 binary soil types). \emph{ID/OOD:} \emph{2{,}000--2{,}500\,m} elevation is ID, \emph{$>$3{,}000\,m} is OOD. Split is \emph{ecological-zone} shift: above 3{,}000 m the forest transitions from mixed broadleaf--conifer to pure subalpine conifer, redistributing soil type and slope.
    \item \textbf{Folktables ACS Income}. Tabular American Community Survey records (10 covariates: age, work class, education, marital status, occupation, etc.). \emph{ID/OOD:} we use a state-pair split following the Folktables release. Split is \emph{geographic} shift: different states have different industrial mixes and educational distributions.
    \item \textbf{Camelyon17} (\citealt{bandi2018detection,koh2021wilds}). Whole-slide histopathology image patches encoded by a frozen DINOv2 ViT-B/14 to 768-dim representations. \emph{ID/OOD:} patches from three of the five WILDS hospitals are ID, the held-out OOD hospital is OOD. Split is \emph{hospital/staining} shift: different scanners and H\&E staining protocols produce systematic color and texture differences.
    \item \textbf{CIFAR-10 $\to$ CIFAR-10.1} (\citealt{recht2018cifar,recht2019imagenet}). Natural-image patches encoded by an ImageNet-pretrained ResNet-18 to 512-dim representations. \emph{ID/OOD:} CIFAR-10 test set is ID, CIFAR-10.1 v6 is OOD. Split is \emph{benchmark-replication} shift: \citet{recht2019imagenet} re-collected CIFAR-10 following the original protocol; the resulting class marginals match but the covariate distributions diverge enough to drop most classifiers' accuracy by 6--10\,pp.
    \item \textbf{PACS} (\citealt{li2017deeper}). Object images from four artistic domains, encoded by frozen DINOv2 ViT-B/14 to 768-dim representations. \emph{ID/OOD:} \emph{Photo} is ID, \emph{Art-painting / Cartoon / Sketch} are OOD. Split is \emph{style/domain} shift, classically used as a domain-generalization benchmark.
    \item \textbf{Waterbirds} (WILDS, \citealt{koh2021wilds}). Bird photographs encoded by frozen DINOv2 ViT-B/14 to 768-dim representations. \emph{ID/OOD:} the majority groups (waterbird-on-water, landbird-on-land) are ID; the minority groups (waterbird-on-land, landbird-on-water) are OOD. Split is \emph{spurious-correlation} shift: the background--label correlation in the training distribution is broken at test time, and the OOD test set is a known failure mode for ERM-trained classifiers.
    \item \textbf{CelebA} (WILDS subpopulation, \citealt{koh2021wilds,sagawa2020distributionally}). Face photographs encoded by frozen DINOv2 ViT-B/14 to 768-dim representations. \emph{ID/OOD:} with target $y$ = blond hair and attribute $a$ = male, the groups with $y = a$ (\emph{non-blond female / blond male}) are ID and the groups with $y \neq a$ (\emph{non-blond male / blond female}) are OOD. Split is a \emph{group-composition} shift that reverses the hair-color--gender pairing between ID and OOD; it is not the WILDS majority/minority split.
    \item \textbf{CivilComments} (WILDS, \citealt{koh2021wilds}). Online-comment text encoded by Jina-v3 to 1024-dim representations. \emph{ID/OOD:} comments with \emph{no} demographic-identity mention (LGBTQ, Muslim, Black, etc.) are ID; comments with at least one identity mention are OOD. Split is \emph{topic-stratum} shift: identity-mentioning comments are systematically more toxic in the training distribution, making the topic--toxicity correlation a known shortcut.
    \item \textbf{Amazon Reviews} (WILDS, \citealt{koh2021wilds}). Customer-review text encoded by Jina-v3 to 1024-dim representations, 5-class star rating. \emph{ID/OOD:} a held-out 20\% of training-set reviewers are ID; the WILDS test set, drawn from a disjoint reviewer pool, is OOD. Split is \emph{reviewer/domain} shift: different reviewers favor different vocabulary, sentence length, and star-distribution profiles.
\end{enumerate}

\subsection{Baselines}
\label{sec:baselines-detail}

The goal of every baseline below is the same as \cald's: given the unlabeled OOD pool described in \S\ref{sec:dataset-description}, decide whether the eval batch is drawn from the support distribution. Each baseline computes a scalar score per round; we calibrate a 5\%-FPR threshold on the ID-round scores and report TPR on the OOD-round scores. Notation: $\Xanc$ and $\yanc$ denote the labeled anchor batch with PFN softmax outputs $p_a \in \Delta^{C-1}$, $\Xeval$ denotes the eval batch with softmax outputs $p$, and $h(x) \in \mathbb{R}^d$ denotes the PFN's pre-decoder representation. We adapt the conventional ``low score = OOD'' direction so that AUROC is reported in $[0.5, 1]$; baseline implementations live at \texttt{detection/baselines/posthoc.py}.

\begin{enumerate}[leftmargin=1.4em]
    \item \textbf{MSP} \citep{hendrycks2017baseline}. Maximum softmax probability, $s_{\mathrm{MSP}}(x) = \max_c p(c \mid x)$. Confident ID predictions score high; uncertain OOD predictions score low.
    \item \textbf{Entropy} \citep{ovadia2019can}. Predictive entropy, $s_{\mathrm{Ent}}(x) = -\sum_c p(c\mid x) \log p(c\mid x)$. Higher entropy is taken as more OOD.
    \item \textbf{Energy} \citep{liu2020energy}. Negative log-sum-exp of the logits, $s_{\mathrm{En}}(x) = -\log \sum_c e^{z_c(x)}$, the softmax-free analogue of MSP. ID samples produce a peakier logit distribution and therefore a more negative energy.
    \item \textbf{Mahalanobis} \citep{lee2018simple}. With a class-conditional Gaussian fit on the support representations, $s_{\mathrm{MH}}(x) = \min_c (h(x) - \mu_c)^\top \Sigma^{-1} (h(x) - \mu_c)$. OOD samples lie far from every class cluster.
    \item \textbf{Relative Mahalanobis} (RMD, \citealt{ren2021simple}). RMD subtracts a class-agnostic background term, $s_{\mathrm{RMD}}(x) = \min_c \mathrm{MD}_c(x) - \mathrm{MD}_{\mathrm{bg}}(x)$ with $\mathrm{MD}_{\mathrm{bg}}(x) = (h(x) - \mu)^\top \Sigma^{-1} (h(x) - \mu)$. Cancelling population-level density isolates the discriminative signal and improves near-OOD detection where standard MD saturates.
    \item \textbf{KNN} \citep{sun2022knn}. Average $\ell_2$ distance from the eval representation to its $k=5$ nearest neighbors among the support representations, $s_{\mathrm{KNN}}(x) = \tfrac{1}{k}\sum_{i \in \mathcal{N}_k(x)} \|h(x) - h(x_i)\|_2$. A non-parametric distance-to-feature-cloud measure.
    \item \textbf{ViM} \citep{wang2022vim}. Virtual-Logit Matching decomposes pre-head representations into a classifier-subspace component and a null-space residual; the OOD score combines residual norm with energy, $s_{\mathrm{ViM}}(x) = \alpha \|\Pi^\perp (h(x) - u)\|_2 - \log \sum_c e^{z_c(x)}$, where $\Pi^\perp$ projects onto the bottom eigendirections of the ID covariance and $u = -W^+ b$ is the virtual origin. ID samples lie within the classifier's principal subspace; OOD samples carry significant null-space mass.
    \item \textbf{BBSD-Softmax} \citep{lipton2018detecting,rabanser2019failing}. Black-Box Shift Detection on softmax probabilities: per-class two-sample Kolmogorov--Smirnov tests between anchor and eval, take the minimum $p$-value, apply Bonferroni correction, return $1 - p_{\mathrm{corr}}$. Sensitive to label-marginal shifts.
    \item \textbf{BBSD-Embed} (BBSD-E). Same procedure as BBSD-Softmax but applied per-dimension on the pre-decoder embeddings $h(x)$ instead of the softmax. Sensitive to feature-distribution shifts that may not propagate to the labels.
    \item \textbf{D-MMD} \citep{gretton2012kernel,liu2020learning}. Maximum Mean Discrepancy between support and eval feature distributions, $\widehat{\mathrm{MMD}}^2(P, Q) = \tfrac{1}{n^2}\sum_{i,j} k(x_i, x_j) + \tfrac{1}{m^2}\sum_{i,j} k(y_i, y_j) - \tfrac{2}{nm}\sum_{i,j} k(x_i, y_j)$, with an RBF kernel at the median pairwise-distance bandwidth. The classical batch-level two-sample test.
    \item \textbf{H-Divergence} (H-Div, \citealt{bendavid2010theory,lopezpaz2017revisiting}). Trains a logistic-regression domain classifier to distinguish support representations from eval representations; the held-out classification accuracy serves as the OOD score. An accuracy near $0.5$ indicates indistinguishable distributions.
    \item \textbf{GradNorm} \citep{huang2021importance}. Gradient norm of the KL divergence to uniform, $s_{\mathrm{GN}}(x) = \|\nabla_W \mathrm{KL}(\mathbf{u} \,\|\, p_\theta(\cdot\mid x))\|_F$. Confident ID samples produce large gradients toward uniform; near-uniform OOD predictions produce small gradients.
    \item \textbf{GradOrth} \citep{behpour2023gradorth}. At the last linear layer $\nabla_W L = \mathrm{error} \otimes h$, so the gradient lives in the row span of the representations; GradOrth projects $h(x)$ onto the top-$k$ singular subspace of the ID-representation matrix and uses the projection norm as an ID-likelihood score (low projection $\Rightarrow$ OOD).
    \item \textbf{GAIA-Z} \citep{chen2023gaia}. Activity of the gradient of the maximum logit at an internal hidden activation, $s_{\mathrm{GAIA}}(x) = \mathbb{E}\big[\big|\partial \max_c z_c(x) / \partial h(x)\big|\big]$. Originally counts exact-zero gradient channels under ReLU; we adapt to the GELU decoder by reporting mean absolute gradient magnitude. ID samples activate more channels and yield larger mean activity.
    \item \textbf{WeiPer} \citep{granz2024weiper}. Mean MSP under random weight perturbations: for $r$ samples of unit-direction noise $\Delta_r$ scaled to $\delta \|w_c\|$ added to each row of the classifier head, $s_{\mathrm{WeiPer}}(x) = \tfrac{1}{r}\sum_r \max_c \mathrm{softmax}(W + \Delta_r) h(x)$. ID predictions are stable under perturbation and retain a high max-softmax; OOD predictions collapse.
\end{enumerate}

\subsection{Performance Deterioration}
\label{sec:deterioration-survey}

A deterioration detector is only meaningful when shift is genuinely harmful. Table~\ref{tab:deterioration} reports TabPFN accuracy on ID and OOD splits for all 14 datasets (10 rounds, eval size 128, $N=1{,}280$ pooled samples per dataset). To classify each shift as \emph{benign} or \emph{harmful} we use a principled threshold derived from Hoeffding's inequality: for a fixed classifier evaluated on $N$ i.i.d.\ samples,
\begin{equation}
  \Pr\!\left[\hat{a}_{\mathrm{ID}} - \hat{a}_{\mathrm{OOD}} > \Delta^* \;\Big|\; \text{no shift}\right] \leq e^{-N(\Delta^*)^2},
  \label{eq:hoeffding-harm}
\end{equation}
which is distribution-free and applies to any non-adaptive classifier. Setting the right-hand side to $\delta = 0.05$ gives
\begin{equation}
\Delta^*(N) = \sqrt{\ln(1/\delta)/N}.
\end{equation}
For $N = 1{,}280$ this yields $\Delta^* \approx 4.8\%$; any observed accuracy drop exceeding this threshold is statistically distinguishable from sampling noise at 95\% confidence under the null of no shift, and we therefore label the dataset as \textbf{harmful}. Datasets whose accuracy gap falls below $\Delta^*$ in absolute value (including those where OOD accuracy is \emph{higher} than ID) are labeled \textbf{benign}: the model's predictive integrity is preserved up to sampling noise, and a label-blind detector that fires on these datasets is incurring an unnecessary false alarm. The Hoeffding threshold is intentionally conservative: it makes no parametric assumption about the accuracy distribution, only bounded support, and is a per-dataset quantity ($\Delta^*$ scales as $1/\sqrt{N}$). Our benchmark spans the full severity spectrum, from catastrophic (\textbf{Covertype}, $\Delta = 83.5\%$) through moderate (\textbf{Camelyon17}, $\Delta = 8.9\%$) to negligible-or-benign (\textbf{Amazon}, $\Delta = +1.5\%$; \textbf{Adult Income}, $\Delta = -9.5\%$ where OOD accuracy is higher than ID), providing a demanding testbed for both correct alarm and correct silence.

\begin{table}[t]
\centering
\setlength{\tabcolsep}{5pt}
\small
\caption{\textbf{TabPFN accuracy under distribution shift, ordered by severity.}
$\Delta = \mathrm{acc}_{\mathrm{ID}} - \mathrm{acc}_{\mathrm{OOD}}$ (positive = harmful drop).
$\Delta^*(N) = \sqrt{\ln(1/\delta)/N}$ is the Hoeffding 95\%-confidence noise floor
(Eq.~\ref{eq:hoeffding-harm}, $\delta=0.05$, $N=1{,}280$); shifts with $|\Delta| \le \Delta^*$
are labeled \emph{benign} ($\dagger$).
Accuracy estimated over 10 rounds of 128 eval samples each.}
\label{tab:deterioration}
\begin{tabular}{lcccc}
\toprule
Dataset & ID Acc & OOD Acc & $\Delta$ (\%) & $\Delta^*$ (\%) \\
\midrule
\multicolumn{5}{l}{\textit{Benign} ($|\Delta| \le \Delta^*$)} \\
~~Adult Income$^\dagger$         & $0.820$ & $0.914$ & $-9.5$ & $4.8$ \\
~~Diabetes Readm.$^\dagger$      & $0.886$ & $0.909$ & $-2.3$ & $4.8$ \\
~~Amazon$^\dagger$               & $0.645$ & $0.630$ & $+1.5$ & $4.8$ \\
~~Credit Default$^\dagger$       & $0.847$ & $0.819$ & $+2.8$ & $4.8$ \\
~~Folktables$^\dagger$           & $0.799$ & $0.768$ & $+3.1$ & $4.8$ \\
~~Bank Marketing$^\dagger$       & $0.902$ & $0.859$ & $+4.4$ & $4.8$ \\
~~CelebA$^\dagger$               & $0.952$ & $0.908$ & $+4.5$ & $4.8$ \\
\midrule
\multicolumn{5}{l}{\textit{Harmful} ($\Delta > \Delta^*$)} \\
~~CivilComments                  & $0.918$ & $0.844$ & $+7.4$ & $4.8$ \\
~~Camelyon17                     & $0.953$ & $0.864$ & $+8.9$ & $4.8$ \\
~~Waterbirds                     & $0.988$ & $0.871$ & $+11.6$ & $4.8$ \\
~~CIFAR-10$\to$10.1              & $0.795$ & $0.670$ & $+12.4$ & $4.8$ \\
~~UCI Heart                      & $0.824$ & $0.668$ & $+15.5$ & $5.0$ \\
~~PACS                           & $0.998$ & $0.720$ & $+27.9$ & $4.8$ \\
~~Covertype                      & $0.904$ & $0.069$ & $+83.5$ & $4.8$ \\
\bottomrule
\end{tabular}
\end{table}

\paragraph{Adult Income deserves a remark.} It is the only dataset in our benchmark where OOD accuracy is substantially \emph{higher} than ID ($+9.5$\,pp); this is consistent with the underlying covariate shift (ID = Male, OOD = Female) being structurally easier under the 1994 Census labels (the female labor-force composition of that era leaves a sharper income decision boundary). Adult Income is therefore a particularly stringent benign test: a label-blind detector \emph{will} register the substantial covariate redistribution, but the predictive harm is not just absent; it is reversed. A well-calibrated harm detector should remain silent here.

\subsection{Main Results with Confidence Intervals}
\label{sec:main-results-full}

For completeness we report each cell of the main-paper Table~\ref{tab:baselines} as $\text{mean}_{\,\text{std}}$ over 10 seeds, split into the benign (Table~\ref{tab:baselines_full_benign}) and harmful (Table~\ref{tab:baselines_full_harmful}) halves. Bolding marks the best mean in each column.

\begin{table}[t]
\centering
\setlength{\tabcolsep}{4pt}
\renewcommand{\arraystretch}{1.05}
\scriptsize
\caption[Full TPR@5\% FPR results, benign shifts]{\textbf{Full TPR@5\%\,FPR results across 7 benign shifts ($\Delta\!\le\!4.8\%$, Hoeffding bound at $N\!=\!1{,}280$, Appendix~\ref{sec:deterioration-survey}; TPR = correct silence rate).} Each cell is $\text{mean}_{\,\text{std}}$ across 10 seeds (seeds 42--51); std is the spread of per-seed TPR (each seed's threshold is its own 5th-percentile ID-score). Means match Table~\ref{tab:baselines} exactly. Bold = best mean in column.}
\label{tab:baselines_full_benign}
\begin{tabular*}{\textwidth}{@{\extracolsep{\fill}}lccccccc@{}}
\toprule
Method & Adult & Diab. & Amazon & Credit & Folktbl. & Bank & CelebA \\
$\Delta$ (\%) & $-$9.4 & $-$2.3 & $+$1.5 & $+$2.8 & $+$3.1 & $+$4.4 & $+$4.5 \\
\midrule
MSP & $.00_{\,.00}$ & $.64_{\,.42}$ & $.78_{\,.18}$ & $.55_{\,.31}$ & $.53_{\,.28}$ & $.00_{\,.00}$ & $.63_{\,.26}$ \\
Entropy & $.00_{\,.00}$ & $.61_{\,.42}$ & $.80_{\,.15}$ & $.61_{\,.30}$ & $.50_{\,.24}$ & $.00_{\,.00}$ & $.39_{\,.22}$ \\
Energy & $.00_{\,.00}$ & $.00_{\,.00}$ & $.77_{\,.19}$ & $.00_{\,.00}$ & $.86_{\,.17}$ & $\mathbf{.42}_{\,.45}$ & $.04_{\,.07}$ \\
Mahal. & $.00_{\,.00}$ & $.00_{\,.00}$ & $.71_{\,.19}$ & $.00_{\,.00}$ & $.12_{\,.11}$ & $.05_{\,.08}$ & $.00_{\,.00}$ \\
RMD & $.69_{\,.32}$ & $.59_{\,.39}$ & $.87_{\,.13}$ & $\mathbf{.92}_{\,.11}$ & $.72_{\,.24}$ & $.05_{\,.08}$ & $.42_{\,.20}$ \\
KNN & $.00_{\,.00}$ & $.00_{\,.00}$ & $.78_{\,.15}$ & $.00_{\,.00}$ & $.23_{\,.13}$ & $.00_{\,.00}$ & $.00_{\,.00}$ \\
ViM & $.00_{\,.00}$ & $.10_{\,.30}$ & $.87_{\,.13}$ & $.00_{\,.00}$ & $.63_{\,.28}$ & $.13_{\,.13}$ & $.81_{\,.23}$ \\
BBSD-S & $.00_{\,.00}$ & $.40_{\,.36}$ & $.75_{\,.29}$ & $.76_{\,.29}$ & $.71_{\,.21}$ & $.02_{\,.06}$ & $.64_{\,.40}$ \\
BBSD-E & $.00_{\,.00}$ & $.00_{\,.00}$ & $.76_{\,.27}$ & $.67_{\,.28}$ & $.66_{\,.34}$ & $.00_{\,.00}$ & $.49_{\,.43}$ \\
D-MMD & $.00_{\,.00}$ & $.00_{\,.00}$ & $.66_{\,.14}$ & $.00_{\,.00}$ & $.00_{\,.00}$ & $.00_{\,.00}$ & $.54_{\,.32}$ \\
H-Div & $.00_{\,.00}$ & $.00_{\,.00}$ & $.19_{\,.38}$ & $.00_{\,.00}$ & $.10_{\,.18}$ & $.32_{\,.17}$ & $.00_{\,.00}$ \\
GradNorm & $.00_{\,.00}$ & $.44_{\,.47}$ & $.85_{\,.19}$ & $.01_{\,.03}$ & $.78_{\,.18}$ & $.27_{\,.42}$ & $.60_{\,.19}$ \\
GradOrth & $.83_{\,.21}$ & $\mathbf{.77}_{\,.18}$ & $.83_{\,.17}$ & $.69_{\,.18}$ & $.86_{\,.18}$ & $.02_{\,.06}$ & $.46_{\,.28}$ \\
GAIA-z & $.00_{\,.00}$ & $.07_{\,.21}$ & --- & $.01_{\,.03}$ & $.80_{\,.20}$ & $.29_{\,.36}$ & --- \\
WeiPer & $.18_{\,.36}$ & $.17_{\,.35}$ & $.89_{\,.10}$ & $.00_{\,.00}$ & $.67_{\,.30}$ & $.17_{\,.35}$ & $.81_{\,.19}$ \\
\midrule
\textsc{Cald} (TabPFN) & $\mathbf{1.00}_{\,.00}$ & $.08_{\,.25}$ & $\mathbf{.92}_{\,.04}$ & $.00_{\,.00}$ & $\mathbf{.98}_{\,.03}$ & $.02_{\,.04}$ & $.27_{\,.28}$ \\
\textsc{Cald} (TabICL) & $\mathbf{1.00}_{\,.00}$ & $.00_{\,.00}$ & $.87_{\,.14}$ & $.76_{\,.12}$ & $.42_{\,.15}$ & $.00_{\,.00}$ & $\mathbf{.97}_{\,.05}$ \\
\bottomrule
\end{tabular*}
\end{table}

\begin{table}[b]
\centering
\setlength{\tabcolsep}{4pt}
\renewcommand{\arraystretch}{1.05}
\scriptsize
\caption[Full TPR@5\% FPR results, harmful shifts]{\textbf{Full TPR@5\%\,FPR results across 7 harmful shifts ($\Delta\!>\!4.8\%$, Hoeffding bound at $N\!=\!1{,}280$, Appendix~\ref{sec:deterioration-survey}; TPR = correct alarm rate).} Each cell is $\text{mean}_{\,\text{std}}$ across 10 seeds (seeds 42--51); std is the spread of per-seed TPR (each seed's threshold is its own 5th-percentile ID-score). Means match Table~\ref{tab:baselines} exactly. Bold = best mean in column.}
\label{tab:baselines_full_harmful}
\begin{tabular*}{\textwidth}{@{\extracolsep{\fill}}lccccccc@{}}
\toprule
Method & CivilC. & Camel. & Waterb. & CIFAR & UCI & PACS & Covert. \\
$\Delta$ (\%) & $+$7.4 & $+$8.9 & $+$11.6 & $+$12.4 & $+$15.6 & $+$27.9 & $+$83.5 \\
\midrule
MSP & $\mathbf{1.00}_{\,.00}$ & $.97_{\,.06}$ & $\mathbf{1.00}_{\,.00}$ & $.96_{\,.09}$ & $.74_{\,.32}$ & $\mathbf{1.00}_{\,.00}$ & $\mathbf{1.00}_{\,.00}$ \\
Entropy & $\mathbf{1.00}_{\,.00}$ & $.98_{\,.04}$ & $\mathbf{1.00}_{\,.00}$ & $.97_{\,.09}$ & $.77_{\,.31}$ & $\mathbf{1.00}_{\,.00}$ & $\mathbf{1.00}_{\,.00}$ \\
Energy & $\mathbf{1.00}_{\,.00}$ & $\mathbf{1.00}_{\,.00}$ & $\mathbf{1.00}_{\,.00}$ & $.97_{\,.09}$ & $.66_{\,.42}$ & $\mathbf{1.00}_{\,.00}$ & $.50_{\,.50}$ \\
Mahal. & $.32_{\,.19}$ & $.43_{\,.16}$ & $.18_{\,.13}$ & $.55_{\,.20}$ & $\mathbf{1.00}_{\,.00}$ & $\mathbf{1.00}_{\,.00}$ & $\mathbf{1.00}_{\,.00}$ \\
RMD & $.51_{\,.30}$ & $.40_{\,.38}$ & $.52_{\,.27}$ & $.67_{\,.25}$ & $.47_{\,.36}$ & $\mathbf{1.00}_{\,.00}$ & $\mathbf{1.00}_{\,.00}$ \\
KNN & $\mathbf{1.00}_{\,.00}$ & $.95_{\,.08}$ & $.97_{\,.05}$ & $.10_{\,.10}$ & $.99_{\,.03}$ & $\mathbf{1.00}_{\,.00}$ & $\mathbf{1.00}_{\,.00}$ \\
ViM & $.23_{\,.26}$ & $.54_{\,.36}$ & $.18_{\,.22}$ & $.15_{\,.11}$ & $.93_{\,.21}$ & $.70_{\,.36}$ & $\mathbf{1.00}_{\,.00}$ \\
BBSD-S & $.61_{\,.40}$ & $.94_{\,.15}$ & $.88_{\,.26}$ & $.89_{\,.10}$ & $.81_{\,.34}$ & $\mathbf{1.00}_{\,.00}$ & $\mathbf{1.00}_{\,.00}$ \\
BBSD-E & $.58_{\,.25}$ & $\mathbf{1.00}_{\,.00}$ & $.99_{\,.03}$ & $.72_{\,.24}$ & $\mathbf{1.00}_{\,.00}$ & $\mathbf{1.00}_{\,.00}$ & $\mathbf{1.00}_{\,.00}$ \\
D-MMD & $.91_{\,.21}$ & $\mathbf{1.00}_{\,.00}$ & $\mathbf{1.00}_{\,.00}$ & $.68_{\,.30}$ & $\mathbf{1.00}_{\,.00}$ & $\mathbf{1.00}_{\,.00}$ & $\mathbf{1.00}_{\,.00}$ \\
H-Div & $.70_{\,.27}$ & $\mathbf{1.00}_{\,.00}$ & $\mathbf{1.00}_{\,.00}$ & $.50_{\,.17}$ & $\mathbf{1.00}_{\,.00}$ & $\mathbf{1.00}_{\,.00}$ & $\mathbf{1.00}_{\,.00}$ \\
GradNorm & $\mathbf{1.00}_{\,.00}$ & $.92_{\,.14}$ & $.98_{\,.06}$ & $.91_{\,.15}$ & $.54_{\,.42}$ & $.98_{\,.06}$ & $.70_{\,.46}$ \\
GradOrth & $.66_{\,.24}$ & $.54_{\,.31}$ & $\mathbf{1.00}_{\,.00}$ & $.95_{\,.08}$ & $.23_{\,.28}$ & $.97_{\,.06}$ & $\mathbf{1.00}_{\,.00}$ \\
GAIA-z & --- & --- & --- & --- & $.41_{\,.42}$ & --- & $.59_{\,.48}$ \\
WeiPer & $.34_{\,.36}$ & $.45_{\,.42}$ & $.26_{\,.27}$ & $.21_{\,.18}$ & $.89_{\,.30}$ & $.62_{\,.42}$ & $.73_{\,.42}$ \\
\midrule
\textsc{Cald} (TabPFN) & $1.00_{\,.01}$ & $.19_{\,.13}$ & $.46_{\,.46}$ & $\mathbf{1.00}_{\,.01}$ & $.71_{\,.35}$ & $\mathbf{1.00}_{\,.00}$ & $\mathbf{1.00}_{\,.00}$ \\
\textsc{Cald} (TabICL) & $.99_{\,.02}$ & $\mathbf{1.00}_{\,.00}$ & $.99_{\,.02}$ & $.99_{\,.02}$ & $.71_{\,.32}$ & $\mathbf{1.00}_{\,.00}$ & $\mathbf{1.00}_{\,.00}$ \\
\bottomrule
\end{tabular*}
\end{table}

\paragraph{Protocol.}
For each (dataset, method) cell we run \textbf{10 independent seeds} (seeds 42--51); within each seed we run \textbf{50 calibration rounds} and \textbf{50 deployment rounds} of Algorithm~\ref{alg:method}. Each round draws a fresh ID anchor / eval pair (calibration) or a fresh OOD eval batch (deployment) of size $|\Xeval|=128$, with the support set $\cD^{\mathrm{sup}}$ (size $n=1024$) and the labeled anchor $(\Xanc, \yanc)$ ($|\Xanc|=128$) held fixed within the seed. The detection threshold is set per seed at the empirical 5th percentile of that seed's 50 calibration anchor-loss values, and the per-seed TPR is the fraction of the 50 deployment anchor-losses that fall below it. The reported cell is the mean and standard deviation of the 10 per-seed TPRs and, crucially, \emph{not} a pooled estimate across seeds, because different seeds draw different supports and thus have anchor losses on different scales. Baselines (\S\ref{sec:baselines-detail}) follow the same per-seed thresholding protocol on their own scalar score, again calibrated at the seed's own 5th-percentile ID-side cutoff. AUROC, when reported, is computed on the union of the seed's $50$ ID and $50$ OOD scores and then averaged across the 10 seeds.

\paragraph{LR-stability protocol for \cald.}
For each (dataset, method) cell we sweep \cald's finetune learning rate over $\eta \in \{10^{-2}, 10^{-3}, 10^{-4}\}$ and \emph{select the largest $\eta$ that produces a stable signal}, where stability is defined on the per-step direction indicator
\begin{equation}
    d_t \;:=\; \mathbb{1}\!\left\{\overline{\lambda^{\mathrm{OOD}}_t} < \overline{\lambda^{\mathrm{ID}}_t}\right\}, \qquad t = 1, \ldots, T,
\end{equation}
where $\overline{\lambda^{\mathrm{ID}}_t}$ and $\overline{\lambda^{\mathrm{OOD}}_t}$ are the round-averaged anchor losses at step $t$. We accept the candidate $\eta$ if and only if $d_t$ flips at most twice over $t \in [30, 50]$ (i.e.\ the trajectory has settled into a single direction in the second half of finetuning), \emph{regardless of which direction it has settled into}. If no LR in the sweep is stable we mark the cell unstable (flagged `!' in the direction column of Table~\ref{tab:detection}); otherwise we report the AUROC and TPR at the largest stable $\eta$.

The stability criterion is \emph{not} a post-hoc filter for direction. \cald only fires when $\lambda_{\mathrm{OOD}} < \lambda_{\mathrm{ID}}$ (Eq.~\ref{eq:verdict}); a stable settled-into-$\lambda_{\mathrm{OOD}} > \lambda_{\mathrm{ID}}$ trajectory simply produces a low TPR, which is what we report. The reason we filter on stability rather than direction is that an unstable trajectory, one whose $d_t$ oscillates between $0$ and $1$ across the second-half steps, contains no reliable signal. Reading off $\lambda$ at any single step in such a trajectory is essentially a coin flip on transient gradient noise: stop one step earlier or later and the test verdict reverses. In particular, at high learning rates the ID/OOD trajectories cross each other multiple times before $t = T$, so the value of $\lambda_T = \cL_{\mathrm{anc}}(\theta_T)$ depends on whether $T$ happens to land in a momentary alignment of the gradient cross-term. Reporting that single sample without checking for oscillation would be reading a randomly-timed slice of a chaotic curve, effectively ``getting lucky'' that the finetune horizon ends inside a brief favorable window. The stability criterion ensures that whatever direction we report, the trajectory has stopped flipping by step $30$ and the value of $\lambda_T$ reflects a converged regime rather than an arbitrary point in an oscillating one.

In practice, $\eta = 10^{-3}$ is the modal choice across the 14 datasets: $\eta = 10^{-2}$ is rejected on the more sensitive datasets (notably CelebA and CivilComments under TabPFN, where the joint loss oscillates across iterations 30--50) and $\eta = 10^{-4}$ is the fallback on four cells (Adult Income under both models, CelebA-TabICL, and Folktables-TabPFN), where neither $10^{-2}$ nor $10^{-3}$ stabilizes; Appendix~\ref{sec:abl-frozen} reports the census. The dataset-by-dataset selected LRs and the stability margins are produced by \texttt{detection/viz/cache\_results.py} and consumed by the main-paper table builder.

\subsection{Comparison to Drift-Resilient TabPFN}
\label{sec:drift-resilient}

A natural alternative to \emph{detecting} distribution shift is to \emph{adapt} the model so that shift no longer harms predictions. Drift-Resilient TabPFN~\citep{helli2024driftresilient} pursues this route: it retrains the PFN prior on synthetic data sampled from \emph{evolving} structural causal models, producing a tabular foundation model whose predictions are intended to be robust under covariate shift. If this adaptation were sufficient on the benchmark of \S\ref{sec:deterioration-survey}, the failure mode that motivates \cald would not arise, and a detector would be redundant. We therefore evaluate whether Drift-Resilient TabPFN closes the ID/OOD accuracy gap on our benchmark; closing the gap is a \emph{necessary} condition for the claim that adaptation removes the need for a detector.

Table~\ref{tab:drift-comparison} repeats the deterioration survey of Table~\ref{tab:deterioration} using Drift-Resilient TabPFN (single-estimator configuration, matching standard TabPFN) on the same ID/OOD splits, support sizes, and round structure. The right-most column reports $\delta\mathrm{OOD} = \mathrm{acc}^{\mathrm{drift}}_{\mathrm{OOD}} - \mathrm{acc}^{\mathrm{std}}_{\mathrm{OOD}}$, which isolates the effect of drift adaptation on OOD accuracy. Across all 15 datasets of Table~\ref{tab:drift-comparison} (including the Camelyon17 ResNet18 variant) the mean $\delta\mathrm{OOD}$ is $-1.4$\,pp; on the eight datasets classified as harmful in Table~\ref{tab:deterioration} the mean is $-2.2$\,pp. Drift adaptation strictly improves OOD accuracy on only two harmful datasets (Covertype, $+3.8$\,pp; Camelyon17 ResNet18, $+3.0$\,pp), and \emph{degrades} OOD accuracy by more than the Hoeffding noise floor ($\Delta^* \approx 4.8$\,pp, Eq.~\ref{eq:hoeffding-harm}) on three (PACS $-8.0$\,pp, CIFAR-10 $-7.8$\,pp, Waterbirds $-4.8$\,pp). The dataset with the largest standard-TabPFN deterioration after Covertype, PACS at $\Delta = +27.9$\,pp, is made measurably \emph{worse} by adaptation rather than better, and ID accuracy is also slightly degraded on a majority of datasets (mean $-1.0$\,pp). The harm classification of Table~\ref{tab:deterioration} is essentially unchanged: every dataset that was harmful under standard TabPFN remains harmful under Drift-Resilient TabPFN by the same Hoeffding criterion. The headline finding is thus that drift adaptation does \emph{not} eliminate the deterioration that \cald is designed to detect; on the present benchmark it slightly aggravates it on average.

\begin{table}[t]
\centering
\setlength{\tabcolsep}{4pt}
\small
\caption{\textbf{TabPFN vs.\ Drift-Resilient TabPFN accuracy under distribution shift.}
Standard TabPFN (left) and Drift-Resilient TabPFN~\citep{helli2024driftresilient} (right)
on the same ID/OOD splits.
$\Delta = \mathrm{acc}_{\mathrm{ID}} - \mathrm{acc}_{\mathrm{OOD}}$ (positive $=$ harmful drop).
$\delta\mathrm{OOD} = \mathrm{acc}^{\mathrm{drift}}_{\mathrm{OOD}} - \mathrm{acc}^{\mathrm{std}}_{\mathrm{OOD}}$
(positive $=$ drift model improves OOD accuracy).
Harm threshold $\Delta^* \approx 4.8\%$ (Hoeffding, $\delta=0.05$, $N=1{,}280$);
benign datasets marked $\dagger$.}
\label{tab:drift-comparison}
\begin{tabular}{lcccccccc}
\toprule
& \multicolumn{3}{c}{\textbf{Standard TabPFN}} & \multicolumn{3}{c}{\textbf{Drift-Resilient TabPFN}} & \\
\cmidrule(lr){2-4}\cmidrule(lr){5-7}
Dataset & ID & OOD & $\Delta$\,(\%) & ID & OOD & $\Delta$\,(\%) & $\delta$OOD\,(\%) \\
\midrule
\multicolumn{8}{l}{\textit{Benign} ($\Delta \le \Delta^*$)} \\
~~CelebA$^\dagger$ & $0.952$ & $0.908$ & $+4.5$ & $0.946$ & $0.873$ & $+7.3$ & $-3.4$ \\
~~Bank Marketing$^\dagger$ & $0.902$ & $0.859$ & $+4.4$ & $0.901$ & $0.859$ & $+4.1$ & $+0.1$ \\
~~Folktables$^\dagger$ & $0.799$ & $0.768$ & $+3.1$ & $0.800$ & $0.766$ & $+3.4$ & $-0.2$ \\
~~Credit Default$^\dagger$ & $0.847$ & $0.819$ & $+2.8$ & $0.842$ & $0.815$ & $+2.7$ & $-0.4$ \\
~~Amazon$^\dagger$ & $0.645$ & $0.630$ & $+1.5$ & $0.633$ & $0.644$ & $-1.1$ & $+1.3$ \\
~~Diabetes Readm.$^\dagger$ & $0.886$ & $0.909$ & $-2.3$ & $0.886$ & $0.909$ & $-2.3$ & $+0.0$ \\
~~Adult Income$^\dagger$ & $0.820$ & $0.914$ & $-9.5$ & $0.800$ & $0.912$ & $-11.2$ & $-0.2$ \\
\midrule
\multicolumn{8}{l}{\textit{Harmful} ($\Delta > \Delta^*$)} \\
~~Covertype & $0.904$ & $0.069$ & $+83.5$ & $0.777$ & $0.106$ & $+67.1$ & $\mathbf{+3.8}$ \\
~~PACS & $0.998$ & $0.720$ & $+27.9$ & $1.000$ & $0.640$ & $+36.0$ & $-8.0$ \\
~~UCI Heart & $0.824$ & $0.668$ & $+15.5$ & $0.818$ & $0.656$ & $+16.2$ & $-1.2$ \\
~~CIFAR-10$\to$10.1 & $0.795$ & $0.670$ & $+12.4$ & $0.741$ & $0.592$ & $+14.9$ & $-7.8$ \\
~~Waterbirds & $0.988$ & $0.871$ & $+11.6$ & $0.987$ & $0.823$ & $+16.4$ & $-4.8$ \\
~~Camelyon17 (DINOv2) & $0.953$ & $0.864$ & $+8.9$ & $0.941$ & $0.847$ & $+9.4$ & $-1.7$ \\
~~Camelyon17 (ResNet18) & $0.923$ & $0.840$ & $+8.3$ & $0.921$ & $0.870$ & $+5.2$ & $\mathbf{+3.0}$ \\
~~CivilComments & $0.918$ & $0.844$ & $+7.4$ & $0.915$ & $0.836$ & $+7.9$ & $-0.8$ \\
\bottomrule
\end{tabular}
\end{table}

The result is consistent with, and indeed predicted by, the design of Drift-Resilient TabPFN. Its prior is constructed by sampling structural causal models and then perturbing their parameters along a temporal axis to simulate \emph{gradual evolution} of the data-generating process. The resulting model is therefore tuned to a specific, structured family of shifts (smoothly drifting SCMs, of the kind found in the Wild-Time benchmarks for which the method was developed) and not to the heterogeneous covariate shifts in our benchmark, which include image-domain transfer (Camelyon17, Waterbirds, PACS), feature-space stratification (Covertype's elevation cut, UCI Heart's hospital split), and large textual-distribution gaps (CivilComments). On several datasets (PACS, CIFAR-10, Waterbirds) the perturbed prior actively misallocates predictive mass, assuming the wrong drift mechanism, which manifests directly as degraded OOD accuracy. More fundamentally, no fixed-prior model can be uniformly robust to arbitrary covariate shift: by a no-free-lunch argument, any inductive bias added to handle one family of shifts must be paid for on another. This is the structural reason adaptation cannot \emph{replace} detection: a deployed practitioner has no way to verify, a priori, that the shift at hand falls within the family the adapted prior was trained for. \cald and Drift-Resilient TabPFN are therefore complementary in their roles, with the detector indicating \emph{whether} a shift has occurred while the adapter (when applicable) indicates \emph{how} to respond. The empirical evidence in Table~\ref{tab:drift-comparison} suggests that on a heterogeneous covariate-shift benchmark, detection is the more dependable of the two.

\subsection{Per-Dataset Detection Performance of $\hat\lambda$}
\label{sec:loss-anchor-detail}

Table~\ref{tab:detection} reports AUROC and TPR@5\%\,FPR for the post-finetune anchor-loss statistic $\hat\lambda$ alone (without baselines) across all datasets and both models, complementing the baseline comparison in the main paper. The protocol matches Table~\ref{tab:baselines}: seeds 42--51, $K=50$ calibration rounds, per-seed 5\%-quantile thresholds, learning rate selected per cell by the tail-stability rule.

\begin{table}[t]
\centering
\setlength{\tabcolsep}{4pt}
\small
\caption{\textbf{Per-dataset detection performance of the post-finetune anchor loss $\hat\lambda$.} AUROC and TPR@5\%\,FPR for distinguishing ID from OOD eval batches, under the same protocol as Table~\ref{tab:baselines} (seeds 42--51, $K=50$ calibration rounds, per-seed 5\%-quantile thresholds, learning rate selected per cell by the tail-stability rule). Direction \checkmark\,: OOD $<$ ID as expected; \textbf{!}: reversed for the majority of seeds. All four reversed cells are benign shifts and coincide with the cells where the learning-rate stability rule falls back to $\eta = 10^{-4}$. The per-cell selection lies outside the conformal guarantee of Proposition~\ref{prop:conformal-main}, which covers a configuration fixed in advance; the certified frozen-$\eta = 10^{-3}$ configuration is reported alongside these results in Appendix~\ref{sec:abl-frozen}.}
\vspace{0.2cm}
\label{tab:detection}
\begin{tabular}{llccc}
\toprule
Dataset & Model & AUROC & TPR@5\% & Dir. \\
\midrule
\multicolumn{5}{l}{\textit{Image}} \\
~~Camelyon17 (DINOv2)  & TabPFN & 0.675 & 0.192 & \checkmark \\
  & TabICL & \textbf{1.000} & 1.000 & \checkmark \\
~~Camelyon17 (ResNet18)  & TabPFN & 0.977 & 0.914 & \checkmark \\
  & TabICL & 0.990 & 0.932 & \checkmark \\
~~CIFAR-10$\to$10.1  & TabPFN & \textbf{1.000} & 0.998 & \checkmark \\
  & TabICL & 0.996 & 0.990 & \checkmark \\
~~PACS  & TabPFN & \textbf{1.000} & 1.000 & \checkmark \\
  & TabICL & \textbf{1.000} & 1.000 & \checkmark \\
\midrule
\multicolumn{5}{l}{\textit{Tabular}} \\
~~UCI Heart  & TabPFN & 0.908 & 0.713 & \checkmark \\
  & TabICL & 0.930 & 0.714 & \checkmark \\
~~Covertype  & TabPFN & \textbf{1.000} & 1.000 & \checkmark \\
  & TabICL & \textbf{1.000} & 1.000 & \checkmark \\
~~Credit Default  & TabPFN & \textbf{1.000} & 1.000 & \checkmark \\
  & TabICL & 0.698 & 0.242 & \checkmark \\
~~Folktables  & TabPFN & 0.624 & 0.020 & \textbf{!} \\
  & TabICL & 0.899 & 0.578 & \checkmark \\
~~Bank Marketing  & TabPFN & 0.997 & 0.978 & \checkmark \\
  & TabICL & \textbf{1.000} & 1.000 & \checkmark \\
~~Diabetes Readm.  & TabPFN & 0.976 & 0.918 & \checkmark \\
  & TabICL & \textbf{1.000} & 1.000 & \checkmark \\
~~Adult Income  & TabPFN & \textbf{1.000} & 0.000 & \textbf{!} \\
  & TabICL & \textbf{1.000} & 0.000 & \textbf{!} \\
\midrule
\multicolumn{5}{l}{\textit{Subpop.}} \\
~~Waterbirds  & TabPFN & 0.921 & 0.462 & \checkmark \\
  & TabICL & 0.993 & 0.986 & \checkmark \\
~~CelebA  & TabPFN & 0.919 & 0.730 & \checkmark \\
  & TabICL & 0.845 & 0.030 & \textbf{!} \\
~~CivilComments  & TabPFN & 0.999 & 0.998 & \checkmark \\
  & TabICL & 0.997 & 0.987 & \checkmark \\
~~Amazon  & TabPFN & 0.550 & 0.082 & \checkmark \\
  & TabICL & 0.600 & 0.127 & \checkmark \\
\bottomrule
\end{tabular}
\end{table}

\newpage
\section{Ablations}
\label{sec:appendix-ablations}

This appendix collects the hyperparameter sweeps, design-choice ablations, and the further ablations deferred from \S\ref{sec:experiments}: the disagreement weight, the finetuning budget and its deployment cost, the frozen deployment configuration, anchor robustness, feature collapse, the theorem's condition and premise, the two-signal detector, and detectability across shift types.

\subsection{Hyperparameter Sweeps}
\label{sec:appendix-hyperparam}

We sweep four hyperparameters of \cald on three representative datasets covering catastrophic, moderate, and structured-image shift: Covertype, Credit Default, and CIFAR-10. The point of the sweeps is twofold: to check that the detector is not living on a knife-edge of its compute budget, and to verify that the operating regime under which the theory applies (small-step finetuning, modest support window) is the regime under which the practical detector actually performs well.

\paragraph{Number of finetune steps $T$.} The natural worry is that we have tuned $T$ to a sweet spot beyond which the signal collapses; if so, the theorem's first-order analysis would be tracking a transient that the deployed detector exits at $T = T_{\mathrm{deploy}}$. AUROC is in fact largely insensitive to $T$ on strong shifts (Covertype, Credit Default: $1.000$ at every $T \in \{1, 3, 5, 10, 20, 50\}$). On the subtler CIFAR-10 shift more steps help, with AUROC rising from $0.522$ at $T=1$ to $0.992$ at $T=50$. The default $T=50$ sits well within the small-step regime that Theorem~\ref{thm:main-separation-main}'s remainder bound requires, and the ID/OOD separation does not depend on a finely tuned step count.

\paragraph{Learning rate $\eta$.} Theorem~\ref{thm:main-separation-main} gives a first-order guarantee that becomes vacuous if the second-order remainder $O(T^2 \eta^2)$ catches up with the leading $T \eta$ term; in plain words, the analysis only applies when $\eta$ is small. Sweeping over a three-decade range $\eta \in \{10^{-4}, 10^{-3}, 5 \times 10^{-3}, 10^{-2}, 5 \times 10^{-2}, 10^{-1}\}$, Credit Default and Covertype keep AUROC at $1.000$ throughout, and CIFAR-10 degrades only at very low $\eta$ ($< 5 \times 10^{-3}$) where the steps are too small to move the decoder into the disagreement regime at all. The detector therefore inhabits the small-step regime under which the theory applies, and the Algorithm~\ref{alg:method} LR-stability heuristic can pick within that regime with substantial slack.

\paragraph{Support size $n$.} The PFN's in-context support set is the conditioning information the entire detector reads against, and using more of it costs both quadratic attention compute and a longer prompt. We sweep $n$ to see how small the in-context window can be before the detection signal degrades. Performance falls gracefully at $n = 100$ (AUROC $0.762$--$0.914$) but recovers fully by $n = 250$, so a modest in-context window suffices to anchor the detector. This rules out the alternative that \cald's separation is a side-effect of the PFN being asked to solve a particularly large in-context task.

\paragraph{Anchor size $|\Xanc|$.} The anchor batch is the calibration substrate against which the post-finetune $\lambda$ is scored, so its size enters the variance of the calibration record and the conformal threshold from Proposition~\ref{prop:conformal-main}. \cald is robust to anchor size across $|\Xanc| \in \{16, 32, 64, 128, 256\}$, with AUROC $> 0.99$ in all but one cell. The anchor set therefore does not need to be large to serve as a reliable ID reference, and the calibration cost of the method is bounded by the eval-batch geometry rather than by the held-out anchor.

\begin{figure}[t]
    \centering
    \includegraphics[width=0.7\linewidth]{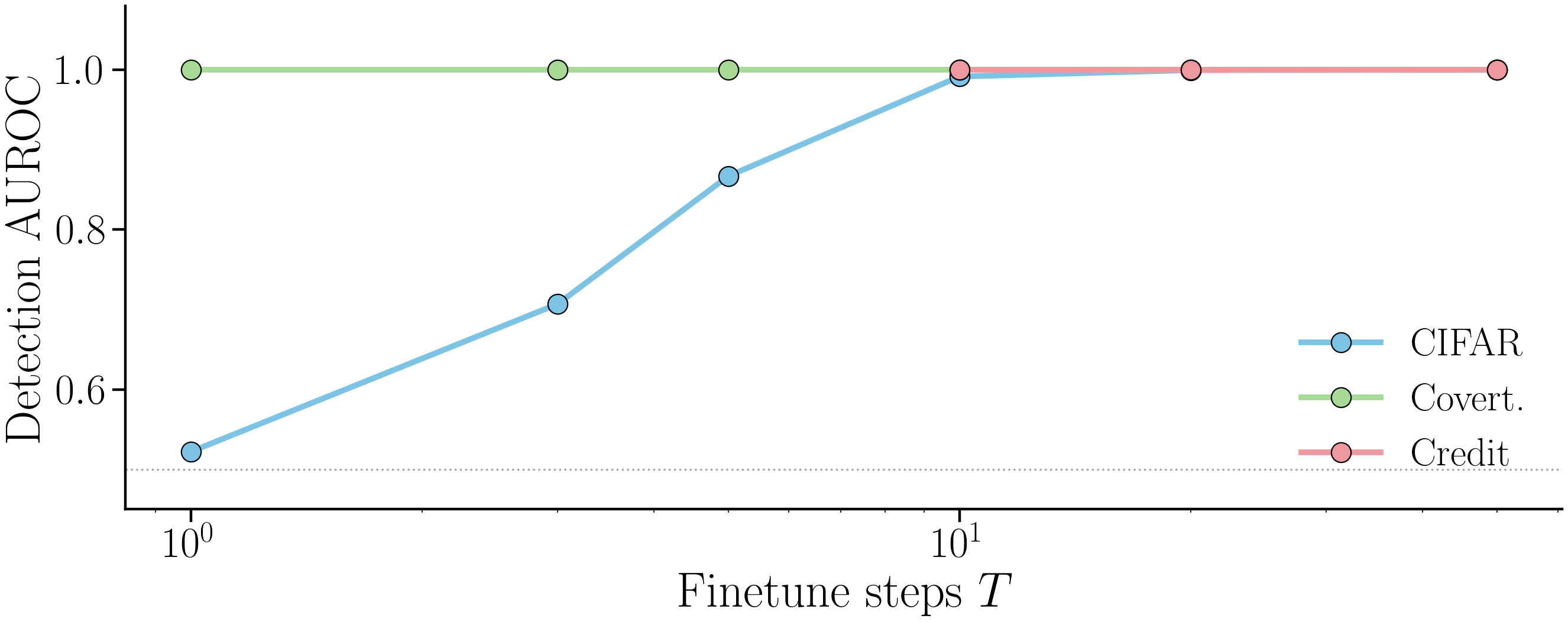}
    \caption{\textbf{Ablation: number of gradient steps $T$.}
    Detection AUROC vs.\ $T$ on three representative datasets. Covertype and Credit Default saturate immediately; CIFAR-10 improves with more steps.}
    \label{fig:ablation_nsteps}
\end{figure}

\subsection{Design-Choice Ablations}
\label{sec:appendix-design}

\begin{figure}[t]
    \centering
    \includegraphics[width=0.92\linewidth]{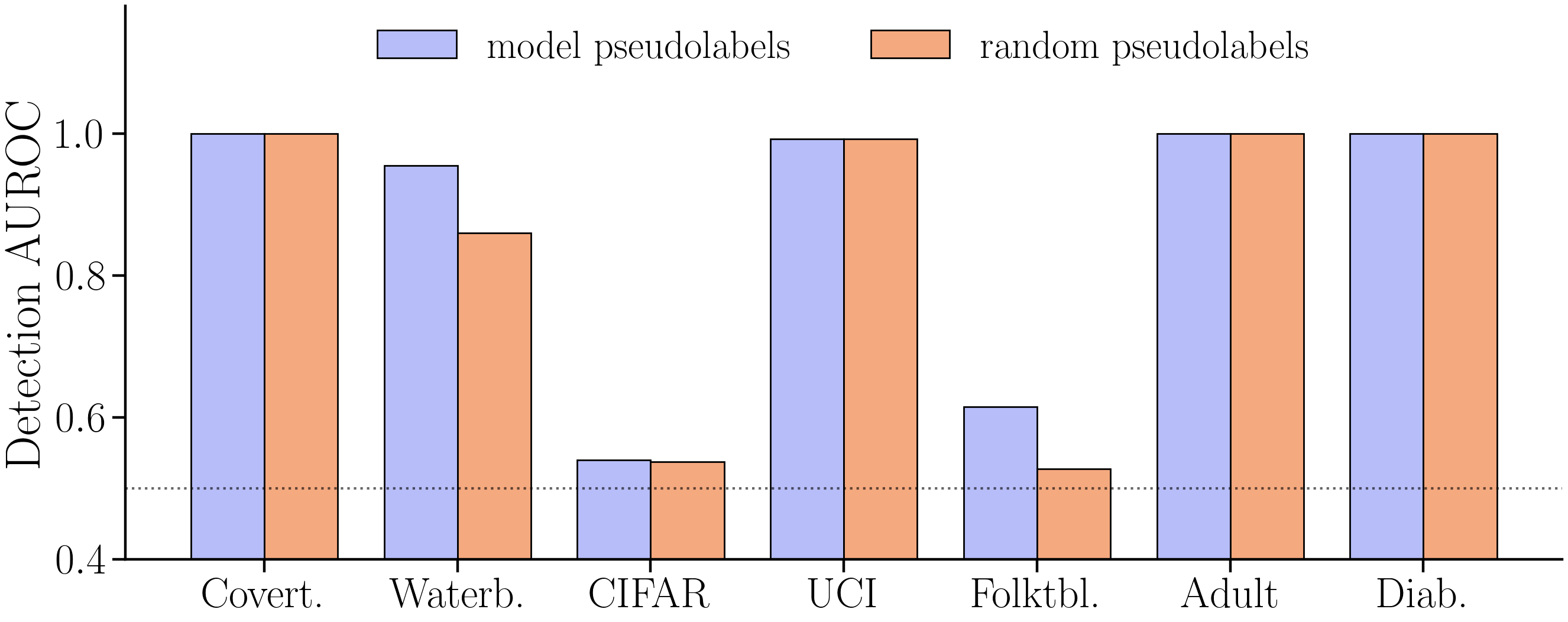}
    \caption{\textbf{Detection AUROC with model-pseudolabels vs.\ random pseudolabels.} Harm-adjusted; dotted line marks chance ($0.5$). Random pseudolabels leave detection unchanged on strong-shift datasets but cost $\sim$10 points of AUROC on moderate-shift datasets (Waterbirds, Folktables) where the PFN is less confident on its own pseudolabels.}
    \label{fig:pred5_random_pseudo}
\end{figure}

\paragraph{Pseudolabel choice.}
\cald's disagree term pushes the eval predictions away from the model's own pseudolabels $\hat{y} = \arg\max\, q_{\theta_0}(y \mid \Xeval, \cD^{\mathrm{sup}})$. Why use the model's pseudolabels rather than any other arbitrary target? If \emph{any} target produced the same ID/OOD separation, the signal would be coming from the disagreement objective in general rather than from the specific direction in label space that a confident model points to. Replacing $\hat{y}$ with random labels drawn uniformly over the $C$ classes (Figure~\ref{fig:pred5_random_pseudo}) leaves AUROC unchanged on the datasets with large covariate shift (Covertype, UCI Heart, Adult, Diabetes Readmission), where any non-degenerate target is far enough from the posterior to drive the disagree gradient. On the moderate-shift datasets where the pretrained PFN is less confident on its own pseudolabels (Waterbirds and Folktables), random pseudolabels cost $9$--$10$ AUROC points. The theoretical residual $r_e = p - u_{\hat{y}}$ depends on $\hat{y}$ matching the modal class prediction; the degradation is therefore concentrated in exactly the regime where the modal-class structure is the key component of the signal, consistent with the gradient form derived in \S\ref{sec:method-theory}.

\paragraph{Pseudolabel scope.}
A natural alternative to freezing $\hat{y}$ at $\theta_0$ is to recompute pseudolabels each step, so the disagree term always points away from the \emph{current} model's predictions. The theoretical analysis in \S\ref{sec:method-theory} assumes $\hat{y}$ is fixed; if recomputing it preserved or improved the signal, the assumption would be a pure analytical convenience. We ablate per-batch vs.\ per-sample pseudolabels and frozen-at-$\theta_0$ vs.\ recomputed-at-step. Freezing at $\theta_0$ is necessary: recomputing destabilizes the target and the anchor-loss signal becomes noisy. The fixed-$\hat{y}$ choice is therefore both an analytical and an operational requirement, not a tractability shortcut.

\begin{figure}[t]
    \centering
    \includegraphics[width=0.7\linewidth]{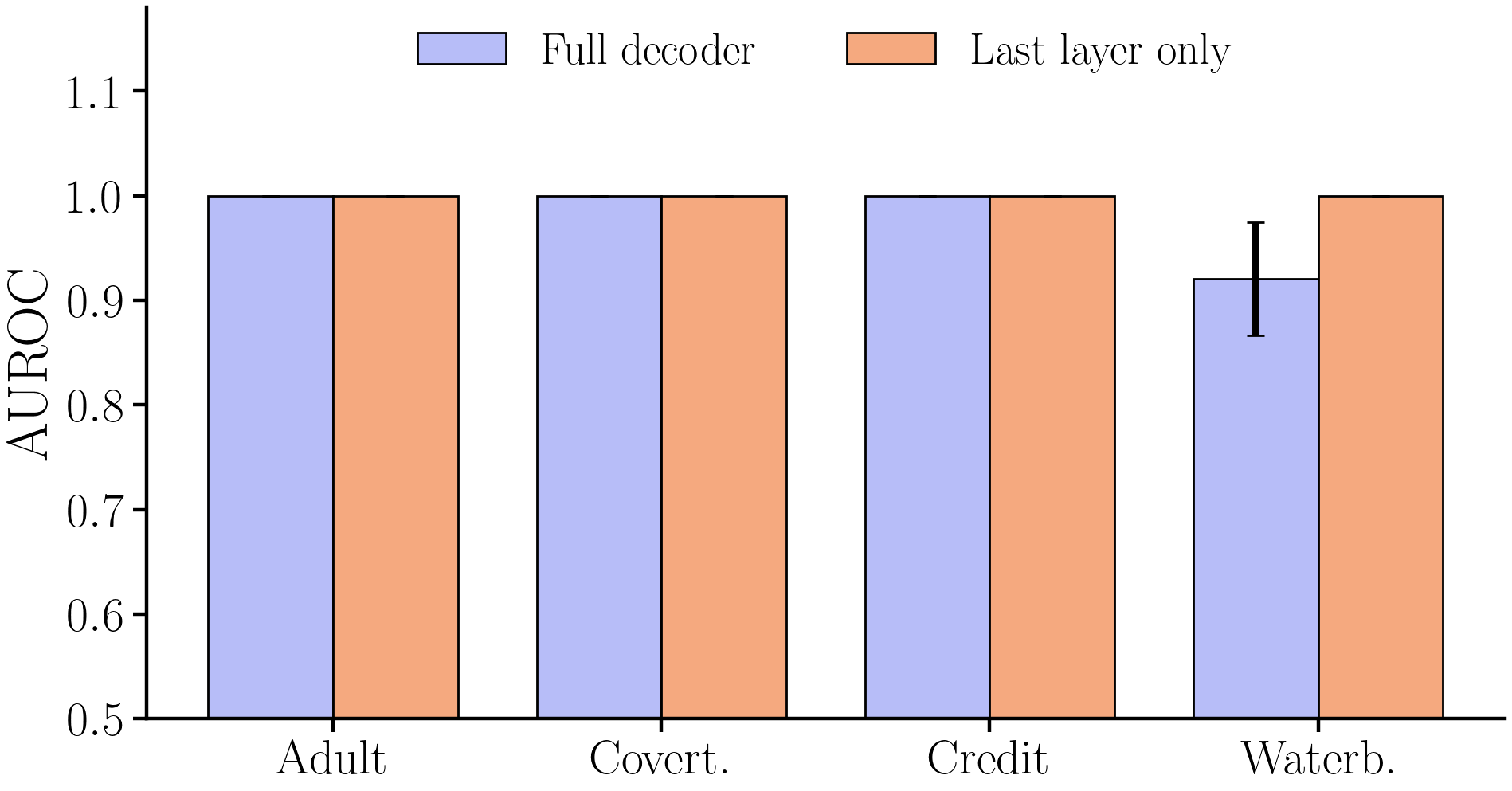}
    \caption{\textbf{Last-layer-only vs.\ full-decoder finetuning.} Detection AUROC matches to within $0.01$ on every dataset, confirming that Theorem~\ref{thm:main-separation-main}'s last-layer assumption is benign in practice.}
    \label{fig:ablation_lastlayer}
\end{figure}

\paragraph{Last-layer-only finetuning.}
Theorem~\ref{thm:main-separation-main} proves a clean rank-one form for the per-sample gradients $g_a$ and $g_e$ only when updates are restricted to the final linear layer: with the full two-layer GELU decoder unfrozen, the gradient mixes contributions across layers and the feature-orthogonality argument in Corollary~\ref{cor:feature-orth} no longer applies. The deployed detector finetunes the whole decoder, so the theory could in principle be analyzing a regime that has nothing to do with how \cald is actually run. Re-running with only the last linear layer trainable matches full-decoder AUROC to within $0.01$ on every dataset (Figure~\ref{fig:ablation_lastlayer}). The two regimes are therefore empirically indistinguishable, which means the analytical restriction is not a tractability tax: the theory describes the regime under which the detector operates, and a practitioner who wants the cheaper last-layer-only protocol pays nothing in detection power.

\subsection{Disagreement Weight $\beta$}
\label{sec:abl-beta}

The joint objective weights the disagreement term by a coefficient $\beta$ that the main experiments fix at $1$, and a natural concern is that this choice is key: if detection degraded away from $\beta = 1$, the reported results would hinge on an untuned constant, and if detection survived $\beta = 0$, the disagreement term would not be the mechanism at all. We are therefore interested in ablating $\beta$ across several orders of magnitude, including the degenerate $\beta = 0$. At first order the question is already constrained: the separation gap obeys $\Delta\lambda \approx \eta\beta\,(\bbE_{\mathrm{ID}}\langle g_a, g_e\rangle - \bbE_{\mathrm{OOD}}\langle g_a, g_e\rangle)$, so $\eta$ and $\beta$ enter only through their product and the learning-rate sweep of Appendix~\ref{sec:appendix-hyperparam} already spans this axis at $\beta = 1$; the sweep below tests whether the first-order picture survives beyond it.

\begin{figure}[t]
\centering
\includegraphics[width=0.8\linewidth]{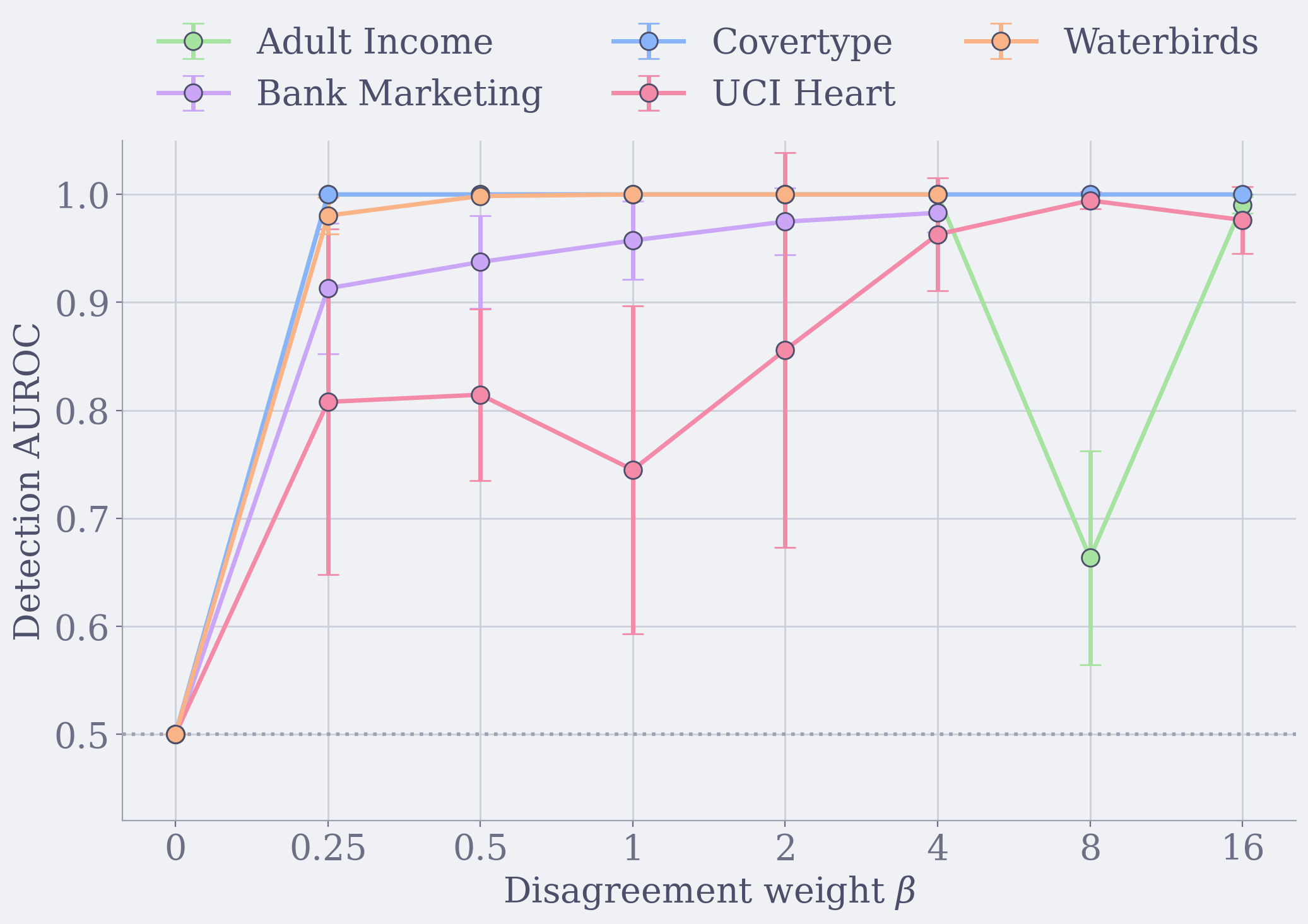}
\caption{\textbf{Ablation of the disagreement weight $\beta$.} Mean detection AUROC over 3 seeds (error bars: $\pm 1$ std; TabPFN, fixed $\eta = 10^{-3}$, 30 rounds; dotted line: chance). At $\beta = 0$ every dataset collapses exactly to chance; detection is flat across $\beta \in [0.25, 4]$ and the first-order picture degrades only at very large $\eta\beta$.}
\label{fig:abl-beta}
\end{figure}

We set up the experiment as follows: TabPFN with fixed $\eta = 10^{-3}$, 30 calibration rounds, and 3 seeds, sweeping $\beta \in \{0, 0.25, 0.5, 1, 2, 4\}$ on five datasets spanning far-OOD, moderate tabular, and subpopulation shift, extended to $\beta \in \{8, 16\}$ on three of them. Figure~\ref{fig:abl-beta} plots mean detection AUROC with per-seed spread; Table~\ref{tab:abl-beta} reports the values.

\begin{table}[t]
\centering
\small
\caption{\textbf{Ablation of the disagreement weight $\beta$.} Mean detection AUROC (TabPFN, fixed $\eta = 10^{-3}$, 30 rounds, 3 seeds). Within-configuration results under a reduced protocol; absolute values are not comparable to Table~\ref{tab:detection}, which selects the learning rate per cell. ``--'': not run.}
\label{tab:abl-beta}
\begin{tabular}{lcccccccc}
\toprule
Dataset & $\beta{=}0$ & $0.25$ & $0.5$ & $1$ & $2$ & $4$ & $8$ & $16$ \\
\midrule
Adult Income    & 0.500 & 1.000 & 1.000 & 1.000 & 1.000 & 1.000 & 0.663 & 0.990 \\
Bank Marketing  & 0.500 & 0.913 & 0.937 & 0.957 & 0.975 & 0.983 & -- & -- \\
Covertype       & 0.500 & 1.000 & 1.000 & 1.000 & 1.000 & 1.000 & 1.000 & 1.000 \\
UCI Heart       & 0.500 & 0.808 & 0.814 & 0.745 & 0.856 & 0.963 & 0.994 & 0.976 \\
Waterbirds      & 0.500 & 0.980 & 0.999 & 1.000 & 1.000 & 1.000 & -- & -- \\
\bottomrule
\end{tabular}
\end{table}

These results show two things. First, at $\beta = 0$ the AUROC collapses to exactly $0.500$ on every run of every dataset: with the disagreement term removed, the post-finetune anchor loss carries no ID/OOD information whatsoever, confirming that the eval-batch disagreement gradient is the mechanism rather than an auxiliary regularizer. Second, detection is flat across $\beta \in [0.25, 4]$, a sixteen-fold range around the default, so $\beta = 1$ sits on a wide plateau rather than a tuned peak. At very large $\beta$ the product $\eta\beta$ exits the small-step regime of Assumption~\ref{assn:small-step} and the first-order picture no longer applies: Adult Income becomes unstable at $\beta = 8$ before recovering at $16$, while UCI Heart, the dataset with the weakest first-order signal (cf.\ Table~\ref{tab:abl-traj}), actually benefits from the stronger push. $\beta = 1$ is principled at first order, where only the product $\eta\beta$ matters, and empirically non-critical over two orders of magnitude.

\subsection{Deployment Cost and the Finetuning Budget $T$}
\label{sec:abl-budget}

Unlike post-hoc scores, \cald must run $T$ gradient steps on every monitored batch, so its deployment cost is a practical concern: a detector that is accurate but orders of magnitude more expensive than the alternatives needs to state exactly what the budget buys. We seek to determine (i) the wall-clock cost per monitored batch and (ii) how detection accuracy depends on the finetuning budget $T$, i.e.\ how much of the paper's $T = 50$ budget is actually needed.

We set up the experiment as follows. We time one finetuning step per monitored batch of 128 points on an NVIDIA H100: 0.168\,s for TabPFN and 0.056\,s for TabICL, so the full $T = 50$ budget costs 8.4\,s and 2.8\,s per batch respectively, against 0.0003--0.67\,s \emph{total} for the 15 post-hoc baselines. We then truncate the detector at $T \in \{5, 10, 20, 50\}$ on all 15 datasets and both models (10 seeds, per-cell selected learning rate as in Table~\ref{tab:detection}) and record $T_{95}$, the smallest $T$ whose mean AUROC reaches 95\% of the $T = 50$ value. Table~\ref{tab:abl-budget} reports the full per-dataset cost--accuracy profile.

\begin{table}[t]
\centering
\setlength{\tabcolsep}{4pt}
\small
\caption{\textbf{Detection AUROC as a function of the finetuning budget $T$.} Mean over 10 seeds at the per-cell selected learning rate (same protocol as Table~\ref{tab:detection}). $T_{95}$ is the smallest $T$ reaching 95\% of the $T{=}50$ AUROC; s@$T_{95}$ is the corresponding wall-clock per monitored batch (H100, batch 128). Aggregates: TabPFN reaches $0.845 \pm 0.159$ at $T{=}10$ vs.\ $0.904 \pm 0.154$ at $T{=}50$ with median $T_{95} = 1$; TabICL reaches $0.851 \pm 0.161$ vs.\ $0.930 \pm 0.124$ with median $T_{95} = 2$.}
\label{tab:abl-budget}
\begin{tabular}{llcccccc}
\toprule
Dataset & Model & @$T{=}5$ & @$10$ & @$20$ & @$50$ & $T_{95}$ & s@$T_{95}$ \\
\midrule
Adult Income & TabPFN & 1.000 & 1.000 & 1.000 & 1.000 & 1 & 0.17 \\
 & TabICL & 0.978 & 1.000 & 1.000 & 1.000 & 2 & 0.11 \\
Amazon & TabPFN & 0.538 & 0.544 & 0.543 & 0.550 & 1 & 0.17 \\
 & TabICL & 0.592 & 0.612 & 0.624 & 0.600 & 1 & 0.06 \\
Bank Marketing & TabPFN & 0.890 & 0.981 & 0.985 & 0.997 & 2 & 0.34 \\
 & TabICL & 1.000 & 0.957 & 1.000 & 1.000 & 1 & 0.06 \\
Camelyon17 (DINOv2) & TabPFN & 0.992 & 0.639 & 0.678 & 0.675 & 1 & 0.17 \\
 & TabICL & 0.578 & 0.999 & 1.000 & 1.000 & 2 & 0.11 \\
Camelyon17 (ResNet18) & TabPFN & 0.994 & 0.811 & 0.882 & 0.977 & 3 & 0.50 \\
 & TabICL & 0.608 & 0.977 & 0.988 & 0.990 & 2 & 0.11 \\
CelebA & TabPFN & 0.819 & 0.734 & 0.778 & 0.919 & 1 & 0.17 \\
 & TabICL & 0.711 & 0.718 & 0.698 & 0.845 & 36 & 2.02 \\
CIFAR-10$\to$10.1 & TabPFN & 0.912 & 0.984 & 0.999 & 1.000 & 7 & 1.17 \\
 & TabICL & 0.758 & 0.964 & 0.996 & 0.996 & 10 & 0.56 \\
CivilComments & TabPFN & 0.998 & 0.821 & 0.913 & 0.999 & 1 & 0.17 \\
 & TabICL & 0.994 & 0.832 & 0.976 & 0.997 & 1 & 0.06 \\
Covertype & TabPFN & 1.000 & 1.000 & 1.000 & 1.000 & 1 & 0.17 \\
 & TabICL & 1.000 & 1.000 & 1.000 & 1.000 & 1 & 0.06 \\
Credit Default & TabPFN & 1.000 & 0.960 & 1.000 & 1.000 & 1 & 0.17 \\
 & TabICL & 0.771 & 0.611 & 0.615 & 0.698 & 1 & 0.06 \\
Diabetes Readmission & TabPFN & 0.993 & 0.956 & 0.962 & 0.976 & 1 & 0.17 \\
 & TabICL & 0.835 & 0.913 & 0.997 & 1.000 & 14 & 0.78 \\
Folktables & TabPFN & 0.573 & 0.574 & 0.577 & 0.624 & 38 & 6.37 \\
 & TabICL & 0.771 & 0.588 & 0.752 & 0.899 & 9 & 0.50 \\
PACS & TabPFN & 0.943 & 0.959 & 1.000 & 1.000 & 10 & 1.68 \\
 & TabICL & 0.713 & 1.000 & 1.000 & 1.000 & 8 & 0.45 \\
UCI Heart & TabPFN & 0.859 & 0.792 & 0.727 & 0.915 & 7 & 1.17 \\
 & TabICL & 0.715 & 0.693 & 0.777 & 0.930 & 21 & 1.18 \\
Waterbirds & TabPFN & 0.726 & 0.919 & 0.920 & 0.921 & 1 & 0.17 \\
 & TabICL & 0.844 & 0.896 & 0.918 & 0.993 & 3 & 0.17 \\
\bottomrule
\end{tabular}
\end{table}

The timing results show a trade-off. In raw wall-clock terms \cald is one to four orders of magnitude more expensive than any post-hoc baseline, and no amount of truncation closes that gap. But the budget needed to reach the reported accuracy is a small fraction of $T = 50$: the median $T_{95}$ is 1 step for TabPFN and 2 for TabICL, and $T \le 10$ (about 1.7\,s and 0.6\,s per monitored batch) recovers 95\% of the final AUROC on 14 of 15 datasets for TabPFN and 12 of 15 for TabICL. The exceptions are the slow-converging cells (Folktables/TabPFN at $T_{95} = 38$; CelebA, UCI Heart, and Diabetes Readmission under TabICL at 36, 21, and 14), which are also among the weaker cells at full budget. One cell is non-monotone: Camelyon17 (DINOv2) under TabPFN peaks early (0.992 at $T{=}5$) and degrades with further finetuning, which is the budget-resolved view of its weak full-budget value in Table~\ref{tab:detection}. $T = 10$ is the practical operating point for deployment, with $T = 50$ retained in the main results as the conservative budget at which the theory's small-step premises were validated.

\subsection{Frozen Deployment Configuration and Stability of the Learning-Rate Selection}
\label{sec:abl-frozen}

Proposition~\ref{prop:conformal-main} bounds the false-alarm rate only when the anchor, support, and finetuning hyperparameters are fixed \emph{before} deployment, yet Table~\ref{tab:detection} selects the learning rate per dataset--model cell with the tail-stability rule, i.e.\ with a procedure that looks at the trajectories being certified. The guarantee therefore covers a fixed configuration, not the selection protocol, and two questions follow: what does it cost to actually deploy the fixed configuration the guarantee speaks about, and how fragile is the selection rule that produces the headline numbers? We measure three things: the accuracy cost of freezing $\eta = 10^{-3}$ everywhere, a census of the selection rule's fallback behavior across all cells and tail windows, and the empirical false-positive rate of the frozen configuration under stress.

We set up the experiment as follows. (a) We recompute all 30 method$\times$dataset cells with the single frozen configuration $\eta = 10^{-3}$ (no selection of any kind) and compare against the selected-LR pipeline of Table~\ref{tab:detection}. (b) We re-run the tail-stability selection over all 30 cells and three tail windows (steps 25--45, 30--50, 35--50). (c) We estimate the false-positive rate of the frozen configuration under the no-shift null over 240 calibration/deployment rounds on six datasets, in four stress regimes: deterministic replay, nondeterministic GPU kernels, re-drawing the anchor every round, and time-correlated eval streams.

\begin{table}[t]
\centering
\setlength{\tabcolsep}{4pt}
\small
\caption{\textbf{Frozen configuration ($\eta = 10^{-3}$, no selection) vs.\ the per-cell selected learning rate of Table~\ref{tab:detection}.} Detection AUROC; $\Delta$ = frozen $-$ selected. Mean $\Delta$ over all 30 cells is $-0.005$ (AUROC) and $+0.008$ (TPR@5\%, not shown). The frozen column is the configuration covered by the conformal guarantee.}
\label{tab:abl-frozen}
\begin{tabular}{lcccccc}
\toprule
 & \multicolumn{3}{c}{TabPFN} & \multicolumn{3}{c}{TabICL} \\
\cmidrule(lr){2-4}\cmidrule(lr){5-7}
Dataset & sel. & frozen & $\Delta$ & sel. & frozen & $\Delta$ \\
\midrule
Adult Income & 1.000 & 1.000 & $+$0.000 & 1.000 & 1.000 & $+$0.000 \\
Amazon & 0.550 & 0.606 & $+$0.056 & 0.600 & 0.624 & $+$0.024 \\
Bank Marketing & 0.997 & 0.935 & $-$0.062 & 1.000 & 1.000 & $+$0.000 \\
Camelyon17 (DINOv2) & 0.675 & 0.999 & $+$0.324 & 1.000 & 1.000 & $+$0.000 \\
Camelyon17 (ResNet18) & 0.977 & 0.998 & $+$0.021 & 0.990 & 0.998 & $+$0.008 \\
CelebA & 0.919 & 0.919 & $+$0.000 & 0.845 & 0.806 & $-$0.039 \\
CIFAR-10$\to$10.1 & 1.000 & 0.896 & $-$0.103 & 0.996 & 0.769 & $-$0.227 \\
CivilComments & 0.999 & 0.990 & $-$0.009 & 0.997 & 0.993 & $-$0.005 \\
Covertype & 1.000 & 1.000 & $+$0.000 & 1.000 & 1.000 & $+$0.000 \\
Credit Default & 1.000 & 1.000 & $+$0.000 & 0.698 & 0.766 & $+$0.068 \\
Diabetes Readmission & 0.976 & 0.984 & $+$0.008 & 1.000 & 0.895 & $-$0.105 \\
Folktables & 0.624 & 0.572 & $-$0.052 & 0.899 & 0.859 & $-$0.040 \\
PACS & 1.000 & 0.940 & $-$0.060 & 1.000 & 0.992 & $-$0.008 \\
UCI Heart & 0.908 & 0.885 & $-$0.024 & 0.930 & 0.903 & $-$0.027 \\
Waterbirds & 0.921 & 1.000 & $+$0.079 & 0.993 & 1.000 & $+$0.007 \\
\bottomrule
\end{tabular}
\end{table}

\begin{table}[t]
\centering
\small
\caption{\textbf{Empirical false-positive rate of the frozen configuration under the no-shift null.} Six datasets $\times$ 240 null rounds per regime; conformal bound $\alpha + 1/(K{+}1) = 0.070$. Regime means sit at or below the bound; four individual dataset--regime cells marginally exceed it (max 0.075), consistent with finite-sample fluctuation at $n = 240$.}
\label{tab:abl-fpr}
\begin{tabular}{lcc}
\toprule
Stress regime & Mean FPR & Per-dataset range \\
\midrule
Deterministic replay & 0.065 & 0.057--0.072 \\
Nondeterministic GPU kernels & 0.065 & 0.057--0.072 \\
Anchor re-drawn every round & 0.064 & 0.052--0.069 \\
Time-correlated eval stream & 0.068 & 0.063--0.075 \\
\bottomrule
\end{tabular}
\end{table}

The certified configuration is deployable at essentially no average cost: freezing $\eta = 10^{-3}$ changes AUROC by $-0.005$ on average over the 30 cells. The average hides real per-cell movement in both directions, which we disclose rather than average away: CIFAR-10 is the worst case ($-0.103$ TabPFN, $-0.227$ TabICL; it needs $\eta = 10^{-2}$), while Camelyon17 (DINOv2)/TabPFN \emph{improves} by $+0.324$, meaning the selection rule itself mis-serves that cell. The census in (b) shows the selection rule is otherwise stable: 4 of 30 cells trigger the fallback to $\eta = 10^{-4}$ (Adult Income on both models at AUROC 1.000 despite the fallback, CelebA/TabICL at 0.845, and Folktables/TabPFN at 0.624), all four on benign shifts, while every severely shifted dataset selects its learning rate stably; across the three tail windows the selected rate changes in 1 of 60 comparisons (Folktables/TabPFN under the 25--45 window, $\Delta$AUROC $-0.052$). Finally, Table~\ref{tab:abl-fpr} shows the frozen configuration's false-alarm rate stays controlled under all four stress regimes. The fixed $\eta = 10^{-3}$ configuration is the certified deployment mode of \cald{} --- the mode Proposition~\ref{prop:conformal-main} actually speaks about --- and per-cell selection is a reporting-time diagnostic that requires labeled OOD trajectories and should be read as such in Table~\ref{tab:detection}.

\subsection{Anchor Construction Robustness}
\label{sec:abl-anchor}

The anchor set is the reference that every statistic in the method is scored against, so the entire pipeline could in principle hinge on drawing a lucky, perfectly clean anchor --- a serious concern in deployment, where held-out labels may be scarce, noisy, or quietly contaminated by the very shift being monitored. Appendix~\ref{sec:appendix-hyperparam} already varies the anchor's \emph{size}; here we seek to determine the sensitivity to its \emph{quality}, stressing the anchor along the remaining axes: label noise, contamination by OOD points, and re-drawing the anchor itself.

We set up the experiment as follows: TabPFN, 3 seeds per cell. For label noise we flip a fraction $\rho \in \{0, 0.05, 0.1, 0.2\}$ of anchor labels uniformly at random; for contamination we replace a fraction $\rho \in \{0, 0.05, 0.1, 0.25\}$ of anchor points with OOD samples; for re-drawing we resample the anchor from the held-out ID pool at every calibration round and measure the false-positive rate under the no-shift null (the ``anchor re-drawn every round'' regime of Table~\ref{tab:abl-fpr}). Table~\ref{tab:abl-anchor} reports detection AUROC on the datasets with complete stress grids.

\begin{table}[t]
\centering
\setlength{\tabcolsep}{4pt}
\small
\caption{\textbf{Anchor-quality stress.} Detection AUROC (TabPFN, 3 seeds) under anchor label noise and anchor OOD contamination. Within-configuration results under a reduced protocol; absolute values are not comparable to Table~\ref{tab:detection}. ``--'': stress level not completed. All movements are within the per-seed standard deviation of the corresponding cell (up to $\pm 0.18$ on PACS and $\pm 0.16$ on CIFAR-10).}
\label{tab:abl-anchor}
\begin{tabular}{lcccccccc}
\toprule
 & \multicolumn{4}{c}{Label noise $\rho$} & \multicolumn{4}{c}{OOD contamination $\rho$} \\
\cmidrule(lr){2-5}\cmidrule(lr){6-9}
Dataset & $0$ & $.05$ & $.10$ & $.20$ & $0$ & $.05$ & $.10$ & $.25$ \\
\midrule
Adult Income & 1.000 & 1.000 & 1.000 & 1.000 & 1.000 & 1.000 & 1.000 & 1.000 \\
Bank Marketing & 0.934 & 0.935 & 0.933 & 0.946 & 0.932 & 0.932 & 0.929 & 0.925 \\
CIFAR-10$\to$10.1 & 0.920 & 0.757 & 0.835 & 0.955 & 0.963 & -- & -- & -- \\
CivilComments & 0.991 & 0.989 & 0.988 & 1.000 & 0.991 & 0.991 & 0.988 & -- \\
Covertype & 1.000 & 1.000 & 1.000 & 1.000 & 1.000 & 1.000 & 1.000 & 1.000 \\
Credit Default & 1.000 & 1.000 & 1.000 & 1.000 & 1.000 & 1.000 & 1.000 & 1.000 \\
Diabetes Readmission & 1.000 & 1.000 & 1.000 & 1.000 & 1.000 & 1.000 & 1.000 & 1.000 \\
Folktables & 0.516 & 0.518 & 0.526 & 0.526 & 0.515 & 0.511 & 0.517 & 0.529 \\
PACS & 0.906 & 0.790 & 0.903 & 0.852 & 0.932 & -- & -- & -- \\
UCI Heart & 0.751 & 0.704 & 0.691 & 0.839 & 0.751 & 0.723 & 0.776 & 0.735 \\
Waterbirds & 1.000 & 0.997 & 0.989 & -- & 1.000 & 1.000 & 1.000 & 0.999 \\
\bottomrule
\end{tabular}
\end{table}

Anchor quality is a low-sensitivity component of the method. Under up to 20\% label noise, no dataset shows a movement exceeding its own per-seed standard deviation: the largest swings (PACS, about $\pm 0.1$ across noise levels against a per-seed std of up to $\pm 0.18$, and CIFAR-10's dip at $\rho = 0.05$ against a std of $\pm 0.16$) are seed noise, not trends, and neither is monotone in $\rho$ as genuine degradation would be. OOD contamination up to 25\% moves no dataset by more than $0.03$. Re-drawing the anchor every round keeps the false-positive rate at 0.052--0.069, within the 0.070 bound. This robustness is also what the theory predicts: in Theorem~\ref{thm:main-separation-main}, $\gamma$ is a population-level constant in which the anchor sample never appears (conditioning the proof on a fixed anchor merely replaces $\gamma$ by $\gamma(\Xanc)$), and in Proposition~\ref{prop:conformal-main} the exchangeable unit of the rank argument is the calibration \emph{round}, not the anchor point, so anchor size and composition never enter either guarantee. Empirically the anchor contributes little even to the \emph{variance} of the statistic: the spread of the per-round calibration scores is essentially flat from anchor size 16 to 256 (median std ratio 0.97), because round-to-round eval-batch resampling, not anchor averaging, dominates the variance of $\lambda$. The anchor needs to be a modest, roughly clean sample of ID data and nothing more.

\subsection{Feature Collapse in Deployed Multi-Layer PFNs}
\label{sec:abl-collapse}

The feature-orthogonality assumption behind Theorem~\ref{thm:main-separation-main} is motivated by Nagler's tilting argument (Appendix~\ref{sec:theory-nagler}), which is proven for single-layer attention, while the deployed TabPFN and TabICL are multi-layer transformers --- so the assumption could in principle describe a model class that the experiments never touch. The multi-layer extension is a hard open problem, but the argument's central \emph{prediction} is directly measurable: as an evaluation batch becomes more OOD, its features should collapse toward a common representation, becoming more orthogonal to the ID anchor features. We seek to determine whether this predicted collapse actually occurs in the deployed multi-layer models, under both far-OOD and near-OOD conditions.

\begin{figure}[t]
\centering
\includegraphics[width=\linewidth]{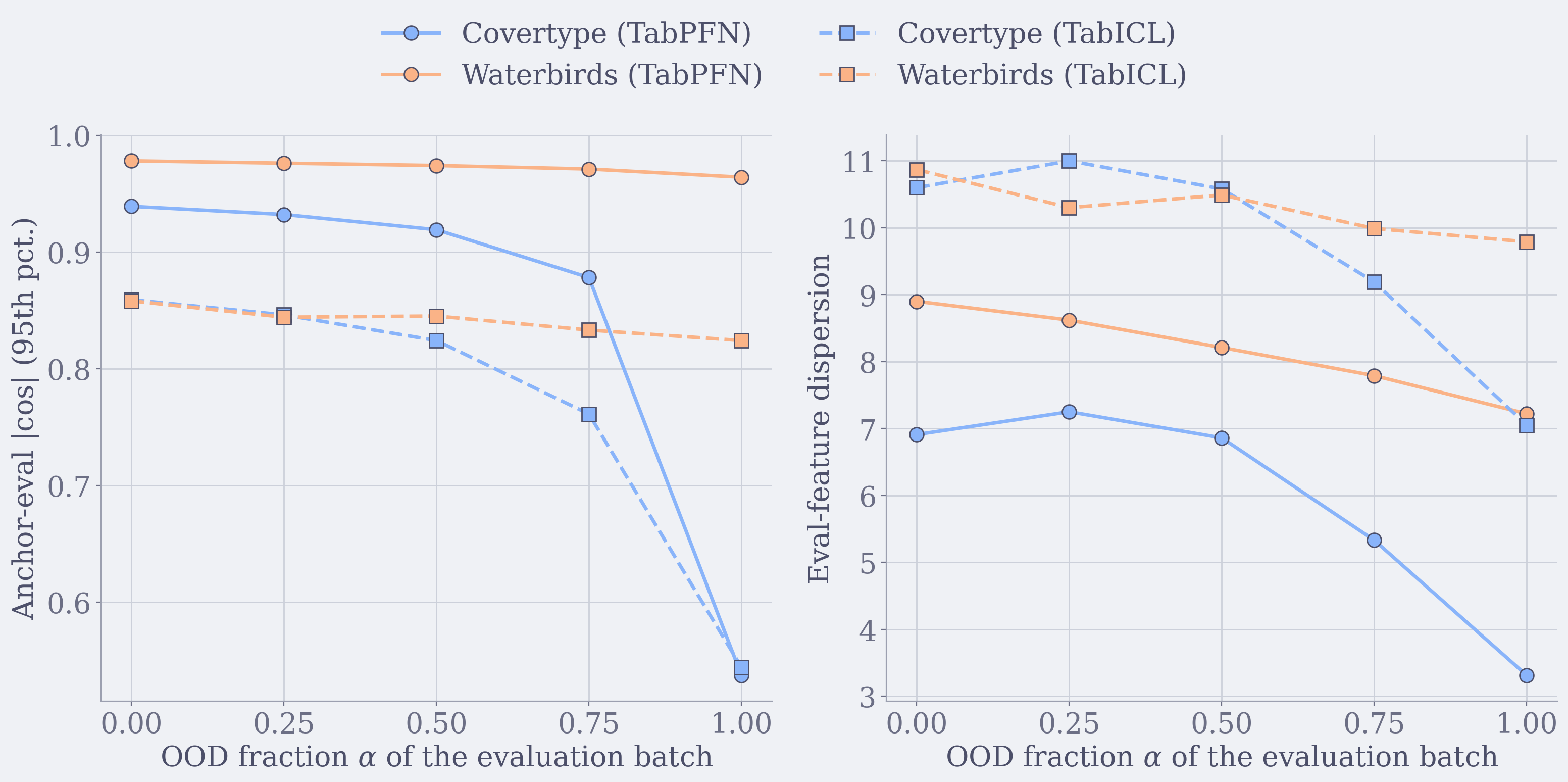}
\caption{\textbf{Feature collapse under controlled OOD mixing.} 95th-percentile absolute anchor--eval cosine (left; the empirical proxy for $\epsilon$) and evaluation-feature dispersion (right) as the OOD fraction $\alpha$ of the evaluation batch grows; solid lines are TabPFN, dashed are TabICL. Both collapse statistics fall sharply on the far-OOD Covertype and mildly on the near-OOD Waterbirds, as the tilting argument predicts.}
\label{fig:abl-collapse}
\end{figure}

We set up the experiment as follows. On Covertype (far-OOD) and Waterbirds (near-OOD) we construct evaluation batches containing a fraction $\alpha \in \{0, 0.25, 0.5, 0.75, 1\}$ of OOD examples and $1 - \alpha$ ID examples. For each $\alpha$ we extract the anchor and evaluation features of the deployed models with no finetuning (TabPFN: the pre-decoder representation; TabICL: the input to the ICL predictor's decoder, i.e.\ the post-LayerNorm output of the dataset-wise ICL transformer) and measure two statistics: the 95th percentile of the absolute cosine similarity across all anchor--evaluation feature pairs, after centering by the pooled $\alpha = 1$ evaluation mean as a proxy for the collapse direction $\phi_\infty$ (this is the empirical proxy for $\epsilon$ in Assumption~\ref{assn:feat-orth}, and should \emph{decrease} under tilting), and the evaluation-feature dispersion, the mean Euclidean distance of each evaluation feature to the mean evaluation feature (which should also \emph{decrease} under collapse). We additionally report detection AUROC of the mixed batch against the $\alpha = 0$ reference. Figure~\ref{fig:abl-collapse} plots the two collapse statistics; Table~\ref{tab:abl-collapse} reports all values.

\begin{table}[t]
\centering
\setlength{\tabcolsep}{3.5pt}
\small
\caption{\textbf{Feature collapse under controlled OOD mixing.} 95th-percentile absolute anchor--eval cosine (the empirical proxy for $\epsilon$), evaluation-feature dispersion, and detection AUROC vs.\ the $\alpha{=}0$ reference, as the OOD fraction $\alpha$ of the evaluation batch grows. Both statistics decrease with $\alpha$, strongly on the far-OOD Covertype and mildly on the near-OOD Waterbirds. AUROC here is within the mixing protocol and not comparable to Table~\ref{tab:detection}.}
\label{tab:abl-collapse}
\begin{tabular}{llccccc@{\hskip 10pt}ccccc}
\toprule
 & & \multicolumn{5}{c}{Covertype (far-OOD), $\alpha =$} & \multicolumn{5}{c}{Waterbirds (near-OOD), $\alpha =$} \\
\cmidrule(lr){3-7}\cmidrule(lr){8-12}
Model & Metric & $0$ & $.25$ & $.5$ & $.75$ & $1$ & $0$ & $.25$ & $.5$ & $.75$ & $1$ \\
\midrule
TabPFN & 95\% $\vert\cos\vert$ & 0.939 & 0.932 & 0.919 & 0.878 & 0.537 & 0.978 & 0.976 & 0.974 & 0.971 & 0.964 \\
 & dispersion & 6.91 & 7.25 & 6.86 & 5.33 & 3.31 & 8.90 & 8.62 & 8.21 & 7.79 & 7.22 \\
 & AUROC & -- & 0.750 & 0.883 & 1.000 & 1.000 & -- & 0.709 & 0.936 & 0.973 & 1.000 \\
\midrule
TabICL & 95\% $\vert\cos\vert$ & 0.859 & 0.846 & 0.824 & 0.761 & 0.544 & 0.858 & 0.844 & 0.845 & 0.833 & 0.824 \\
 & dispersion & 10.60 & 11.00 & 10.58 & 9.19 & 7.05 & 10.87 & 10.30 & 10.49 & 9.99 & 9.79 \\
 & AUROC & -- & 1.000 & 1.000 & 1.000 & 1.000 & -- & 0.772 & 0.932 & 1.000 & 1.000 \\
\bottomrule
\end{tabular}
\end{table}

The tilting prediction holds in the deployed multi-layer models, with exactly the severity profile the theory assigns it. On Covertype, both collapse statistics fall sharply and monotonically with the OOD fraction: the cosine proxy drops from 0.939 to 0.537 (TabPFN) and from 0.859 to 0.544 (TabICL), and dispersion contracts by factors of 2.1 and 1.5. On Waterbirds the same decreases appear but are mild (cosine ratios 0.99 and 0.96), consistent with Theorem~\ref{thm:main-separation-main}'s scope: the collapse is a far-OOD limit phenomenon, and a near-OOD subpopulation shift should, and does, produce only a weak echo of it. Detection AUROC rises with $\alpha$ in step with the collapse statistics in both models. Although the tilting argument is unproven for general transformer stacks, its measurable prediction is what the deployed multi-layer models actually do, precisely where Theorem~\ref{thm:main-separation-main} invokes it; the multi-layer proof itself remains open (Appendix~\ref{sec:theory-nagler}).

\subsection{Theorem~\ref{thm:main-separation-main}'s Condition and Premise at Finite Severity}
\label{sec:abl-certificate}

Section~\ref{sec:abl-cross-term} of the main paper gives an abridged version of this analysis (Figure~\ref{fig:premise_delta_traj}); the full estimates and the per-dataset $\lambda$-gap are collected here.

Theorem~\ref{thm:main-separation-main} certifies the signed separation only when $\epsilon < \gamma / (B_\phi^2 B_C)$, a condition stated in worst-case constants, and the question is whether it is ever satisfied on the benchmark at finite shift severity --- or whether the theory certifies a regime the experiments never visit. We seek to determine, per dataset, (i) whether the theorem's condition holds as stated, and (ii) when it does not, whether the gradient mechanism the proof is built on is nonetheless present, by measuring the proof's premise directly instead of routing it through the worst-case constants.

We set up the experiment as follows. On ten datasets (TabPFN) we estimate every quantity in the condition from the factorization of Lemma~\ref{lem:factorization}: $\hat\epsilon$ is the 95th percentile of the absolute anchor--eval feature cosines (centered by the pooled-OOD-mean proxy for $\phi_\infty$; the 95th percentile is a \emph{lenient} estimate, since the assumption formally requires a uniform bound), and $\hat\gamma$, $\hat B_\phi$, $\hat B_C$ are the empirical versions of the ID-interference and boundedness constants. Separately, we measure the proof's premise on the finetuning trajectories themselves: $\Delta_{\mathrm{traj}} := \bbE_{\mathrm{OOD}}[\langle \bar g_a, \bar g_e\rangle] - \bbE_{\mathrm{ID}}[\langle \bar g_a, \bar g_e\rangle]$, the gap in the time-averaged anchor--eval gradient alignment of Proposition~\ref{prop:T-step} computed by exact full-decoder autograd, together with the $\lambda$-gap $\bbE[\lambda \mid \mathrm{ID}] - \bbE[\lambda \mid \mathrm{OOD}]$, the raw separation available to the detector (here and below we write $\bbE[\lambda \mid \mathrm{ID}]$ for the main paper's $\bbE[\lambda_{\mathrm{ID}}]$, and likewise for OOD). The proof's mechanism operates iff $\Delta_{\mathrm{traj}} > 0$: the disagreement gradient interferes with anchor descent more strongly under ID than under OOD inputs.

\begin{table}[t]
\centering
\small
\caption{\textbf{Theorem~\ref{thm:main-separation-main}'s condition at finite severity.} Empirical estimates of both sides of $\epsilon < \gamma/(B_\phi^2 B_C)$ (TabPFN). The condition fails on every dataset: $\hat\epsilon$ is $O(1)$ while the right-hand side is near zero or negative. The AUROC column repeats Table~\ref{tab:detection}'s selected-LR values for reference.}
\label{tab:abl-certificate}
\begin{tabular}{lccc}
\toprule
Dataset & $\hat\epsilon$ (q95) & $\hat\gamma/(\hat B_\phi^2 \hat B_C)$ & AUROC (Table~\ref{tab:detection}) \\
\midrule
Covertype & 0.497 & $-0.0033$ & 1.000 \\
CIFAR-10$\to$10.1 & 0.771 & $-0.0010$ & 1.000 \\
Camelyon17 (DINOv2) & 0.920 & $-0.0003$ & 0.675 \\
Camelyon17 (ResNet18) & 0.928 & $+0.0006$ & 0.977 \\
CelebA & 0.951 & $+0.0083$ & 0.919 \\
Credit Default & 0.951 & $-0.0951$ & 1.000 \\
Waterbirds & 0.967 & $-0.0047$ & 0.921 \\
UCI Heart & 0.971 & $-0.1744$ & 0.908 \\
Bank Marketing & 0.973 & $-0.1138$ & 0.997 \\
Diabetes Readmission & 0.991 & $-0.0791$ & 0.976 \\
\bottomrule
\end{tabular}
\end{table}

\begin{table}[t]
\centering
\small
\caption{\textbf{The proof's premise, measured directly on the finetuning trajectories.} $\Delta_{\mathrm{traj}} > 0$ iff the gradient-interference mechanism of Theorem~\ref{thm:main-separation-main} operates; the $\lambda$-gap is the raw ID/OOD separation of the statistic. Both columns are computed under fixed $\eta = 10^{-3}$ (within-configuration); they are not comparable to Table~\ref{tab:detection} or to the selected-LR AUROC column of Table~\ref{tab:abl-certificate}.}
\label{tab:abl-traj}
\begin{tabular}{lcc}
\toprule
Dataset & $\Delta_{\mathrm{traj}}$ & $\lambda$-gap \\
\midrule
Waterbirds & $+87.2$ & $+0.255$ \\
Credit Default & $+40.3$ & $+0.198$ \\
Diabetes Readmission & $+22.6$ & $+0.133$ \\
Bank Marketing & $+10.8$ & $+0.044$ \\
Covertype & $+8.2$ & $+0.158$ \\
Camelyon17 (ResNet18) & $+7.6$ & $+0.228$ \\
CelebA & $+3.0$ & $+0.080$ \\
Camelyon17 (DINOv2) & $+1.5$ & $+0.234$ \\
CIFAR-10$\to$10.1 & $+0.11$ & $+0.006$ \\
UCI Heart & $-4.4$ & $+0.008$ \\
\bottomrule
\end{tabular}
\end{table}

These results show a sharp split between the condition and the mechanism. The condition fails everywhere: $\hat\epsilon \ge 0.497$ on every dataset while the right-hand side never exceeds $+0.0083$ and is negative on eight of ten, so at finite severity every benchmark is near-OOD in the theorem's sense and the formal guarantee certifies none of them (Covertype, the most separated, comes closest). The failure, however, lives in the worst-case constants, not in the mechanism: the premise $\Delta_{\mathrm{traj}} > 0$ holds on nine of ten datasets, including the moderate-overlap benchmarks where the static condition is vacuous. More tellingly, the two datasets where the premise is weakest or fails --- CIFAR-10 ($+0.11$) and UCI Heart ($-4.4$) --- are exactly the two with near-zero $\lambda$-gap ($+0.006$, $+0.008$): the measurement predicts the detector's weakest cells in advance rather than rationalizing its successes after the fact. We conclude that Theorem~\ref{thm:main-separation-main} certifies detection in the far-OOD regime only, that the gradient-interference mechanism it is built on is empirically present across the near-OOD benchmarks, and that a finite-$\epsilon$ bound with measurable constants in place of worst-case ones is the right target for future theory.

\subsection{Two-Signal Harm Detection: Empirical Evaluation}
\label{sec:abl-twosignal}

The deployed statistic $\lambda$ detects distributional difference, not harm: Lemma~\ref{lem:density-ratio} shows the accuracy gap depends only on mass moved onto the model's mistake region, which no label-free statistic can see, and empirically $\lambda$ false-alarms on the benign Bank Marketing, Diabetes Readmission, Credit Default, and CelebA shifts. Appendix~\ref{sec:theory-harm} proposes the confidence-drop statistic $\hat\mu$ (Proposition~\ref{prop:harm-theorem}) as a harm-magnitude estimator, with $\lambda$ as a shift-confirmation gate; here we are interested in whether the composed detector actually restores benign-shift silence, and at what cost in harmful-shift detection. Two properties hold before any measurement: the conjunction fires only when both tests fire, so its false-alarm rate under the null is at most the smaller of the two individual rates, under arbitrary dependence and with no Bonferroni correction (the conformal bound $\alpha + 1/(K{+}1)$ carries over unchanged); and under \emph{any} shifted distribution the $\hat\mu$ gate can only remove alerts produced by the $\lambda$-test, never add them, so it cannot increase benign false alarms.

\begin{figure}[t]
\centering
\includegraphics[width=\linewidth]{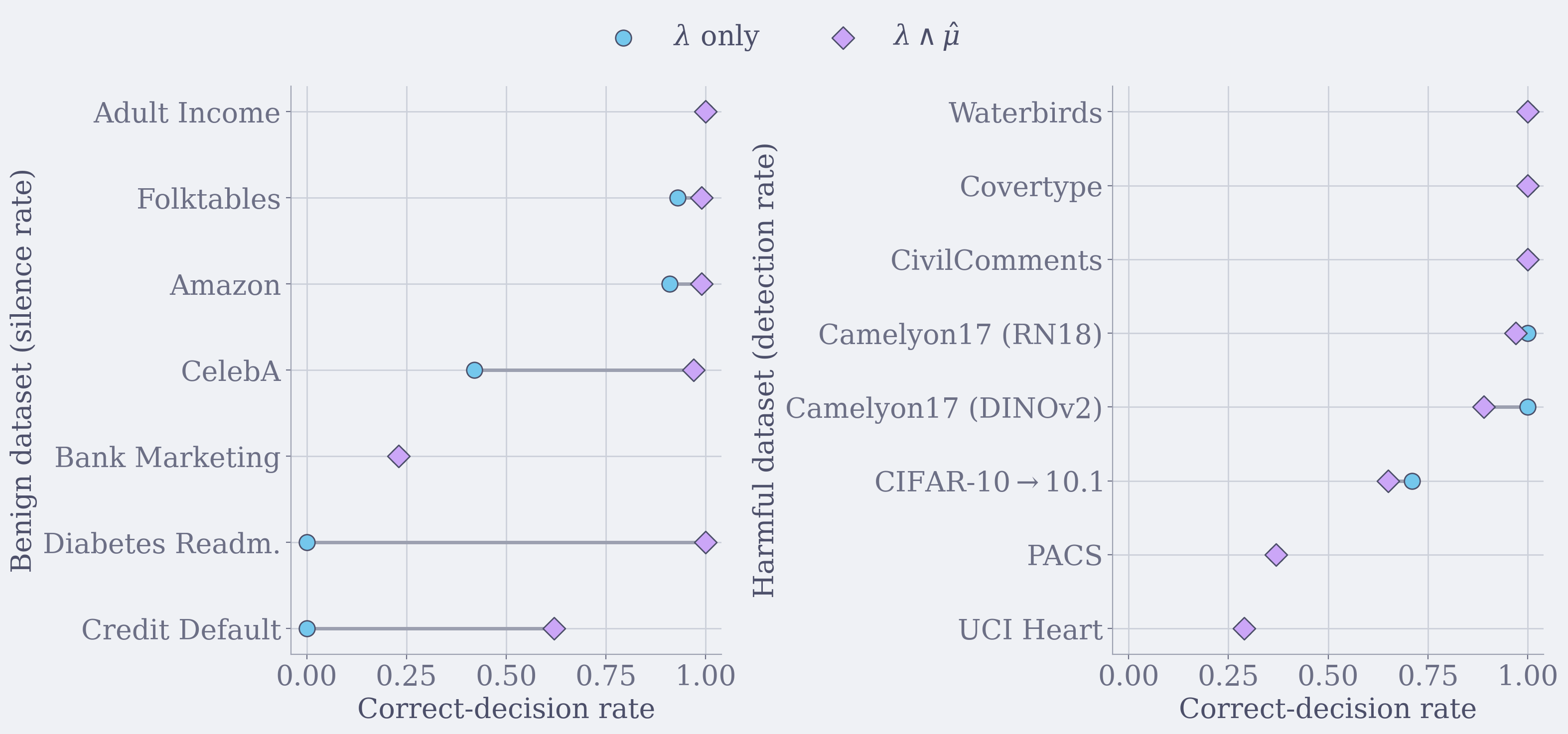}
\caption{\textbf{Two-signal detector vs.\ $\lambda$ alone.} Correct-decision rate per dataset (benign: silence; harmful: detection), TabPFN with fixed $\eta = 10^{-3}$. A diamond to the right of its circle is an improvement from the $\hat\mu$ gate. The gate removes the benign false alarms on CelebA, Credit Default, and Diabetes Readmission, cannot help Bank Marketing (whose confidence collapses), and costs little on harmful shifts.}
\label{fig:abl-twosignal}
\end{figure}

We set up the experiment as follows: all 15 datasets, TabPFN, fixed $\eta = 10^{-3}$, with $\hat\mu = \hat\bbE_{\mathrm{ID}}[p_x(y)] - \hat\bbE_{\mathrm{eval}}[\max_c p_x(c)]$ estimated per Definition~\ref{def:two-signal}. The composed detector fires iff the $\lambda$-test fires and $\hat\mu$ exceeds its threshold. Figure~\ref{fig:abl-twosignal} shows the per-dataset movement from $\lambda$ alone to the conjunction; Table~\ref{tab:abl-twosignal} reports the correct-decision rate of each configuration: the silence rate on benign datasets and the fire rate on harmful ones.

\begin{table}[t]
\centering
\small
\caption{\textbf{Two-signal detector: correct-decision rates.} Benign rows report the silence rate (correct = no alarm); harmful rows report the detection rate (correct = alarm). All 15 datasets, TabPFN, fixed $\eta = 10^{-3}$ (within-configuration; not comparable to Table~\ref{tab:detection}). Means: benign silence $0.50 \rightarrow 0.83$, harmful detection $0.80 \rightarrow 0.77$.}
\label{tab:abl-twosignal}
\begin{tabular}{llccc}
\toprule
Dataset & Shift & $\lambda$ only & $\hat\mu$ only & $\lambda \wedge \hat\mu$ \\
\midrule
Adult Income & benign & 1.00 & 1.00 & 1.00 \\
Amazon & benign & 0.91 & 0.97 & 0.99 \\
Bank Marketing & benign & 0.23 & 0.00 & 0.23 \\
CelebA & benign & 0.42 & 0.97 & 0.97 \\
Credit Default & benign & 0.00 & 0.62 & 0.62 \\
Diabetes Readmission & benign & 0.00 & 1.00 & 1.00 \\
Folktables & benign & 0.93 & 0.75 & 0.99 \\
\midrule
Camelyon17 (DINOv2) & harmful & 1.00 & 0.89 & 0.89 \\
Camelyon17 (ResNet18) & harmful & 1.00 & 0.97 & 0.97 \\
CIFAR-10$\to$10.1 & harmful & 0.71 & 0.90 & 0.65 \\
CivilComments & harmful & 1.00 & 1.00 & 1.00 \\
Covertype & harmful & 1.00 & 1.00 & 1.00 \\
PACS & harmful & 0.37 & 1.00 & 0.37 \\
UCI Heart & harmful & 0.29 & 0.98 & 0.29 \\
Waterbirds & harmful & 1.00 & 1.00 & 1.00 \\
\bottomrule
\end{tabular}
\end{table}

The gate works where the theory says it should. Mean benign silence rises from 0.50 to 0.83: the gate removes the false alarms on three of the four problematic benign datasets (CelebA $0.42 \to 0.97$, Credit Default $0.00 \to 0.62$, Diabetes Readmission $0.00 \to 1.00$). Bank Marketing remains a failure ($0.23 \to 0.23$) because the model's confidence collapses under that benign shift, so both signals fire --- the one benign case where the gate's own premise is violated. On harmful shifts the mean detection rate drops only from 0.80 to 0.77, with the losses concentrated where the $\lambda$-test is already weak under this fixed configuration (PACS, UCI Heart) rather than created by the gate. The construction has a real limitation that bounds how far it should be trusted: $\hat\mu$ reads the model's own confidence, so an overconfident or poorly calibrated model would silence true alarms, and Proposition~\ref{prop:harm-theorem}'s error terms are exactly calibration gaps. A formal treatment of when the confidence signal can be trusted is beyond the scope of this work. The two-signal construction is an evaluated proposal that demonstrably recovers benign-shift silence at modest cost, while $\lambda$ remains the primary, conformally certified statistic of \cald.

\subsection{Detectability Across Shift Types}
\label{sec:abl-shift-type}

\cald{} monitors unlabeled covariates, so its reach is bounded by a structural fact: a shift is visible only insofar as it changes the covariate marginal $p(X)$. The PFN's pretraining prior suggests organizing shifts by where they enter the data-generating process --- TabPFN's synthetic tasks are drawn from randomly sampled structural causal models, so one can ask which interventions on the SCM a label-free monitor can see. A shift in the label prior $p(y)$ changes the mixture of class-conditional covariate distributions and therefore moves $p(X)$; a concept shift that rewrites $p(y \mid X)$ while leaving $p(X)$ fixed is invisible in principle, however harmful. We seek to determine whether the deployed detector respects this taxonomy, using controlled shift injections where the benchmark's natural shifts cannot separate the two.

\begin{figure}[t]
\centering
\includegraphics[width=0.8\linewidth]{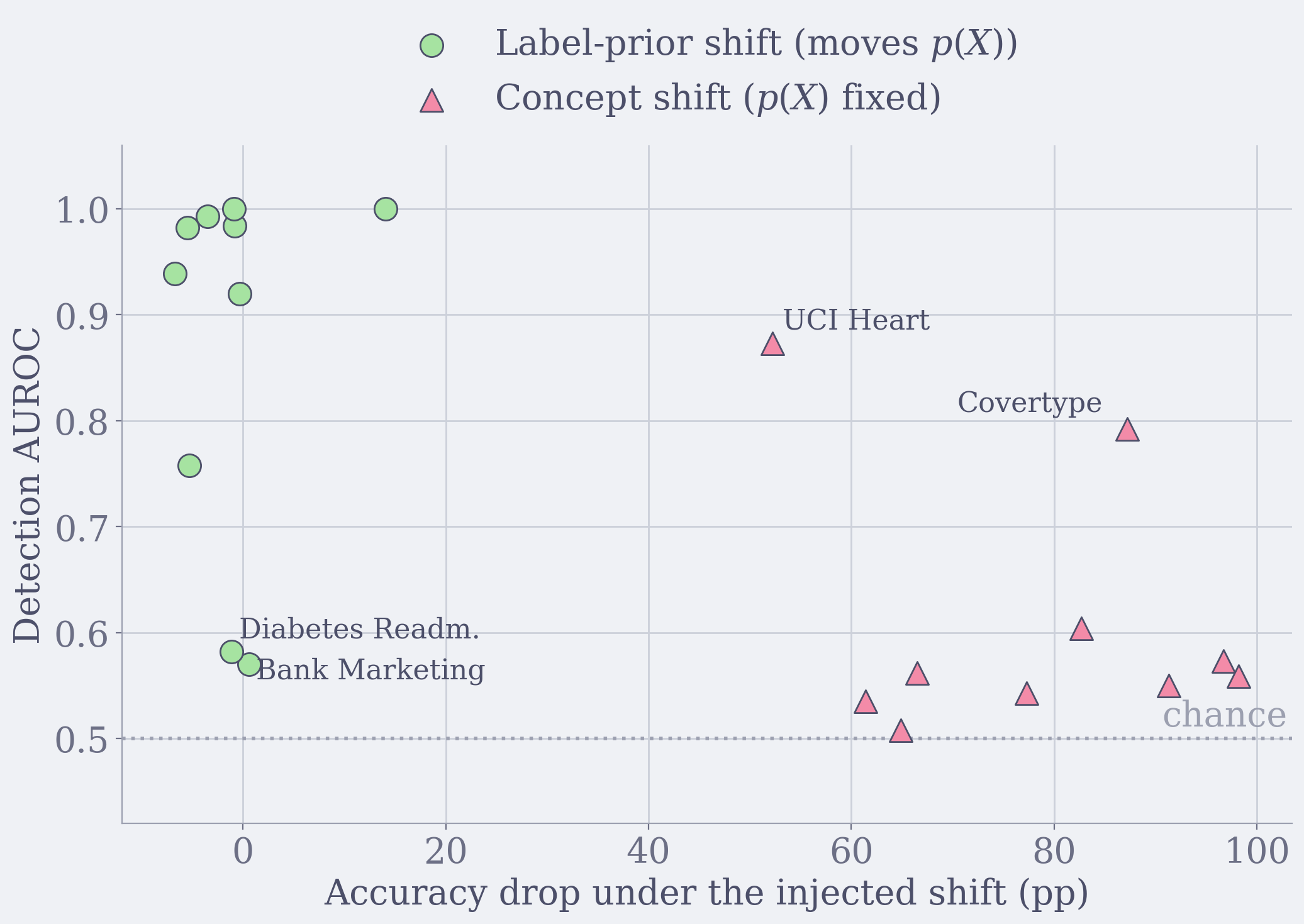}
\caption{\textbf{Label-prior vs.\ concept shift at full strength.} Each point is one dataset under one injected shift (TabPFN, fixed $\eta = 10^{-3}$, 2 seeds; dotted line: chance). Concept shifts (triangles) destroy accuracy yet sit at their no-shift detection baseline; label-prior shifts (circles) barely move accuracy yet are detected wherever they move $p(X)$. The annotated points are the exceptions discussed in the text.}
\label{fig:abl-shift-type}
\end{figure}

We set up the experiment as follows: TabPFN, fixed $\eta = 10^{-3}$, 2 seeds. For label-prior shift we resample the evaluation batch's class proportions toward a skewed target prior at full strength, keeping each class-conditional covariate distribution intact; for concept shift we relabel the evaluation batch through a permuted class map at full strength, changing $p(y \mid X)$ while leaving the covariates untouched. Figure~\ref{fig:abl-shift-type} plots detection against the harm each injected shift inflicts; Table~\ref{tab:abl-shift-type} reports detection AUROC at full shift strength against the no-shift baseline, alongside the accuracy drop the concept shift inflicts.

\begin{table}[t]
\centering
\small
\caption{\textbf{Label-prior vs.\ concept shift at full strength.} Detection AUROC (TabPFN, fixed $\eta = 10^{-3}$, 2 seeds; within-configuration, not comparable to Table~\ref{tab:detection}). The no-shift column is the same protocol at strength 0 and marks each dataset's chance baseline. Concept shift destroys accuracy while leaving the covariates untouched, and detection stays at its baseline; label-prior shift moves the covariate marginal and is detected wherever that movement is substantial.}
\label{tab:abl-shift-type}
\begin{tabular}{lcccc}
\toprule
Dataset & No shift & Label-prior & Concept & Concept $\Delta$acc (pp) \\
\midrule
Adult Income & 0.51 & 0.98 & 0.51 & $-$64.9 \\
Bank Marketing & 0.61 & 0.57 & 0.60 & $-$82.7 \\
Camelyon17 (DINOv2) & 0.55 & 0.98 & 0.55 & $-$91.3 \\
Covertype & 0.57 & 1.00 & 0.79 & $-$87.2 \\
Credit Default & 0.57 & 0.94 & 0.56 & $-$66.5 \\
Diabetes Readmission & 0.55 & 0.58 & 0.54 & $-$77.3 \\
Folktables & 0.56 & 0.99 & 0.54 & $-$61.4 \\
PACS & 0.60 & 1.00 & 0.57 & $-$96.7 \\
UCI Heart & 0.88 & 0.76 & 0.87 & $-$52.2 \\
Waterbirds & 0.56 & 0.92 & 0.56 & $-$98.2 \\
\bottomrule
\end{tabular}
\end{table}

The detector obeys the taxonomy almost exactly. Concept shift is invisible: on nine of ten datasets, AUROC under a full-strength concept shift stays within $0.03$ of the no-shift baseline even as accuracy falls by 52 to 98 percentage points --- this is the impossibility of Lemma~\ref{lem:density-ratio} realized in its purest form, since no label-free monitor of any kind could detect these shifts. Covertype's $0.79$ is the one visible exception, and it is instructive: its class-stratified evaluation sampling slightly perturbs the covariate distribution as a side effect of the relabeling, and the detector picks up precisely that covariate residue, not the concept change itself. Label-prior shift, by contrast, is detected strongly wherever the resampled prior substantially moves the covariate mixture (seven of ten datasets at $0.92$--$1.00$); it is missed on Bank Marketing and Diabetes Readmission, whose class-conditional covariate distributions overlap heavily, so that even a full prior swap barely moves $p(X)$ --- consistent with the mechanism, since the detector sees only the covariate marginal. UCI Heart is confounded in both directions: a residual covariate quirk of its ID split elevates its no-shift baseline to $0.88$, and neither injected shift moves the detector far from that baseline. \cald{} detects exactly the SCM interventions that reach $p(X)$ and is structurally blind to those that do not; shifts expressible as re-sampling the SCM's exogenous inputs are the detector's natural domain, while mechanism changes with fixed covariates require labels and are out of scope (Appendix~\ref{sec:scope}).

\newpage
\section*{NeurIPS Paper Checklist}

\begin{enumerate}

\item {\bf Claims}
    \item[] Question: Do the main claims made in the abstract and introduction accurately reflect the paper's contributions and scope?
    \item[] Answer: \answerYes{} %
    \item[] Justification: Each claim in the abstract and introduction is substantiated: the algorithm (Section~\ref{sec:method}), calibration-to-harm and benchmark results (Section~\ref{sec:experiments}, Table~\ref{tab:baselines}), and the theoretical guarantees (Theorem~\ref{thm:main-separation-main}, Proposition~\ref{prop:conformal-main}).

\item {\bf Limitations}
    \item[] Question: Does the paper discuss the limitations of the work performed by the authors?
    \item[] Answer: \answerYes{}
    \item[] Justification: Appendix~\ref{sec:scope} catalogues out-of-scope failure modes (label shift, concept shift, adversarial inputs) and the regimes where Theorem~\ref{thm:main-separation}'s guarantee becomes vacuous (near-OOD with overlapping supports).

\item {\bf Theory assumptions and proofs}
    \item[] Question: For each theoretical result, does the paper provide the full set of assumptions and a complete (and correct) proof?
    \item[] Answer: \answerYes{}
    \item[] Justification: Section~\ref{sec:method-theory} presents simplified statements of the main results; complete assumptions and proofs are in Appendix~\ref{sec:theory}, with all assumptions consolidated in Appendix~\ref{sec:theory-assumptions}.

    \item {\bf Experimental result reproducibility}
    \item[] Question: Does the paper fully disclose all the information needed to reproduce the main experimental results of the paper to the extent that it affects the main claims and/or conclusions of the paper (regardless of whether the code and data are provided or not)?
    \item[] Answer: \answerYes{}
    \item[] Justification: Algorithm~\ref{alg:method} gives the full pseudocode; Section~\ref{sec:experimental-setup} and Appendix~\ref{sec:experimental-details} list all datasets, baselines, hyperparameters, and per-dataset preprocessing.

\item {\bf Open access to data and code}
    \item[] Question: Does the paper provide open access to the data and code, with sufficient instructions to faithfully reproduce the main experimental results, as described in supplemental material?
    \item[] Answer: \answerYes{}
    \item[] Justification: Code with reproduction scripts is released at \url{https://github.com/teivng/cald}; all datasets are publicly available and their sources are described in Appendix~\ref{sec:dataset-description}.

\item {\bf Experimental setting/details}
    \item[] Question: Does the paper specify all the training and test details (e.g., data splits, hyperparameters, how they were chosen, type of optimizer) necessary to understand the results?
    \item[] Answer: \answerYes{}
    \item[] Justification: Section~\ref{sec:experimental-setup} specifies the setup; Appendix~\ref{sec:experimental-details} lists all hyperparameter values ($T=50$, $\eta \in \{10^{-2}, 10^{-3}, 10^{-4}\}$ selected per cell by the tail-stability rule, $K=50$, $\alpha=0.05$) and per-dataset configurations.

\item {\bf Experiment statistical significance}
    \item[] Question: Does the paper report error bars suitably and correctly defined or other appropriate information about the statistical significance of the experiments?
    \item[] Answer: \answerYes{}
    \item[] Justification: Results are reported as means with standard deviations across 10 seeds (Appendix~\ref{sec:main-results-full}); Type-I error is closed-form-bounded by $\alpha + 1/(K+1)$ via Proposition~\ref{prop:conformal-main}.

\item {\bf Experiments compute resources}
    \item[] Question: For each experiment, does the paper provide sufficient information on the computer resources (type of compute workers, memory, time of execution) needed to reproduce the experiments?
    \item[] Answer: \answerYes{}
    \item[] Justification: All experiments run on a single NVIDIA H100 GPU; total compute is on the order of tens of GPU-hours, with per-step runtimes in Appendix~\ref{sec:abl-budget}.
    
\item {\bf Code of ethics}
    \item[] Question: Does the research conducted in the paper conform, in every respect, with the NeurIPS Code of Ethics \url{https://neurips.cc/public/EthicsGuidelines}?
    \item[] Answer: \answerYes{}
    \item[] Justification: We have reviewed and adhered to the NeurIPS Code of Ethics; the work involves no human subjects or scraped personal data.

\item {\bf Broader impacts}
    \item[] Question: Does the paper discuss both potential positive societal impacts and negative societal impacts of the work performed?
    \item[] Answer: \answerYes{}
    \item[] Justification: Positive impact is safer deployment of tabular foundation models in high-stakes domains; the main risk is over-trusting the test outside its proven scope, which Appendix~\ref{sec:scope} explicitly catalogues to mitigate.
    
\item {\bf Safeguards}
    \item[] Question: Does the paper describe safeguards that have been put in place for responsible release of data or models that have a high risk for misuse (e.g., pre-trained language models, image generators, or scraped datasets)?
    \item[] Answer: \answerNA{}
    \item[] Justification: The paper releases no high-misuse-risk artifacts; \cald is a deployment-time monitor over existing pretrained PFNs.

\item {\bf Licenses for existing assets}
    \item[] Question: Are the creators or original owners of assets (e.g., code, data, models), used in the paper, properly credited and are the license and terms of use explicitly mentioned and properly respected?
    \item[] Answer: \answerYes{}
    \item[] Justification: All models (TabPFN, TabICL) and datasets are cited and used under their original licenses; dataset sources are described in Appendix~\ref{sec:dataset-description}.

\item {\bf New assets}
    \item[] Question: Are new assets introduced in the paper well documented and is the documentation provided alongside the assets?
    \item[] Answer: \answerYes{}
    \item[] Justification: The \cald implementation is released at \url{https://github.com/teivng/cald} with a README documenting the API and run commands.

\item {\bf Crowdsourcing and research with human subjects}
    \item[] Question: For crowdsourcing experiments and research with human subjects, does the paper include the full text of instructions given to participants and screenshots, if applicable, as well as details about compensation (if any)?
    \item[] Answer: \answerNA{}
    \item[] Justification: The paper does not involve crowdsourcing or research with human subjects.

\item {\bf Institutional review board (IRB) approvals or equivalent for research with human subjects}
    \item[] Question: Does the paper describe potential risks incurred by study participants, whether such risks were disclosed to the subjects, and whether Institutional Review Board (IRB) approvals (or an equivalent approval/review based on the requirements of your country or institution) were obtained?
    \item[] Answer: \answerNA{}
    \item[] Justification: The paper does not involve human subjects, so IRB approval is not applicable.

\item {\bf Declaration of LLM usage}
    \item[] Question: Does the paper describe the usage of LLMs if it is an important, original, or non-standard component of the core methods in this research? Note that if the LLM is used only for writing, editing, or formatting purposes and does \emph{not} impact the core methodology, scientific rigor, or originality of the research, declaration is not required.
    \item[] Answer: \answerNA{}
    \item[] Justification: LLMs are not used as any component of the core methodology. The PFNs studied (TabPFN, TabICL) are tabular in-context learners, not LLMs.

\end{enumerate}
    
\end{document}